%% file: stylereiser.tex
\documentclass{article}

\usepackage{arxiv}

\usepackage{booktabs} 
\usepackage{xcolor}
\usepackage{import}
\usepackage{array}
\usepackage{multirow}
\usepackage{hyperref}
\usepackage{graphicx}
\usepackage{tabularx}
\usepackage{url}
\usepackage{amsfonts}
\usepackage{microtype}
\usepackage{natbib}
\usepackage{doi}
\usepackage{orcidlink}
\usepackage{picture}
\usepackage{color}
\usepackage{amsmath}
\usepackage{enumitem}

\newcommand{\shortcite}[1]{\citeyearpar{#1}}

\newcommand{\targetframefigure}{$\mathbf{y}$}

\usepackage[ruled]{algorithm2e} 

\SetAlFnt{\small}
\SetAlCapFnt{\small}
\SetAlCapNameFnt{\small}
\SetAlCapHSkip{0pt}

\begin{document}

\title{StructuReiser: A Structure-preserving Video Stylization Method} 

\author{Radim Spetlik \orcidlink{0000-0002-1423-7549} \\
	CTU in Prague, FEE\\
	\texttt{spetlrad@fel.cvut.cz} \\
	\And
	David Futschik \orcidlink{0000-0003-3254-0290}\\
	Google\\
	\And
	Daniel S\'{y}kora \orcidlink{0000-0002-6145-5151}\\
	CTU in Prague, FEE\\
}

\def\fg#1{Fig.~\ref{fig:#1}}

\def\img#1#2#3{\def\svgwidth{\hsize}\import{figures/#1/}{#1.tex}\caption{#3}\label{fig:#2}}

\keywords{video style transfer, exemplar-based}

\maketitle

\begin{abstract}
We introduce StructuReiser, a novel video-to-video translation method that transforms input videos into stylized sequences using a set of user-provided keyframes. Unlike existing approaches, StructuReiser maintains strict adherence to the structural elements of the target video, preserving the original identity while seamlessly applying the desired stylistic transformations. This enables a level of control and consistency that was previously unattainable with traditional text-driven or keyframe-based methods. Furthermore, StructuReiser supports real-time inference and custom keyframe editing, making it ideal for interactive applications and expanding the possibilities for creative expression and video manipulation.
\end{abstract}

\section{Introduction}

Guided video stylization has received considerable attention recently. The example-based methods~\cite{jamriska_stylizing_2019,texler_interactive_2020,futschik_stalp_2021} offer users full creative freedom by allowing them to directly manipulate the appearance through one or more stylized keyframes. Alternatively, diffusion models~\cite{yang_rerender_2023,ceylan_pix2video_2023,geyer_tokenflow_2024} provide a simplified workflow, enabling users to modify appearance using various text prompts. 

Despite their practicality and impressive results, both keyframe-based and text-driven approaches share a significant limitation that can impact the quality and usability of generated content:
they transfer the style without adequately preserving the structure of the target content. 
For example, when the input video features a distinct character whose identity is crucial (see~\fg{teaser}a), there is no guarantee that the stylized sequence will retain this identity (see~\fg{teaser}d--e). To address this issue in keyframe-based approaches, users must provide a set of keyframes that accurately capture existing and newly appearing structural elements in the input video -- a process that can be tedious and time-consuming. The challenge is even greater with text-driven methods, where crafting a prompt that ensures the preservation of structural details is nearly impossible.

To overcome these challenges, we propose a novel approach to keyframe-based video stylization. The core idea of our solution is a new formulation of guided video stylization that, in addition to ensuring fidelity to the transferred style, also focuses on preserving the structural elements of the input video (cf.~\fg{teaser}b--c). 
To the best of our knowledge, this is the first attempt to address structural fidelity in both text-driven and keyframe-based video stylization. This improvement not only reduces the manual effort required for keyframe preparation, but also prevents the unintended introduction of structural inconsistencies that can arise in text-driven approaches. 

To validate the effectiveness of our method, we conducted extensive qualitative and quantitative evaluations, including an online user survey, which demonstrated significant improvements over the current state-of-the-art. Moreover, our method supports real-time inference, making it suitable for interactive scenarios such as video conferencing, where text-driven approaches are difficult to apply.

\paragraph{To summarize our contributions:}
\begin{enumerate}[nosep]
\renewcommand{\labelenumi}{(\roman{enumi})}
    \item We formulate the task of structurally faithful stylization of videos, establishing a practical framework that emphasizes both stylistic fidelity and structural preservation.
    \item Building on this framework, we introduce a novel keyframe-based video stylization method that effectively transfers style from a single stylized keyframe while preserving essential structural elements in the entire video sequence.
    \item We validate our approach through extensive qualitative and quantitative evaluations, showing a significant improvement over current state-of-the-art in maintaining structural fidelity while closely adhering to the desired stylization.
\end{enumerate}

\begin{figure}
\def\svgwidth{\hsize}\import{figures/teaser/}{teaser.tex}\caption{StructuReiser transfers the style from a single stylized keyframe~(a) to the entire video sequence~(b) generating stylized frames~(c) that are both stylistically consistent and structurally faithful. The keyframe~(a) was created using the text-guided video-to-video diffusion model by Ceylan et al.~\shortcite{ceylan_pix2video_2023}. However, when applied directly to other frames in the sequence, this model often introduces significant structural inconsistencies~(d). The state-of-the-art keyframe-based video stylization method of Futschik et al.~\shortcite{futschik_stalp_2021} faces similar issues~(e). In contrast, our approach~(c) maintains the structural integrity of the target video sequence while ensuring coherent stylization throughout.}\label{fig:teaser}
\end{figure}
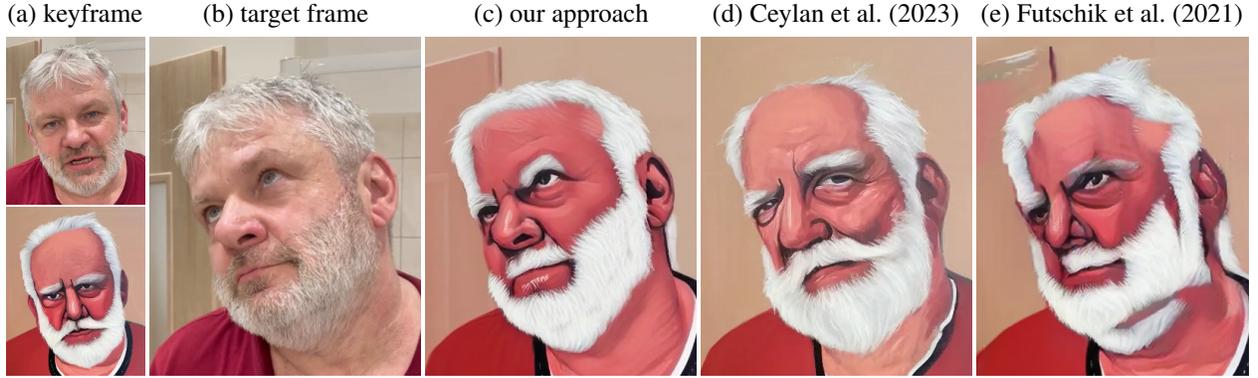

\section{Related work}

The origin of image and video stylization techniques can be traced back several decades. Early stylization approaches were typically based on hand-crafted algorithmic solutions that were restricted to a certain range of styles and specific target domains. For instance, Curtis et al.~\shortcite{curtis_computer-generated_1997} runs a physically-based simulation to mimic the appearance of a watercolor media, Salisbury et al.~\shortcite{salisbury_orientable_1997} produces painterly artworks automatically using a set of predefined brush strokes, while Praun et al.~\shortcite{praun_real-time_2001} can generate brush strokes procedurally. Despite the impressive results these early stylization techniques produce, their main limitation lies in the fact that slight modification of an existing style or creation of a new one usually requires a significant effort and expertise.

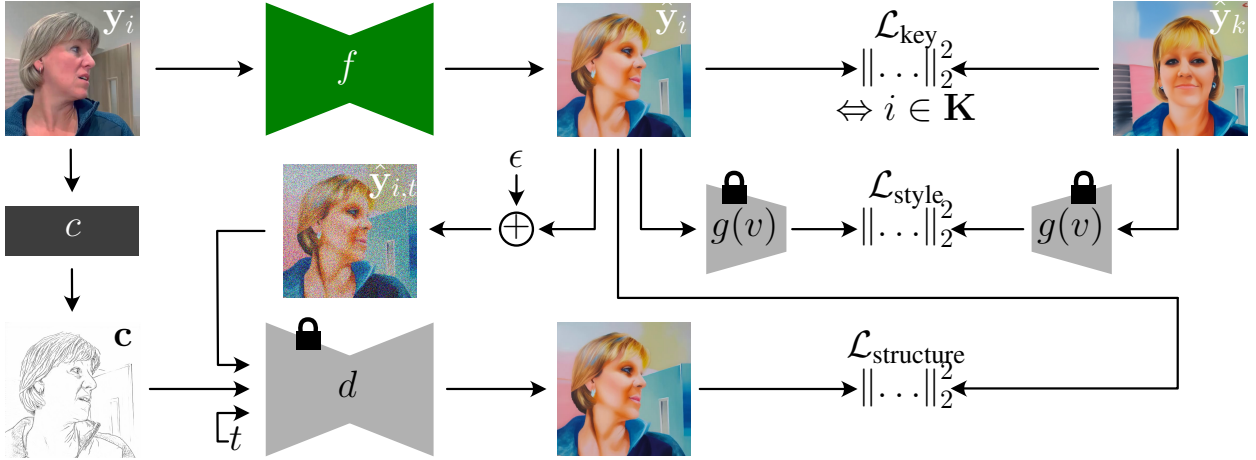
\begin{figure*}
\def\svgwidth{\hsize}\import{figures/overview/}{overview.tex}\caption{An overview of our approach. Given images from a source domain~$\mathbf{y}_i \in \mathcal{Y}$, we optimize the operator~$f$ to produce images~$\hat{\mathbf{y}}_i$ with a similar appearance as images from a target domain~$\hat{\mathbf{x}}_i$. The \emph{key} loss~$\mathcal{L}_{\text{key}}$~(\ref{eq:loss_key}) encourages reconstruction of keyframes~$\hat{\mathbf{x}}_i$, the \emph{style} loss~$\mathcal{L}_{\text{style}}$~(\ref{eq:loss_vgg}) ensures style consistency between frames and keyframes using Gram correlation matrices~$g$ of extracted VGG network responses~$v$~\cite{gatys_image_2016}, and finally the \emph{structure} loss~$\mathcal{L}_{\text{structure}}$~(\ref{eq:loss_sds}) enforces fidelity to structural elements present in the input video frames~$\mathbf{y}_i$. The \emph{structure} loss requires a pre-trained ControlNet~\cite{zhang_adding_2023} model consisting of diffusion model~$d$ initialized by adding a random Gaussian noise~$\mathbf{\epsilon}\sim\mathcal{N}(\mathbf{0},\mathbf{I})$ into the synthesized image~$\hat{\mathbf{y}}_i$, a time step~$t$, and a function~$c$ transforming the input image~$\mathbf{y}_i$ to a condition~$\mathbf{c}$ (in this case Canny edge detector~\cite{canny_computational_1986}).}\label{fig:overview}
\end{figure*}

To overcome this limitation, Hertzmann et al.~\shortcite{hertzmann_image_2001} introduced the idea of image analogies. In their approach, the user can provide an example pair of unstyled and stylized images that specify the intended stylization analogy. The resulting image is then constructed by copying patches from a stylized exemplar so that the corresponding pixels in the unstyled patches match the patches in the target unstyled image. The framework of image analogies later proved to be a viable solution also for example-based video stylization~\cite{benard_stylizing_2013,jamriska_stylizing_2019} that can deliver temporally consistent sequences that faithfully preserve the user-specified visual style.  A key limitation of those techniques is that they treat the target video as a guide for style transfer, and thus larger structural changes that may appear in the target domain are not taken into account. To overcome this limitation, the user needs to provide multiple consistently stylized keyframes, of which manual preparation can be labor intensive. When stylized keyframes are generated synthetically, it is, on the other hand, difficult to ensure their consistency.

Those limitations were addressed by Frigo et al.~\shortcite{frigo_split_2016} and Gatys et al.~\shortcite{gatys_image_2016} who perform stylization using only the style image and try to better respect the structural changes in the target domain. Frigo et al.~search for the optimal mapping between the adaptively sized patches in the target image and patches in the style-exemplar while Gatys et al.~iteratively optimizes the output image so that when fed into the VGG network~\cite{simonyan_very_2014} its responses correspond to VGG responses of the style exemplar and the target image. This approach inspired follow-up works~\cite{li_universal_2017,kolkin_style_2019} of which aim is to increase the faithfulness of the generated image to the style exemplar by employing more sophisticated loss functions. Chen et al.~\shortcite{chen_coherent_2017} and Ruder et al.~\shortcite{ruder_artistic_2018} later demonstrated how to extend the framework of Gatys et al. to example-based video stylization delivering temporally consistent sequences. Although those approaches are fully automatic and do not require preparation of a larger number of stylized keyframes, their artistic control over the final output is fairly limited. The transfer is usually not semantically meaningful and lacks faithfulness to the original artistic media.  

To perform a semantically meaningful transfer while respecting structural changes in the target domain, image-to-image translation networks were proposed~\cite{johnson_perceptual_2016,isola_image--image_2017}. However, these require a large amount of training data to work reliably. Only in some domain-specific scenarios, such as portrait stylization~\cite{futschik_real-time_2019} training pairs can be generated automatically~\cite{fiser_example-based_2017}. 

To mitigate the requirement for larger paired datasets, few-shot learning approaches~\cite{liu_few-shot_2019,wang_few-shot_2019} and deformation-based approaches~\cite{siarohin_first_2019,siarohin_animating_2019} were proposed. However, these methods require pretraining on large domain-specific datasets and thus are not applicable in the general case. Texler et al.~\shortcite{texler_interactive_2020} proposed a few-shot patch-based training strategy for which only a~few stylized keyframes are necessary to deliver compelling video stylization results without the requirement of domain-specific pre-training. However, their method has limitations comparable to the image analogies approach of Jamri\v{s}ka et al.~\shortcite{jamriska_stylizing_2019}, i.e., when new structural details appear in the target sequence, it is necessary to provide additional stylized keyframes. Although in the follow-up work of Futschik et al.~\shortcite{futschik_stalp_2021} the amount of required keyframes decreased significantly, the newly appearing structural changes cause difficulties as the underlying approach is still predominantly focused on style preservation.

As an alternative approach to video stylization, unwrapping techniques have been proposed~\cite{rav-acha_unwrap_2008,kasten_layered_2021}. In those approaches, input video frames are first projected onto a static atlas where edits can be performed at one snap and later transferred back to the original video domain. The quality of results is highly dependent on the quality of the generated unwrap. For small local edits, those techniques produce impressive results, but larger changes typically cause difficulties. 

Recent approaches to video stylization utilize large pre-trained text-to-image diffusion models~\cite{rombach_high-resolution_2022} and can edit videos globally using text prompts~\cite{khandelwal_infusion_2023,yang_rerender_2023,zhang_towards_2023,chu_medm_2024,geyer_tokenflow_2024,kim_collaborative_2024,shin_edit--video_2024}. However, the outputs of these techniques heavily depend on a particular version of the pre-trained text-to-image diffusion model, which may tend to produce unpredictable results that do not consistently reproduce structural changes in the target video sequence. This leads to a typical ubiquitous structural flicker that can be disturbing to the observer. Moreover, in addition to the text prompt, additional control over the stylization process is difficult to achieve, in contrast to keyframe-based methods where the user has full creative freedom~\cite{jamriska_stylizing_2019,texler_interactive_2020,futschik_stalp_2021}.

\section{Our Approach}

The input of our method is a set of~$N$ video frames~$\mathcal{T}$ which is partitioned into two subsets:~$\mathcal{T} = \mathcal{X} \cup \mathcal{Y}$.
\begin{enumerate}
    \item \emph{Keyframes}~$\mathcal{X}= \{ \mathbf{x}_i \in \mathcal{T} \mid i \in \mathbf{K} \}$, where~$\mathbf{K}$ is a set of~$K$ frame indices. For each keyframe $\mathbf{x}_i$, a ground truth stylized version~~$\mathbf{\hat{x}} \in \mathcal{\hat{X}}$ is provided.
    \item \emph{Non-keyframes}~$\mathcal{Y} = \{ \mathbf{y}_i \in \mathcal{T} \mid i \not\in \mathbf{K} \}$, consisting of the remaining~$N - K$ frames without stylized counterparts. 
\end{enumerate}
We define the set of keyframe pairs as $\mathcal{K} = \{(\mathbf{x}_i, \mathbf{\hat{x}}_i) \in \mathcal{X} \times \mathcal{\hat{X}} \mid i \in \mathbf{K} \}$, which contains $K$~tuples of original keyframes and their corresponding stylized versions.

The aim of our approach is to learn a stylization operator~$f$ that maps any input video frame~$\mathbf{z}_i \in \mathcal{Z}$ into its stylized counterpart~$\hat{\mathbf{z}}_i \in \mathcal{\hat{Z}}$, i.e., $\hat{\mathbf{z}}_i = f(\mathbf{z}_i)$, by minimizing the following loss function:
\begin{align}\label{eq:loss}
    \mathcal{L}(\mathcal{K}, \mathcal{Y}) = 
      \lambda_k \mathcal{L}_{\text{key}}(\mathcal{K})
    + \lambda_v \mathcal{L}_{\text{style}}(\mathcal{K}, \mathcal{Y})
    + \lambda_s \mathcal{L}_{\text{structure}}(\mathcal{Y}).
\end{align}

\begin{figure*}
\def\svgwidth{\hsize}\import{figures/sota_comparison_dad/}{sota_comparison_dad.tex}\caption{%
Results of our approach in comparison with the state-of-the-art in diffusion-based video stylization: The target video sequence (see a representative target frame~\targetframefigure{}) has been stylized using diffusion-based approaches (top row):  
(a)~Ceylan et al.~\shortcite{ceylan_pix2video_2023}, (b)~Yang et al.~\shortcite{yang_rerender_2023}, (c)~Chu et al.~\shortcite{chu_medm_2024}, and (d)~Geyer et al.~\shortcite{geyer_tokenflow_2024}. One frame from those stylized sequences was used as a keyframe (see small insets). The style of this keyframe has been propagated to the rest of the target sequence~$\textbf{y} \in \mathcal{Y}$ using our approach~(bottom row). Note how our approach better preserves the structural details seen in the target frame. Also, see our supplementary video to compare consistency across the entire sequence. Diffusion-based approaches tend to suffer from notable structural flicker, whereas our approach keeps the structure consistent, yielding considerably more stable results.
}\label{fig:sota_comparison_dad}
\end{figure*}

The~\emph{reconstruction loss}~$\mathcal{L}_{\text{key}}$ guides the~$f$ operator to generate outputs that closely match the user-provided stylized keyframes, i.e.:
\begin{equation}\label{eq:loss_key}
    \mathcal{L}_{\text{key}}(\mathcal{K}) = \frac{1}{K} \sum_{i \in \mathbf{K}} \left\| f \left(\mathbf{x}_i\right) - \mathbf{\hat{x}}_i \right\|_2^2.
\end{equation}

The~\emph{style loss}~$\mathcal{L}_{\text{style}}$ originally proposed by Gatys et al.~\shortcite{gatys_image_2016} ensures that the distribution of stylized frames~$f(\mathbf{y}_i) \in \mathcal{\hat{Y}}$ aligns with that of the stylized keyframes~$f(\mathbf{x}_i) \in \mathcal{\hat{X}}$. This is achieved by comparing responses of the VGG network~\cite{simonyan_very_2014} as follows:
\begin{align}
\label{eq:loss_vgg}
    \mathcal{L}_{\text{style}}(\mathcal{K}, \mathcal{Y})
    = \frac{1}{|\mathbf{K}| |\mathbf{L}| |\mathcal{Y}|} \sum_{i \in \mathbf{K}} \sum_{j \not\in \mathbf{K}} &\sum_{l \in \mathbf{L}} \left\| g\left(\mathbf{\hat{x}}_i, l\right) - g\left(f(\mathbf{y}_j), l\right) \right\|_2^2
\end{align}
where~$g(\mathbf{\cdot})$ computes the Gram correlation matrix of VGG network responses at the layer~$l\in \mathbf{L}$ and~$\mathbf{L}$ is a set of indices of all considered VGG layers.

The~\emph{structure loss}~$\mathcal{L}_{\text{structure}}$ is the key component and the main contribution of our framework. It guides the operator~$f$ in transferring essential structural elements from the input video sequence to the stylized output. We compute this loss using a diffusion model~$d$, conditioned by~$c$, to ensure preservation of important structural details:
\begin{equation}\label{eq:loss_sds}
    \mathcal{L}_{\text{structure}}(\mathcal{Y}) = \frac{1}{|\mathcal{Y}|} \sum_{i \not\in \mathbf{K}} \left\| d\left(\hat{\mathbf{y}}_{i,t}, c(\textbf{y}_i), t\right) - \epsilon \right\|_2^2,
\end{equation}
where $\hat{\mathbf{y}}_{i,t} = \sqrt{\bar{\alpha}_t} f(\mathbf{y}_i) + \sqrt{1 - \bar{\alpha}_t}\mathbf{\epsilon}$ is a noisy version of the stylized input frame~$\mathbf{y}_i$ corrupted by a Gaussian noise~$\mathbf{\epsilon} \sim \mathcal{N}(\mathbf{0}, \mathbf{I})$ at a pre-defined time step~$t$.

In \fg{teaser}c, $d$ is a pre-trained diffusion model ControlNet~\cite{zhang_adding_2023}, conditioned by a line art detector~$c$. 
However, this is just one of many possible choices. We further explore different conditionings~$c$ of the diffusion model~$d$ in~Sec.~\ref{sec:conditioning}.

Our strategy to computing~$\mathcal{L}_{\text{structure}}$ shares some aspects with the \emph{score distillation sampling} (SDS) method proposed by Poole et al.~\shortcite{poole_dreamfusion_2022}, however, a key difference in our solution is its ability to preserve prescribed structural details, which is not possible with the original SDS. In~Sec.~\ref{sec:difflineart} we also demonstrate the importance of this particular formulation in contrast to a straightforward approach, i.e., directly computing the structure loss using a line art detector.

\paragraph{Definition of ``structure''} We use terms like ``structure'' and ``structural fidelity'' to describe key visual elements from the input video we aim to preserve. However, we acknowledge that precise definitions of these terms are challenging and interpretations may vary.
Fortunately, ``structure'' is defined inherently in our approach -- by the choice of conditioning~$c$ in the diffusion model~$d$,  determining the meaning of ``structure'' by design.

\begin{table}[t]
\centering
\begin{tabular}{l|c|c|c}
 & SSIM $\uparrow$ & LPIPS $\downarrow$ & \reflectbox{F}LIP $\downarrow$ \\
\hline
\hline
ours & \textbf{0.75} $\pm$ 0.12 & \textbf{0.28} $\pm$ 0.09 & \textbf{0.30} $\pm$ 0.06 \\
Geyer et al.~\shortcite{geyer_tokenflow_2024} & 0.72 $\pm$ 0.10 & 0.30 $\pm$ 0.08 & 0.31 $\pm$ 0.05 \\
\hline
ours & \textbf{0.71} $\pm$ 0.10 & \textbf{0.37} $\pm$ 0.09 & \textbf{0.43} $\pm$ 0.08 \\
Yang et al.~\shortcite{yang_rerender_2023} & 0.65 $\pm$ 0.09 & 0.42 $\pm$ 0.09 & 0.47 $\pm$ 0.08 \\
\hline
ours & \textbf{0.71} $\pm$ 0.09 & \textbf{0.38} $\pm$ 0.10 & \textbf{0.40} $\pm$ 0.10 \\
Ceylan et al.~\shortcite{ceylan_pix2video_2023} & 0.67 $\pm$ 0.08 & 0.41 $\pm$ 0.10 & 0.42 $\pm$ 0.10 \\
\hline
ours & \textbf{0.62} $\pm$ 0.07 & \textbf{0.48} $\pm$ 0.10 & \textbf{0.59} $\pm$ 0.11 \\
Chu et al.~\shortcite{chu_medm_2024} & 0.56 $\pm$ 0.07 & 0.55 $\pm$ 0.07 & 0.61 $\pm$ 0.10 \\
\hline
ours & \textbf{0.69} $\pm$ 0.14 & \textbf{0.33} $\pm$ 0.12 & \textbf{0.37} $\pm$ 0.08 \\
Jamri\v{s}ka et al.~\shortcite{jamriska_stylizing_2019} & 0.67 $\pm$ 0.11 & 0.36 $\pm$ 0.10 & 0.38 $\pm$ 0.06 \\
Futschik et al.~\shortcite{futschik_stalp_2021} & 0.66 $\pm$ 0.11 & 0.35 $\pm$ 0.11 & 0.39 $\pm$ 0.07 \\
Texler et al.~\shortcite{texler_interactive_2020} & 0.66 $\pm$ 0.11 & 0.36 $\pm$ 0.09 & 0.38 $\pm$ 0.06 \\
\hline\hline
\end{tabular}
\caption{Quantitative comparison of structural fidelity. Averages and standard deviations between input and stylized frames shown in five groups. 
Since diffusion-based methods cannot perform exemplar-based stylization, we trained our method using style exemplars generated by these methods for the first four groups (see Sec.~\ref{sec:quantitative}). The fifth group includes exemplar-based approaches trained on a union of the diffusion-generated and artist-created exemplars. Our method outperforms existing approaches across all metrics.}
\label{tab:quantitative}
\end{table}

\begin{figure*}
\def\svgwidth{\hsize}\import{figures/sota_comparison_lili/}{sota_comparison_lili.tex}\caption{%
Results of our approach in comparison with the state-of-the-art in diffusion-based video stylization (cont.): See~\fg{sota_comparison_dad} for a detailed explanation.
}\label{fig:sota_comparison_lili}
\end{figure*}

\begin{figure*}
\def\svgwidth{\hsize}\import{figures/sota_comparison_zuzka/}{sota_comparison_zuzka.tex}\caption{%
Results of our approach in comparison with the state-of-the-art in diffusion-based video stylization (cont.): See \fg{sota_comparison_dad} for a detailed explanation.
}\label{fig:sota_comparison_zuzka}
\end{figure*}

\begin{figure*}
\def\svgwidth{\hsize}\import{figures/sota_comparison_joli/}{sota_comparison_joli.tex}\caption{%
Results of our approach in comparison with the state-of-the-art in diffusion-based video stylization (cont.): See \fg{sota_comparison_dad} for a detailed explanation.
}\label{fig:sota_comparison_joli}
\end{figure*}

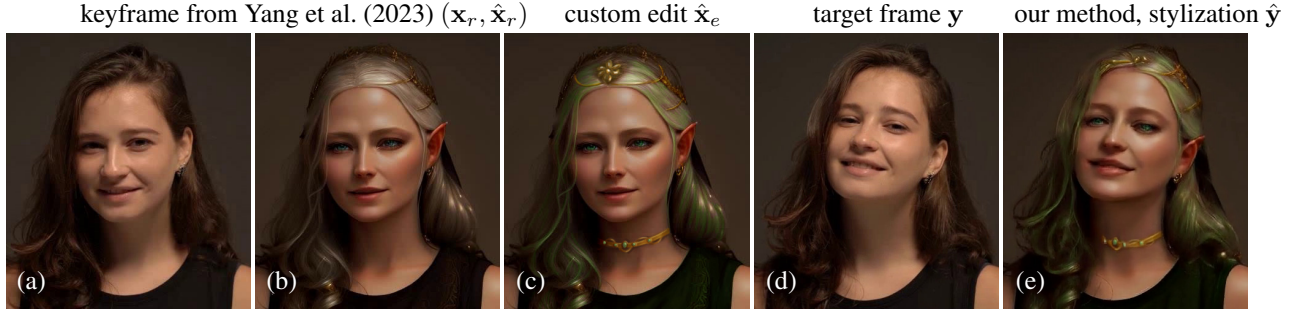
\begin{figure*}
\def\svgwidth{\hsize}\import{figures/edits/}{edits.tex}\caption{Custom edit of results generated by the diffusion-based method of Yang et al.~\shortcite{yang_rerender_2023}. Left to right: (a, b)~keyframe~$(\mathbf{x}_r, \hat{\mathbf{x}}_r)$ is the result of Yang et al.~\shortcite{yang_rerender_2023} conditioned with the text prompt ``Galadriel, the royal Elf, silver-golden hair,'' (c)~custom edit~$\hat{\mathbf{x}}_e$ of the stylized keyframe~$\hat{\mathbf{x}}_r$, (d)~target frame~$\mathbf{y}$, (e)~stylization~$\hat{\mathbf{y}}$ produced by our method with a single keyframe~$(\mathbf{x}_r, \hat{\mathbf{x}}_e)$. A key advantage of our method is that it allows custom edits of videos stylized by diffusion-based methods, which typically offer only limited control over the generated results through textual prompts.}\label{fig:edits}
\end{figure*}

\begin{figure*}
\def\svgwidth{\hsize}\import{figures/sota_comparison_dad_2/}{sota_comparison_dad_2.tex}\caption{%
Results of our approach in comparison with the state-of-the-art in keyframe-based video stylization: The diffusion-based method of Ceylan et al.~\shortcite{ceylan_pix2video_2023} has been used to generate a stylized sequence~(a) from which~$4$~keyframes were selected to perform video stylization using methods of Jamri\v{s}ka et al.~\shortcite{jamriska_stylizing_2019}~(b) and~Texler et al.~\shortcite{texler_interactive_2020}~(c), and~$1$ keyframe was selected for the method of Futschik et al.~\shortcite{futschik_stalp_2021}~(d) and our approach~(e). Note how our approach better preserves the structural details seen in the target frame. In our supplementary video, it is also visible that our approach keeps the structure consistent, yielding notably more stable results.
}\label{fig:sota_comparison_dad_2}
\end{figure*}

\begin{figure*}
\def\svgwidth{\hsize}\import{figures/sota_comparison_lili_2/}{sota_comparison_lili_2.tex}\caption{%
Results of our approach in comparison with the state-of-the-art in keyframe-based video stylization (cont.): 
Diffusion-based method of Geyer et al.~\shortcite{geyer_tokenflow_2024} has been used to generate the initial stylized sequence~(a).
See Fig.~\ref{fig:sota_comparison_dad_2} for a detailed explanation.
}\label{fig:sota_comparison_lili_2}
\end{figure*}

\begin{figure*}
\def\svgwidth{\hsize}\import{figures/sota_comparison_zuzka_2/}{sota_comparison_zuzka_2.tex}\caption{%
Results of our approach in comparison with the state-of-the-art in keyframe-based video stylization (cont.): 
Diffusion-based method of Yang et al.~\shortcite{yang_rerender_2023} has been used to generate the initial stylized sequence~(a).
See Fig.~\ref{fig:sota_comparison_dad_2} for a detailed explanation.
}\label{fig:sota_comparison_zuzka_2}
\end{figure*}

\subsection{Implementation details}\label{sec:implementation}

We implemented our approach in PyTorch~\cite{paszke_pytorch_2019} using the AdamW~\cite{loshchilov_decoupled_2017} optimizer with fixed learning rate~$3 \cdot 10^{-5}$. To model the stylization operator~$f$, we adopt the network architecture originally proposed by Futschik et al.~\shortcite{futschik_real-time_2019} that proved to be suitable for style transfer tasks~\cite{texler_arbitrary_2020,futschik_stalp_2021} allowing for reproduction of important high-frequency details, critical for generating complex and believable artistic styles. The batch size was set to~$1$ and instance normalization~\cite{ulyanov_instance_2016} used instead of batch normalization. We set~$\lambda_{k} = 1.0$ and~$\lambda_{v} = 100.0$. The parameters for~$\mathcal{L}_{\text{structure}}$~(\ref{eq:loss_sds}) were selected experimentally and differ between the presented sequences: $\lambda_s \in \{10^{-5}, 10^{-6}\}$ and $t \in \{20, 28\}$. As~$d$, we adopt the default noise scheduler of ControlNet v1.1~\cite{zhang_adding_2023} and the UniPC scheduler~\cite{zhao_unipc_2024} with the number of steps set to~$30$. The parameters of the VGG network and ControlNet were fixed and line art conditioning was used for~$c$ (if not stated otherwise). The training was performed on a single NVIDIA A100 GPU with~$40$ GB of RAM for~$4$ hours and the model with the lowest total loss has been selected to produce the stylized sequences. 

\begin{figure*}[t]    
\def\svgwidth{\hsize}\import{figures/perceptual_study_temporal_consistency_style_structure/}{perceptual_study_temporal_consistency_style_structure.tex}\caption{Perceptual study. Each point represents the ratio of votes preferring the results of our method over those of other methods, based on responses from a total of $54$ participants. Comparisons were made against four diffusion-based methods -- 
    Geyer et al.~\shortcite{geyer_tokenflow_2024} (pink), Yang et al.~\shortcite{yang_rerender_2023} (orange), Chu et al.~\shortcite{chu_medm_2024} (brown), Ceylan et al.~\shortcite{ceylan_pix2video_2023} (purple) -- and three exemplar-based methods -- 
    Futschik et al. \shortcite{futschik_stalp_2021} (green), Texler et al. \shortcite{texler_interactive_2020} (blue), and
    Jamri\v ska et al. \shortcite{jamriska_stylizing_2019} (red). The bottom x-axis displays the ratio of answers favoring our method for preserving temporal consistency (hollow squares), the top x-axis shows the ratio favoring our method for structure preservation (hollow triangles), and the y-axis represents the style reproduction ratio. The graph illustrates that our method significantly outperforms previous works in reproducing input structures and maintaining temporal consistency. Conversely, prior methods are preferred for style preservation, which is expected given our focus on structural fidelity.}\label{fig:perceptual_study_temporal_consistency_style_structure}
\end{figure*}
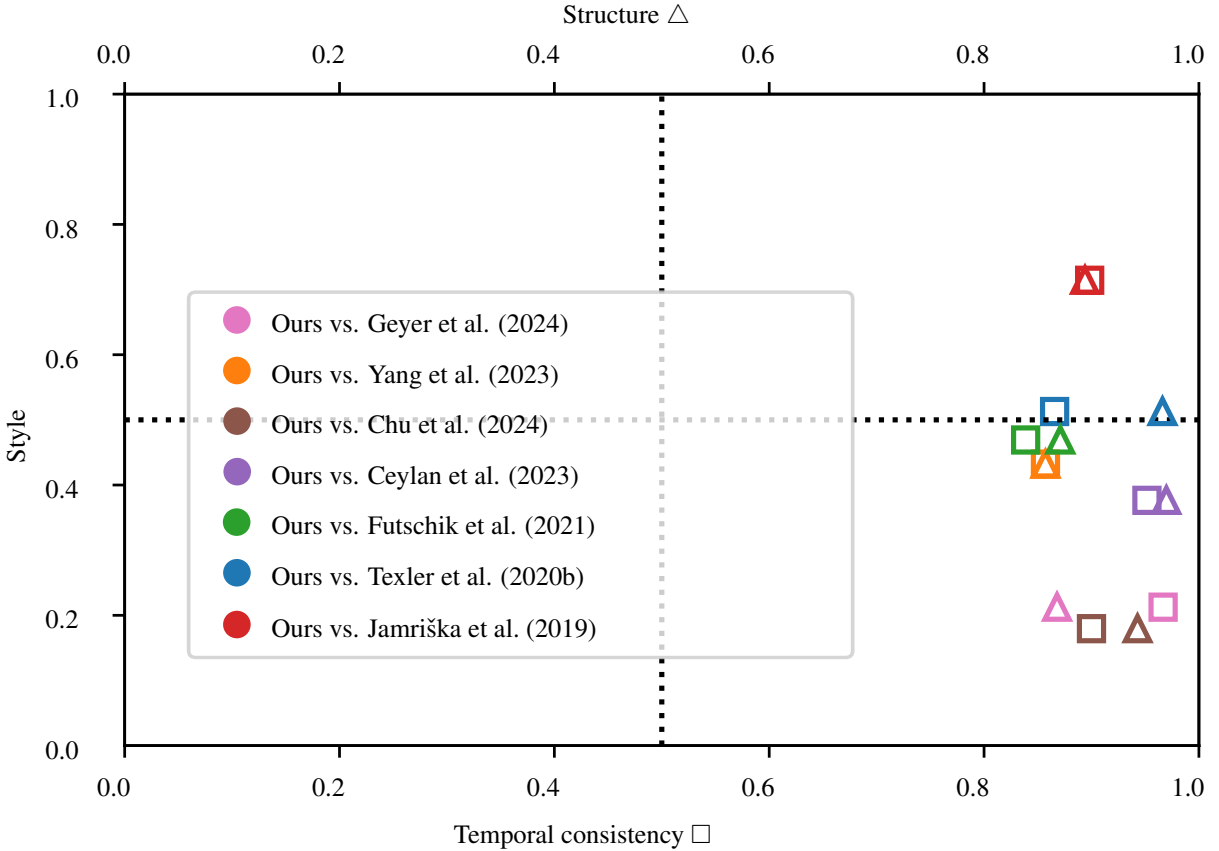

\section{Results and Comparison}\label{sec:results_and_comparison}

The results are presented in Figures~\ref{fig:teaser}, \ref{fig:sota_comparison_dad}, \ref{fig:sota_comparison_lili},  \ref{fig:sota_comparison_zuzka}, and \ref{fig:sota_comparison_joli}. See also our supplementary material for additional results. We compare them with the output of recent diffusion-based methods~\cite{ceylan_pix2video_2023,chu_medm_2024,geyer_tokenflow_2024,yang_rerender_2023} and keyframe-based approaches~\cite{futschik_stalp_2021,jamriska_stylizing_2019,texler_interactive_2020}.

\subsection{Perceptual study}

To qualitatively evaluate our approach, we conducted a perceptual study comparing our method's outputs with those from four state-of-the-art diffusion-based techniques (\cite{ceylan_pix2video_2023,chu_medm_2024,geyer_tokenflow_2024,yang_rerender_2023}), and three state-of-the-art exemplar-based methods (~\cite{futschik_stalp_2021,jamriska_stylizing_2019,texler_interactive_2020}). The study assessed how well each method reproduced the artistic style, preserved structural content, and maintained temporal consistency of the input video. We collected data from $54$ participants through an online survey, where participants were presented with randomized two-alternative forced-choice (2AFC) comparisons. Each participant completed $36$ questions, selecting which anonymized stylization better reproduced style (12 questions), preserved content (12 questions), and maitained temporal consistency (12 questions). In each comparison, an output from our method was paired with one from another method using the same input data.

The preference scores for our method versus others are presented in \fg{perceptual_study_temporal_consistency_style_structure}. In the figure, hollow squares represent preferences regarding temporal consistency, while hollow triangles indicate preferences concerning structural elements. Our method significantly outperforms previous works in reproducing input structures, even though it reproduces the exemplar's style slightly less accurately -- which is expected due to our focus on structural preservation. Moreover, our method excels in maintaining the temporal consistency of the stylized output compared to previous approaches.

\subsection{Quantitative evaluation}
\label{sec:quantitative}
To quantitatively evaluate of our method's ability to preserve structural elements from input videos, we conducted a comprehensive comparison across published video stylization techniques. We calculated the averages and standard deviations between corresponding input and stylized frames across all available video sequences, using three metrics: SSIM, LPIPS \cite{zhang_unreasonable_2018}, and \reflectbox{F}LIP \cite{andersson_flip_2020}. The style exemplars were sourced from Geyer et al. (2024), Yang et al. (2023), Ceylan et al. (2023), Chu et al. (2024), and a combination including artist-created stylizations.

Since the published diffusion-based methods cannot perform  exemplar-based stylization, we trained our method using a single style exemplar selected from the output of each diffusion method. We then compared the results of our method with those of the diffusion-based methods, as shown in the first four groups of results in Table~\ref{tab:quantitative}. In the last group, we compare our method with the exemplar-based stylization approaches of Jamri\v{s}ka et al.~\shortcite{jamriska_stylizing_2019}, Futschik et al.~\shortcite{futschik_stalp_2021}, and Texler et al.~\shortcite{texler_interactive_2020}, trained on a set of diffusion-based exemplars and artist-created exemplars. Our method demonstrates superior performance across all sequence groups and metrics, effectively maintaining structural fidelity across diverse style sources.

\subsection{Comparison with diffusion-based methods}
\label{sec:comparison_with_diffusion}
Since diffusion-based methods only support textual guidance and cannot perform keyframe-based stylization directly, we adopt a two-step approach for each video sequence. First, we generate a stylized version using a combined text prompt~$\mathbf{p} = \mathbf{p}_{\text{edit}}\mathbf{p}_{\text{desc}}$, where~$\mathbf{p}_{\text{edit}}$ (e.g., ``hyperrealistic detailed oil painting of'') defines the desired style, and~$\mathbf{p}_{\text{desc}}$ (e.g., ''an old man with a white beard``) describes the content. For instance, $\mathbf{p} =$ ``hyperrealistic detailed oil painting of an old man with a white beard''.
 From the resulting stylized sequence, we then select one frame as a keyframe and propagate this keyframe's style throughout the entire target sequence~$\mathbf{y} \in \mathcal{Y}$ using our proposed keyframe-based stylization method.
As a result, each diffusion-based method is presented with a unique stylization.
Note that in the methods proposed by~Ceylan et al.~\shortcite{ceylan_pix2video_2023} and~Geyer et al.~\shortcite{geyer_tokenflow_2024}, the context description $\mathbf{p}_{\text{desc}}$ serves as an inversion prompt. 

From the results presented in Figures~\ref{fig:sota_comparison_dad}, \ref{fig:sota_comparison_lili}, 
\ref{fig:sota_comparison_zuzka}, and~\ref{fig:sota_comparison_joli} it is apparent that our approach maintains structural details better than diffusion-based approaches. See also our supplementary video that demonstrates structural consistency across the entire sequence contrasting the flicker common in diffusion-based methods.

A key advantage of our approach in contrast to diffusion-based techniques is that it enables the user to incorporate custom edits by manually modifying the output of the diffusion-based method and use it as a newly stylized keyframe for training. This option is beneficial especially in cases when it is difficult to find an accurate text prompt that precisely expresses the desired artistic vision. In Fig.~\ref{fig:edits}, we show the results of a custom edit of an image~$\hat{\mathbf{x}}_r$ stylized by the method of Yang et al.~\shortcite{yang_rerender_2023} with the text prompt ``Galadriel, the royal Elf, silver-golden hair.'' This image was edited by an artist producing the image~$\hat{\mathbf{x}}_e$. Our method was then trained with the keyframe~$(\mathbf{x}_r, \hat{\mathbf{x}}_e)$, rendering the stylization $\hat{\mathbf{y}}$. 


\subsection{Comparison with keyframe-based methods}

To compare our method with other keyframe-based approaches (see Figures~\ref{fig:sota_comparison_dad_2}, \ref{fig:sota_comparison_lili_2},  and~\ref{fig:sota_comparison_zuzka_2}), we used sequences generated by diffusion-based methods. From each stylized sequence, we selected keyframe to train different methods: four keyframes were chosen to train the methods of Jamriška et al.~\shortcite{jamriska_stylizing_2019} and Texler et al.~\shortcite{texler_interactive_2020}, while one keyframe was used to train the method of Futschik et al.~\shortcite{futschik_stalp_2021} and our own method. In all the results presented, our approach consistently maintains the structural details of the target frame~\targetframefigure{} while faithfully replicating important style features of the stylized keyframe. Please refer to our supplementary video to compare the structural consistency across the entire sequence.

\section{Experiments}

In this section, we present a set of experiments that provide better insight into how our approach performs in various settings. 
We first examine its convergence rate~(Sec.~\ref{sec:convergence_rate}), then we present an ablation study on our loss components~(Sec.~\ref{sec:loss_ablation}). 
We investigate the influence of parameters~$\lambda_{s}$ and~$t$~(Sec.~\ref{sec:lambda_t}) and conditioning prior~$\mathbf{c}$ (Sec.~\ref{sec:conditioning}). 
Finally, we compare the results of training with $\mathcal{L}_\text{structure}$ and without it using only the line art guidance function $c(\mathbf{y})$ (Sec.~\ref{sec:difflineart}).

\subsection{Convergence rate}\label{sec:convergence_rate}

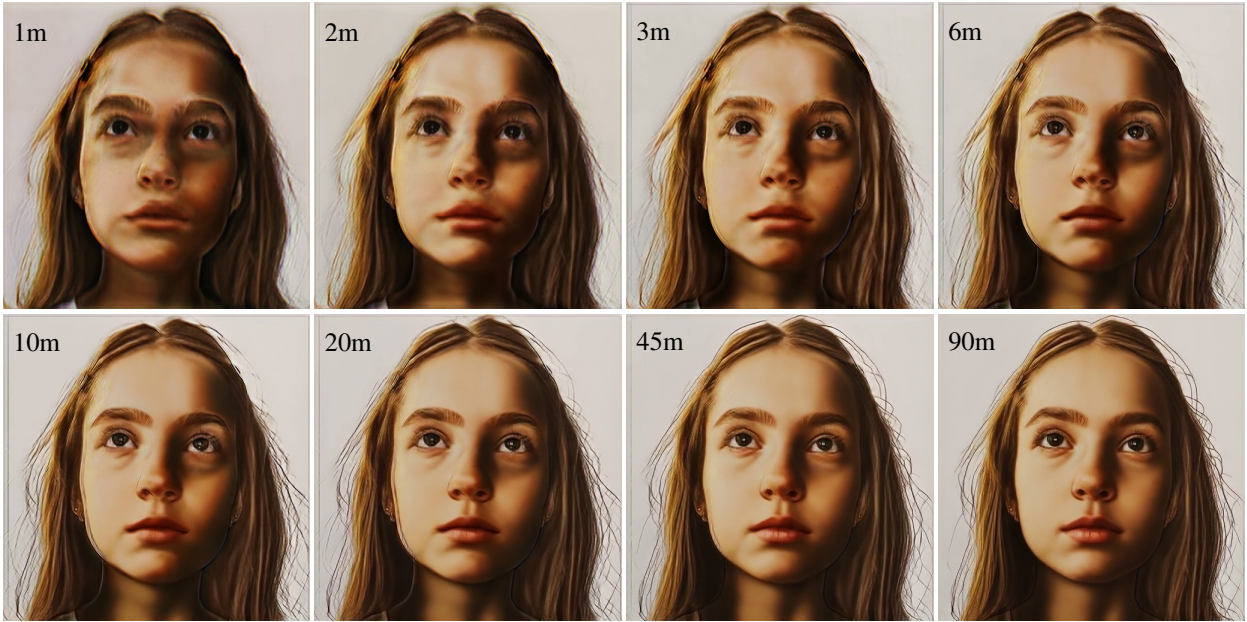
\begin{figure}
\def\svgwidth{\hsize}\import{figures/timelapse/}{timelapse.tex}\caption{Convergence speed of our method. Results captured at~$1$, $2$, $3$, $6$, $10$, $20$, $45$, and~$90$ minutes of training reveal the progressive improvement of the stylization. Reasonable results are retrieved after~$6$ minutes of training, with no notable improvement beyond $90$ minutes on NVIDIA A100 GPU.}\label{fig:timelapse}
\end{figure}

In the first experiment, we explore the optimization convergence speed of our method. The results presented in Fig.~\ref{fig:timelapse} indicate that a reasonably stylized output could be obtained after~$6$ minutes of training, and training for more than~$90$ minutes does not bring a significant improvement in stylization quality. Once the network has been trained, it is capable of performing a real-time stylization of a live video stream. This gives our method a key advantage over the diffusion-based video-to-video methods~\cite{ceylan_pix2video_2023,chu_medm_2024,geyer_tokenflow_2024,yang_rerender_2023}, which require at least~$5$ seconds of computation per frame on a single NVIDIA A100 GPU. 

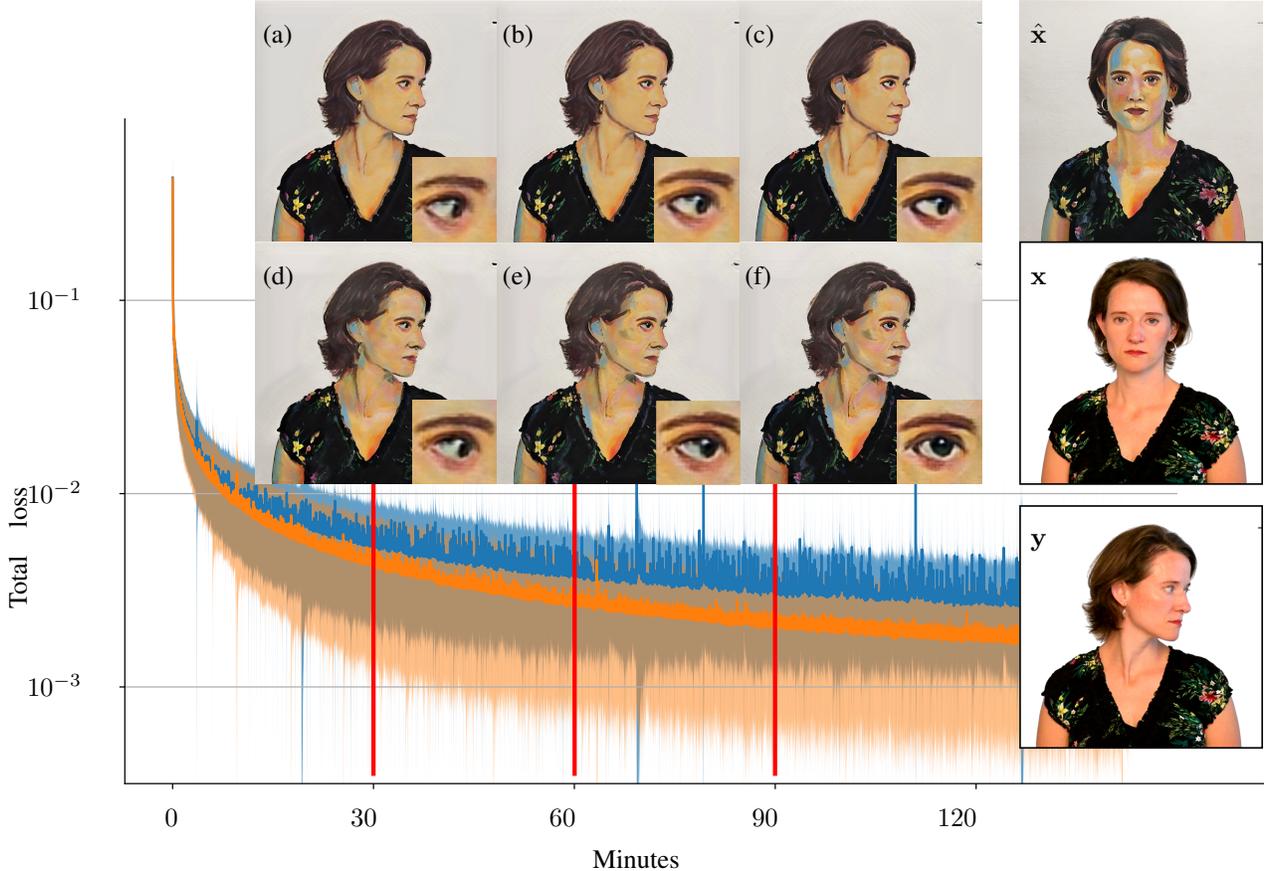
\begin{figure}
\def\svgwidth{\hsize}\import{figures/loss_stability/}{loss_stability.tex}\caption{Comparative analysis of stylization results over training duration. In each of the first three columns, a stylized target frame~$\mathbf{y}$ is shown after~$30$, $60$, and~$90$ minutes of training. Top row~(a), (b), (c) and blue color: our method, bottom row~(d), (e), (f) and orange color: Futschik et al.~\shortcite{futschik_stalp_2021}. The method of Futschik et al.~\shortcite{futschik_stalp_2021} prioritizes the preservation of the original style. However, over time, this leads to a gradual loss of fine details in the structure of the target sequence (cf.~zoom-in insets). A key advantage of our method is that we explicitly strive to retain these details, thereby maintaining their fidelity. The plots of the total loss are averaged over~$15$ different sequences, the border of the opaque area depicts the measured standard deviation.}\label{fig:loss_stability}
\end{figure}

In~Fig.~\ref{fig:loss_stability}, we compare stylization results of our method (Figures~\ref{fig:loss_stability}a--c, blue color plot) with Futschik et al.~\shortcite{futschik_stalp_2021} (Figures~\ref{fig:loss_stability}d--f, orange color plot) after~$30$, $60$, and~$90$ minutes of training (plots are averages over~$15$ different sequences; the border of the opaque area is the standard deviation).
The approach of~Futschik et al.~\shortcite{futschik_stalp_2021} emphasizes the preservation of the original style. 
However, this emphasis results in a gradual loss of fine details within the structure of the target sequence, as illustrated by the zoom-in insets in~Fig.~\ref{fig:loss_stability}. Note how the eye in~Fig.~\ref{fig:loss_stability}f is directly copied from the stylized keyframe~$\mathbf{y}_k$.
A key advantage of our approach is that we explicitly retain these fine structural details, thus maintaining fidelity to the target frames~$\mathbf{y}$.

\subsection{Ablation study on our loss components}\label{sec:loss_ablation}

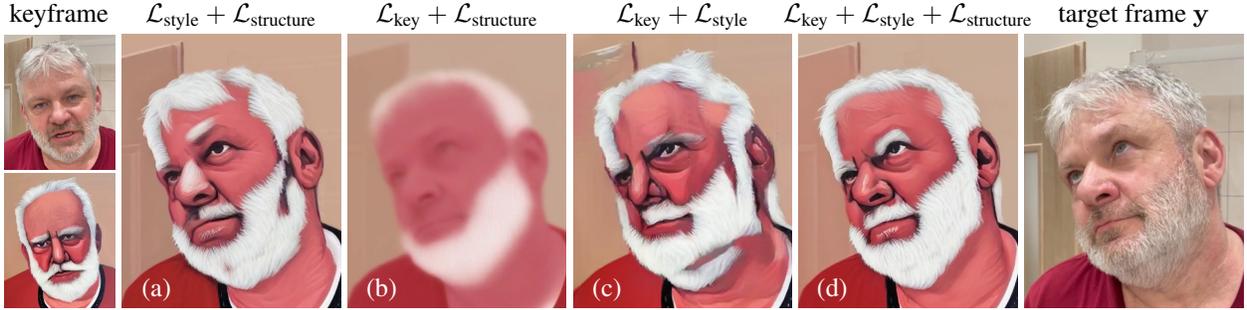
\begin{figure*}
\def\svgwidth{\hsize}\import{figures/loss_ablation/}{loss_ablation.tex}\caption{An ablation study demonstrating the importance of individual terms in our objective function (\ref{eq:loss}). A neural network is trained to transfer the style from the stylized keyframe~$\mathbf{y}$ to the target frame~$\mathbf{y}$. Each of the three terms~$\mathcal{L}_{\text{key}}$ (\ref{eq:loss_key}), $\mathcal{L}_{\text{style}}$ (\ref{eq:loss_vgg}), and~$\mathcal{L}_{\text{structure}}$ (\ref{eq:loss_sds}) is left out in the training. Leaving out~$\mathcal{L}_{\text{key}}$~(a) causes the style transfer to be less semantically meaningful  (see, e.g., white nose), excluding~$\mathcal{L}_{\text{style}}$~(b) leads to complete style washout, and leaving out~$\mathcal{L}_{\text{structure}}$~(c) results in poor replication of target frame structures. Only the combination of all three terms yield satisfactory results~(d).}\label{fig:loss_ablation}
\end{figure*}

In this experiment, we performed an ablation study to analyze the impact of individual loss terms within the optimization of our model. We systematically set multiplication constant of each of the three loss terms~$\lambda_k$, $\lambda_v$, and~$\lambda_s$~(\ref{eq:loss}) to zero. The results are presented in Fig.~\ref{fig:loss_ablation} for the model with the lowest total loss after training on a GPU for~$4$ hours. We specifically selected an input frame~$\mathbf{x}$ with a significant appearance change with respect to the style exemplar~$\mathbf{y}$ to highlight the ability of our method to produce results in the style of the exemplar even when new, previously unseen structures appear in the input frame. The~$t=28$ and~$\lambda_{\text{structure}} = 5 \cdot 10^{-6}$.

Excluding~$\mathcal{L}_{\text{key}}$ from the loss in~Fig.~\ref{fig:loss_ablation}a results in an output that preserves structure of the target frame~$\mathbf{y}$, however, the transfer is not semantically meaningful, i.e., colors from the stylized keyframe~$\mathbf{y}$ are placed to improper locations in the target frame~$\mathbf{y}$ (cf.~white spots located on the nose and forehead). 


When the term~$\mathcal{L}_{\text{style}}$ is omitted in~Fig.~\ref{fig:loss_ablation}b, we get a result with an overall structure of the target frame~$\mathbf{y}$ but with corrupted style details. This is caused by the fact that the term~$\mathcal{L}_{\text{structure}}$ enforces the structure of the target frame~$\mathbf{y}$ but not details of the stylized keyframe~$\mathbf{y}$ that are maintained by the term~$\mathcal{L}_{\text{style}}$. 

When~$\mathcal{L}_{\text{structure}}$ is omitted in~Fig.~\ref{fig:loss_ablation}c, the term~$\mathcal{L}_{\text{style}}$ enforces the presence of the style details in the keyframe image~$\mathbf{y}$, and the term~$\mathcal{L}_{\text{key}}$ keeps the style transfer to be semantically meaningful. However, when previously unseen structures appear, such as new shapes of eyes, they are poorly reconstructed since the loss without term~$\mathcal{L}_{\text{structure}}$ forces the network to reproduce only structures appearing in the stylized keyframe~$\mathbf{y}$. 


Only by combining the three loss terms together in~Fig.~\ref{fig:loss_ablation}d a result is produced that preserves the structural integrity of the target image~$\mathbf{x}$ as well as incorporates the style features of the keyframe~$\mathbf{y}$.

\subsection{The influence of parameters~$\lambda_{s}$ and~$t$}\label{sec:lambda_t}

In this experiment, we manipulate the parameter~$t$ in~$\mathcal{L}_{\text{structure}}$~(\ref{eq:loss_sds}) and the loss weight~$\lambda_s$ (\ref{eq:loss}). We trained the network for~$4$ hours on a GPU and selected the network with the lowest total loss.

\begin{figure}
\def\svgwidth{\hsize}\import{figures/style_as_function_of_sds_attributes_3/}{style_as_function_of_sds_attributes_3.tex}\caption{Reinforcement of input video structure as function of parameters~$\lambda_s$~(\ref{eq:loss}) and~$t$~(\ref{eq:loss_sds}). The stronger~$\lambda_s$, the more pronounced the structure of~$\mathbf{x}$ is in the result. Note that the ControlNet used by the structural guidance term~$\mathcal{L}_{\text{structure}}$ (\ref{eq:loss_sds}) uses a~$30$ step denoising schedule. At~$t = 0$, the guidance input is pure Gaussian noise~$\epsilon \sim \mathcal{N}(\mathbf{0}, \mathbf{I})$, and at~$t = 29$, it contains minimal noise.}\label{fig:style_as_function_of_sds_attributes_3}
\end{figure}
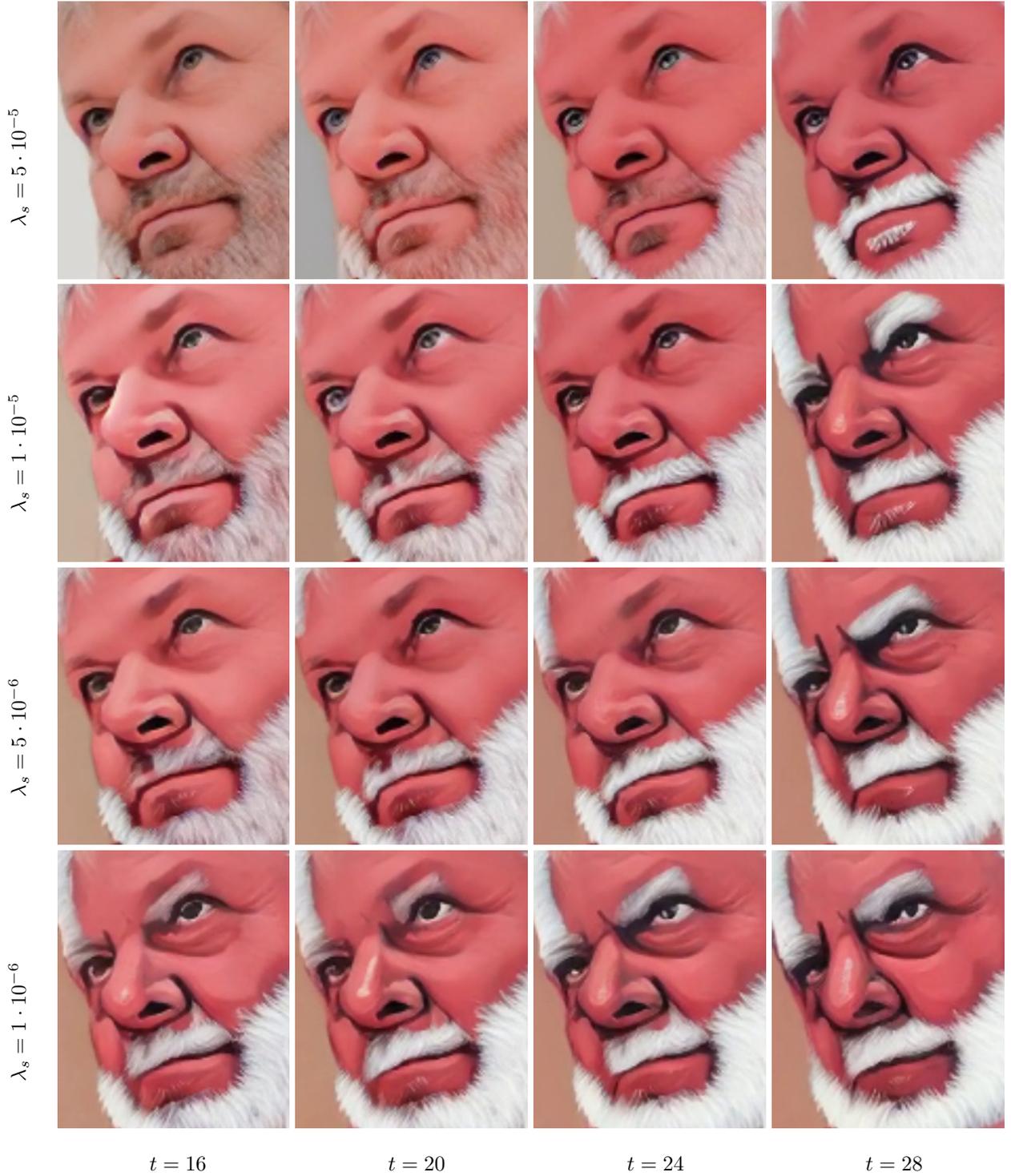


The results in~Fig.~\ref{fig:style_as_function_of_sds_attributes_3} show that the stronger~$\lambda_s$, the more pronounced the structure of~$\mathbf{x}$ is in the result. Since the value of~$\lambda_s$ modifies the strength of~$\mathcal{L}_{\text{structure}}$ in the total loss term~$\mathcal{L}$~(\ref{eq:loss}), setting it to a higher value results in a stronger pronunciation of the input video structures in the output of our method. Another interesting dimension of structure reinforcement control strength is the parameter~$t$. The ControlNet we utilize for the purpose of structure guidance uses a~$30$ step denoising schedule. Every time the loss term~$\mathcal{L}_{\text{structure}}$ is evaluated, a convex combination~$\hat{\mathbf{y}}_{i, t}$~(\ref{eq:loss_sds}) of the output of our method and a Gaussian noise~$\epsilon \sim \mathcal{N}(\mathbf{0},\mathbf{I})$ is performed. Then, the pre-trained ControlNet estimates the original noise~$\epsilon$, and the residual between the estimated noise and~$\epsilon$ is propagated through our method. This means that when~$t=0$, ControlNet gets a pure Gaussian noise and when~$t=29$, it gets the output of our network with only a little Gaussian noise present. According to this, we see that the lower the value of~$t$, the stronger the focus is on the high-level structures compared to the higher values of~$t$.

\subsection{The influence of image conditioning prior~$\mathbf{c}$}\label{sec:conditioning}

In our method, we use a ControlNet diffusion network~\cite{zhang_adding_2023} to ensure that the structural elements of the target frame~$\mathbf{y}$ are accurately reflected in the stylized output. 
We hypothesize that by sampling from the ControlNet network, we leverage both the structural conditioning provided by the guidance image~$\mathbf{c}$ and the network's inherent ability to guide denoised samples toward the learned manifold of real images.

In this experiment, we evaluate the effectiveness of four pre-trained ControlNet models for providing structural guidance in stylization. Each ControlNet uses a different image guidance function~$c$: (i)~``line art detection'' neural model, (ii)~``depth estimation'' neural model, (iii)~Canny edge detector~\cite{canny_computational_1986}, and~(iv)~``soft edge estimation`` neural model.

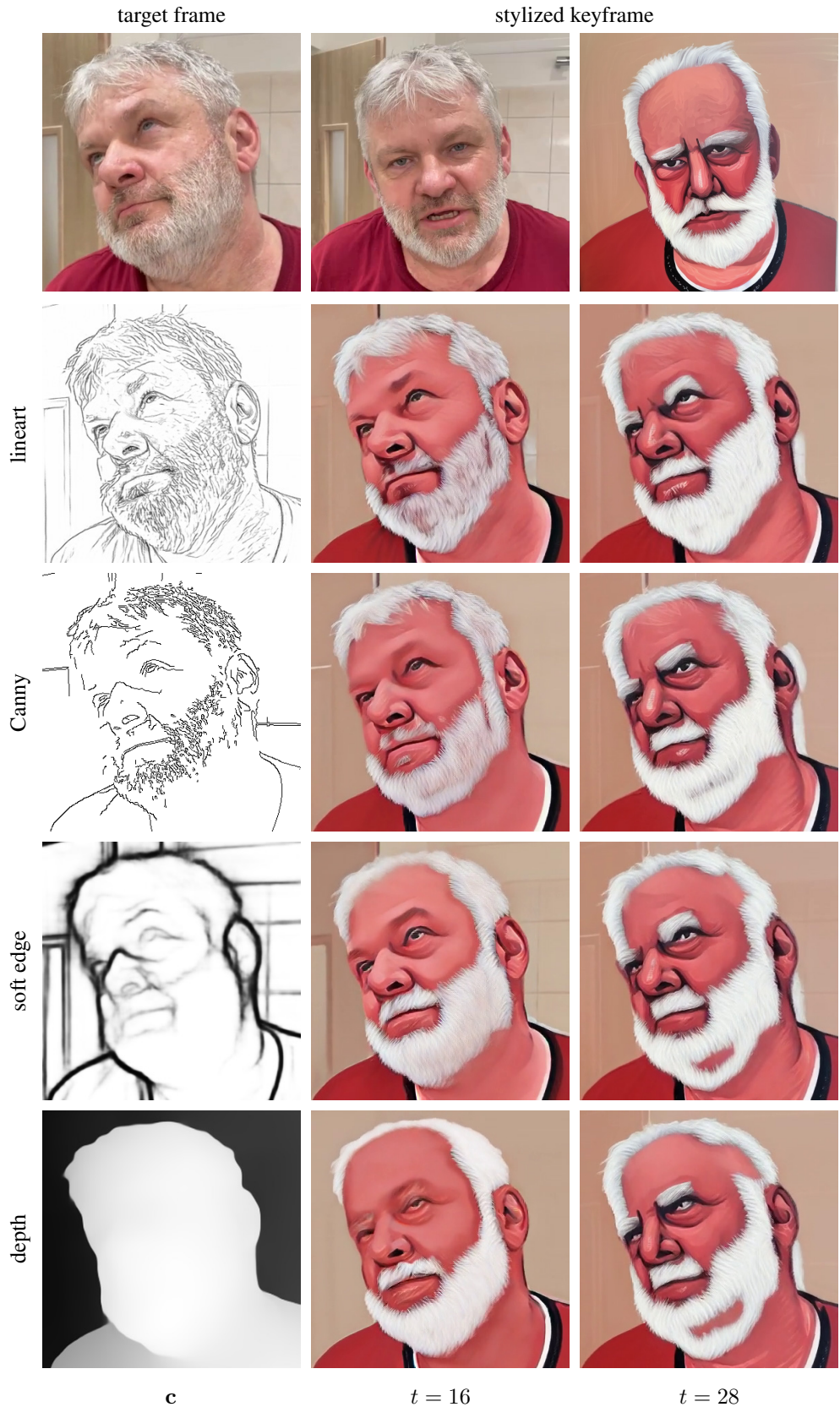
\begin{figure}
\def\svgwidth{\hsize}\import{figures/control_type_2/}{control_type_2.tex}\caption{Structure reinforcement as function of image guidance~$\mathbf{c}$ and diffusion time steps~$t$. The best results are achieved with the line art guidance within ControlNet's~$\mathcal{L}_{\text{structure}}$ (\ref{eq:loss_sds}) at~$t=28$.}\label{fig:control_type_2}
\end{figure}

The results for~$t = 16$ and~$t = 28$ are presented in~Fig.~\ref{fig:control_type_2}. Both depth and soft edge conditioning result in stylization that lacks the finer details of the target frame~$\mathbf{y}$, likely due to the coarse structure of their respective guidance images~$\mathbf{c}$. The Canny edge detector provides more detailed signal to the ControlNet, while line art guidance offers the most precise structural information. For applications requiring rapid training, the quality of line art can be sacrificed for the approximately~$10$-fold faster Canny edge guidance. In contrast, depth guidance is less practical, taking about~$10$ times longer than line art, with soft edge guidance falling between the two in terms of computational cost.

In this experiment, we kept~$\lambda_s = 5 \cdot 10^{-6}$ and trained on a single NVIDIA A100 GPU for~$4$ hours. We selected the model with the lowest total loss.

\subsection{Direct optimization with the image conditioning prior}\label{sec:difflineart}

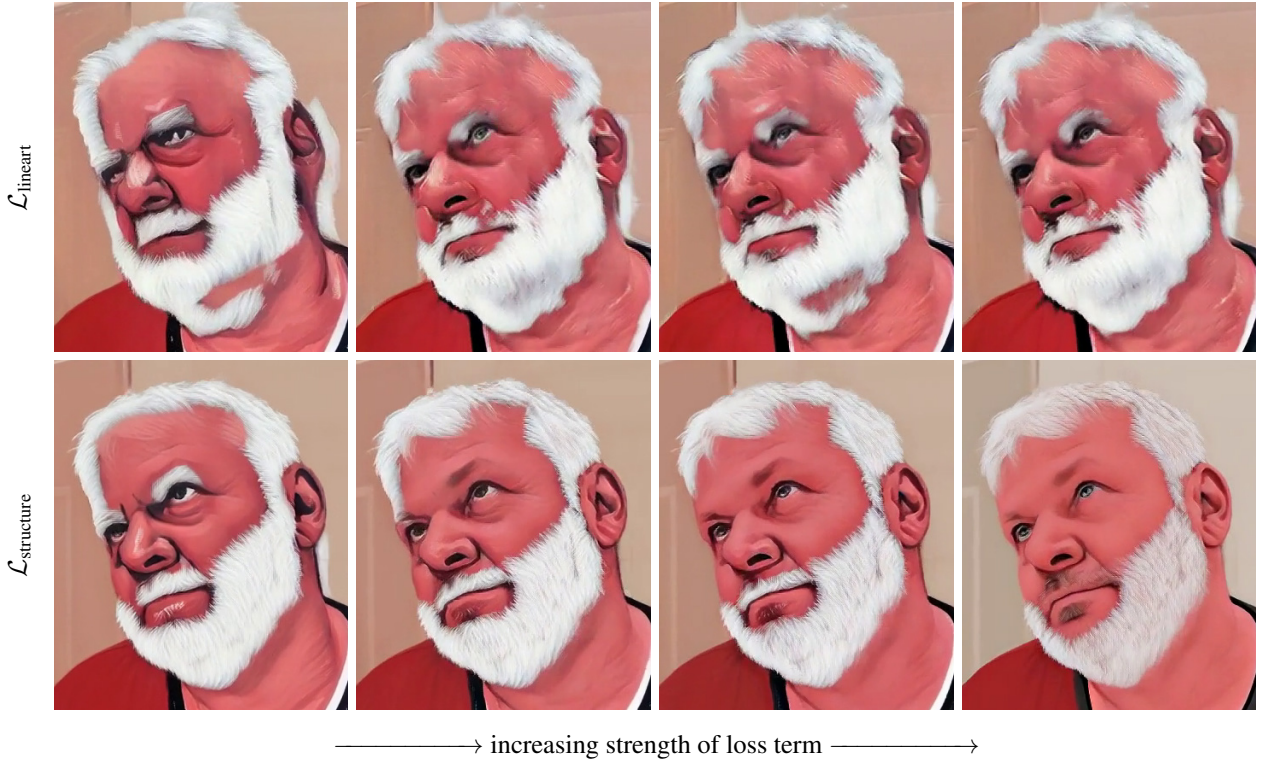
\begin{figure}
\def\svgwidth{\hsize}\import{figures/difflineart/}{difflineart.tex}\caption{Comparison of stylization results using the line art loss term~$\mathcal{L}_\text{lineart}$~(\ref{eq:loss_lineart}) versus our line art-guided structure loss term~$\mathcal{L}_{\text{structure}}$~(\ref{eq:loss_sds}). Increasing the strength of~$\mathcal{L}_\text{lineart}$ fails to effectively transfer structural elements from the target frame to the stylized output, whereas~$\mathcal{L}_{\text{structure}}$ successfully preserves the target's structural details.}\label{fig:difflineart}
\end{figure}

In our experiments, we show that our proposed loss term~$\mathcal{L}_\text{structure}$~(\ref{eq:loss_sds}) enforces structures from the target frame~$\mathbf{y}$ to the stylized output~$\hat{\mathbf{y}}$ of our method. The purpose of this experiment is two fold: first, to demonstrate that the~$\mathcal{L}_\text{structure}$ loss based on the ControlNet model cannot be replaced by a simpler loss that uses the line art image conditioning model~$c(\mathbf{y})$, which is already integrated into ControlNet as the conditioning~$c$; and second, to present this simpler loss as an alternative structure-preserving mechanism, serving as a baseline for comparison.
In Fig.~\ref{fig:difflineart}, we show that replacing our proposed line art-guided structure loss~$\mathcal{L}_\text{structure}$ with the line art term
\begin{equation}\label{eq:loss_lineart}
    \mathcal{L}_\text{lineart}(\mathcal{Y}) = \frac{1}{|\mathcal{Y}|} \sum_{i \not\in \mathbf{K}} \left\| c\left(\hat{\mathbf{y}}_i\right) - c\left(\mathbf{y}_i\right) \right\|_2^2
\end{equation}
 fails to enforce the transfer of structural details from the target frames~$\mathbf{y}_i$ to the stylized outputs~$f(\mathbf{y}_{i})=\hat{\mathbf{y}}_i$. In this experiment, we trained on a single NVIDIA A100 GPU for~$4$ hours and selected the model with the lowest total loss.



\section{Limitations}

Although our approach represents an improvement over the current state-of-the-art in both diffusion-based and keyframe-based video stylization methods, we have identified certain limitations in its application.

In style transfer methods, there exists a delicate balance between maintaining fidelity to the features of the style exemplar images and preserving the structural characteristics present in the content that is being stylized. 
The current state-of-the-art in keyframe-based video stylization mainly emphasizes fidelity to the original features in the style image. 
Our approach aims to produce stylized content that aligns both with both the style exemplar and the structural characteristics of the target video sequence. Although we enable users to find the right balance between style and structure using parameters~$\lambda_s$ and~$t$, we acknowledge that in some situations, decreasing the fidelity to the style features may be perceived as a~potential limitation.

Our method achieves training times comparable to the approach of Futshik et al.~\shortcite{futschik_stalp_2021} which is relatively long and can limit the possibility for interactive updates available in methods like that of Texler et al.~\shortcite{texler_interactive_2020}. To address this drawback, in future work we plan to combine Texler et al.'s rapid patch-based training strategy with the calculation of the \emph{style} loss in a full-frame setting.

In our proposed workflow, the user is expected to select a keyframe that will be used for training. Although certain guidelines can be followed, a mechanism that enables automatic selection of suitable keyframes could simplify the preparation phase.

\section{Conclusion}

In this work, we introduced a novel keyframe-based video stylization method that balances the preservation of essential structural elements with adherence to a prescribed visual style.
By integrating the line art-guided structure loss term~$\mathcal{L}_{\text{structure}}$, our approach overcomes limitations of existing diffusion-based and keyframe-based stylization techniques by preserving structural detail from the input video, enhancing the quality and coherence of stylized sequences, and reducing the need for additional correction keyframes.
Real-time inference of our method enables interactive applications such as consistently stylized video calls, which are challenging with existing approaches. 
By blending style fidelity with structural preservation, our method significantly advances video stylization in both quality and usability.

%

\include{supp/supplementary.tex}

\input{stylereiser.bbl}

\end{document}

%% file: figures/teaser/teaser.tex
\begingroup%
  \makeatletter%
  \providecommand\color[2][]{%
    \errmessage{(Inkscape) Color is used for the text in Inkscape, but the package 'color.sty' is not loaded}%
    \renewcommand\color[2][]{}%
  }%
  \providecommand\transparent[1]{%
    \errmessage{(Inkscape) Transparency is used (non-zero) for the text in Inkscape, but the package 'transparent.sty' is not loaded}%
    \renewcommand\transparent[1]{}%
  }%
  \providecommand\rotatebox[2]{#2}%
  \newcommand*\fsize{\dimexpr\f@size pt\relax}%
  \newcommand*\lineheight[1]{\fontsize{\fsize}{#1\fsize}\selectfont}%
  \ifx\svgwidth\undefined%
    \setlength{\unitlength}{1237.23174328bp}%
    \ifx\svgscale\undefined%
      \relax%
    \else%
      \setlength{\unitlength}{\unitlength * \real{\svgscale}}%
    \fi%
  \else%
    \setlength{\unitlength}{\svgwidth}%
  \fi%
  \global\let\svgwidth\undefined%
  \global\let\svgscale\undefined%
  \makeatother%
  \begin{picture}(1,0.2985666)%
    \lineheight{1}%
    \setlength\tabcolsep{0pt}%
    \put(0,0){\includegraphics[width=\unitlength,page=1]{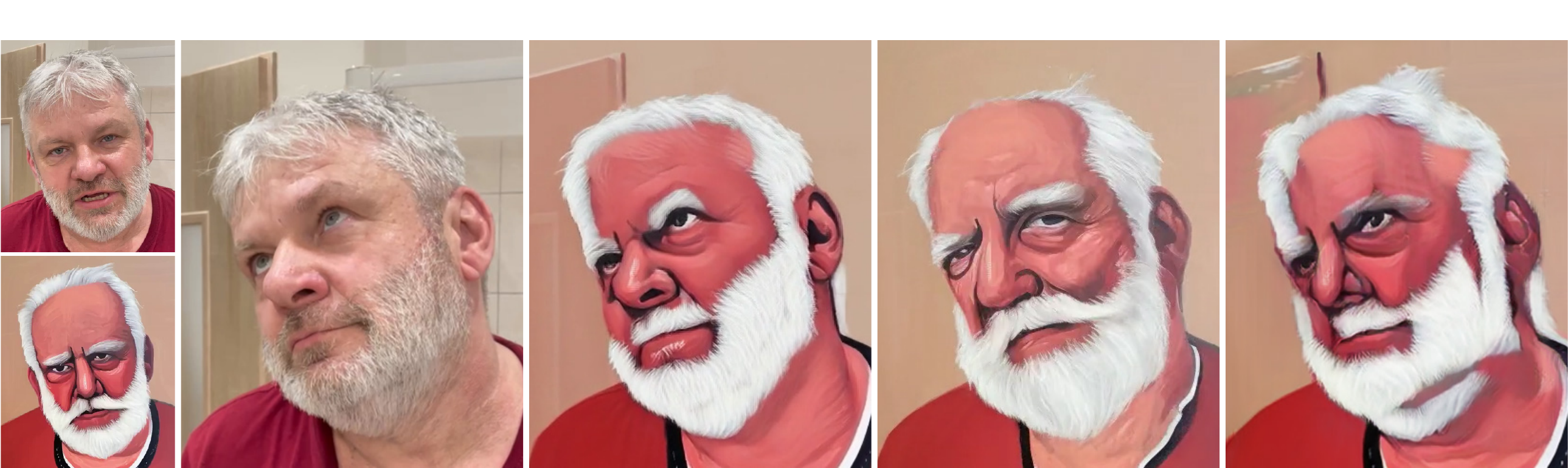}}%
    \put(0.05547471,0.285){\color[rgb]{0,0,0}\makebox(0,0)[t]{\lineheight{1.25}\smash{\begin{tabular}[t]{c}(a)~keyframe\end{tabular}}}}%
    \put(0.22519270,0.285){\color[rgb]{0,0,0}\makebox(0,0)[t]{\lineheight{1.25}\smash{\begin{tabular}[t]{c}(b)~target frame\end{tabular}}}}%
    \put(0.44706267,0.285){\color[rgb]{0,0,0}\makebox(0,0)[t]{\lineheight{1.25}\smash{\begin{tabular}[t]{c}(c)~our approach\end{tabular}}}}%
    \put(0.66823410,0.285){\color[rgb]{0,0,0}\makebox(0,0)[t]{\lineheight{1.25}\smash{\begin{tabular}[t]{c}(d)~Ceylan et al.~\shortcite{ceylan_pix2video_2023}\end{tabular}}}}%
    \put(0.89045331,0.285){\color[rgb]{0,0,0}\makebox(0,0)[t]{\lineheight{1.25}\smash{\begin{tabular}[t]{c}(e)~Futschik et al.~\shortcite{futschik_stalp_2021}\end{tabular}}}}%
  \end{picture}%
\endgroup%

%% file: figures/overview/overview.tex
\begingroup%
  \makeatletter%
  \providecommand\color[2][]{%
    \errmessage{(Inkscape) Color is used for the text in Inkscape, but the package 'color.sty' is not loaded}%
    \renewcommand\color[2][]{}%
  }%
  \providecommand\transparent[1]{%
    \errmessage{(Inkscape) Transparency is used (non-zero) for the text in Inkscape, but the package 'transparent.sty' is not loaded}%
    \renewcommand\transparent[1]{}%
  }%
  \providecommand\rotatebox[2]{#2}%
  \newcommand*\fsize{\dimexpr\f@size pt\relax}%
  \newcommand*\lineheight[1]{\fontsize{\fsize}{#1\fsize}\selectfont}%
  \ifx\svgwidth\undefined%
    \setlength{\unitlength}{514.04207485bp}%
    \ifx\svgscale\undefined%
      \relax%
    \else%
      \setlength{\unitlength}{\unitlength * \real{\svgscale}}%
    \fi%
  \else%
    \setlength{\unitlength}{\svgwidth}%
  \fi%
  \global\let\svgwidth\undefined%
  \global\let\svgscale\undefined%
  \makeatother%
  \begin{picture}(1,0.36712617)%
    \lineheight{1}%
    \setlength\tabcolsep{0pt}%
    \put(0,0){\includegraphics[width=\unitlength,page=1]{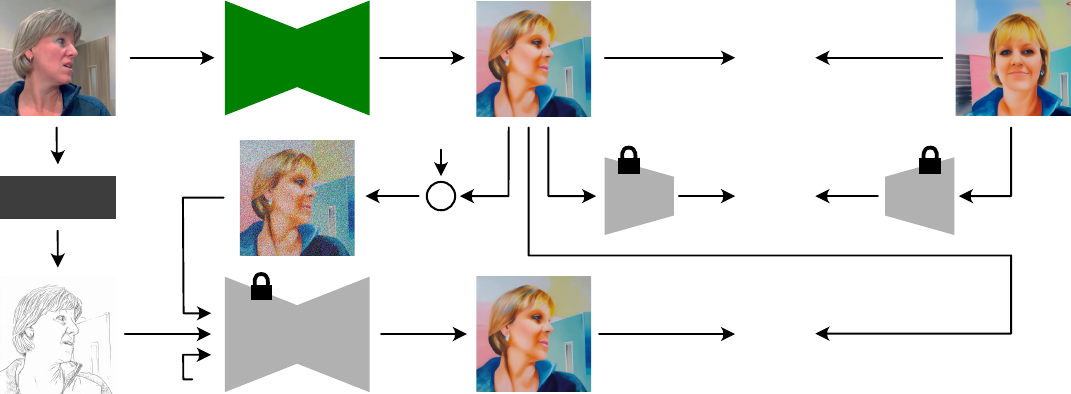}}\Large
    \put(0.27576450,0.30679183){\makebox(0,0)[t]{\lineheight{1.25}\smash{\color[rgb]{1,1,1}$f$}}}%
    \put(0.72580945,0.08079404){\makebox(0,0)[t]{\lineheight{1.25}\smash{$\mathcal{L}_{\text{structure}}$}}}%
    \put(0.72580945,0.20958123){\makebox(0,0)[t]{\lineheight{1.25}\smash{$\mathcal{L}_{\text{style}}$}}}%
    \put(0.72580945,0.33879183){\makebox(0,0)[t]{\lineheight{1.25}\smash{$\mathcal{L}_{\text{key}}$}}}%
    \put(0.72580945,0.04879404){\makebox(0,0)[t]{\lineheight{1.25}\smash{$\left\|\ldots\right\|^2_2$}}}%
    \put(0.72580945,0.30679183){\makebox(0,0)[t]{\lineheight{1.25}\smash{$\left\|\ldots\right\|^2_2$}}}%
    \put(0.72580945,0.17808123){\makebox(0,0)[t]{\lineheight{1.25}\smash{$\left\|\ldots\right\|^2_2$}}}%
    \put(0.72551520,0.27209824){\makebox(0,0)[t]{\lineheight{1.25}\smash{$\Leftrightarrow{i}\in\mathbf{K}$}}}%
    \put(0.85817986,0.17808123){\makebox(0,0)[t]{\lineheight{1.25}\smash{$g(v)$}}}%
    \put(0.59554529,0.17808123){\makebox(0,0)[t]{\lineheight{1.25}\smash{$g(v)$}}}%
    \put(0.27576450,0.04879404){\makebox(0,0)[t]{\lineheight{1.25}\smash{$d$}}}%
    \put(0.18574702,0.00577530){\makebox(0,0)[t]{\lineheight{1.25}\smash{$t$}}}%
    \put(0.05453563,0.17808123){\makebox(0,0)[t]{\lineheight{1.25}\smash{\color[rgb]{1,1,1}$c$}}}%
    \put(0.41156893,0.23423699){\makebox(0,0)[t]{\lineheight{1.25}\smash{$\mathbf{\epsilon}$}}}%
    \put(0.41156893,0.17808123){\makebox(0,0)[t]{\lineheight{1.25}\smash{$+$}}}%
    \put(0.092,0.347){\makebox(0,0)[t]{\lineheight{1.25}\smash{\color[rgb]{1,1,1}$\mathbf{y}_i$}}}%
    \put(0.537,0.347){\makebox(0,0)[t]{\lineheight{1.25}\smash{\color[rgb]{1,1,1}$\hat{\mathbf{y}}_i$}}}%
    \put(0.313,0.215){\makebox(0,0)[t]{\lineheight{1.25}\smash{\color[rgb]{1,1,1}$\hat{\mathbf{y}}_{i,t}$}}}%
    \put(0.985,0.347){\makebox(0,0)[t]{\lineheight{1.25}\smash{\color[rgb]{1,1,1}$\hat{\mathbf{y}}_k$}}}%
    \put(0.095,0.090){\makebox(0,0)[t]{\lineheight{1.25}\smash{$\mathbf{c}$}}}%
  \end{picture}%
\endgroup%

%% file: figures/sota_comparison_dad/sota_comparison_dad.tex
\begingroup%
  \makeatletter%
  \providecommand\color[2][]{%
    \errmessage{(Inkscape) Color is used for the text in Inkscape, but the package 'color.sty' is not loaded}%
    \renewcommand\color[2][]{}%
  }%
  \providecommand\transparent[1]{%
    \errmessage{(Inkscape) Transparency is used (non-zero) for the text in Inkscape, but the package 'transparent.sty' is not loaded}%
    \renewcommand\transparent[1]{}%
  }%
  \providecommand\rotatebox[2]{#2}%
  \newcommand*\fsize{\dimexpr\f@size pt\relax}%
  \newcommand*\lineheight[1]{\fontsize{\fsize}{#1\fsize}\selectfont}%
  \ifx\svgwidth\undefined%
    \setlength{\unitlength}{373.57923132bp}%
    \ifx\svgscale\undefined%
      \relax%
    \else%
      \setlength{\unitlength}{\unitlength * \real{\svgscale}}%
    \fi%
  \else%
    \setlength{\unitlength}{\svgwidth}%
  \fi%
  \global\let\svgwidth\undefined%
  \global\let\svgscale\undefined%
  \makeatother%
  \begin{picture}(1,0.41499197)%
    \lineheight{1}%
    \setlength\tabcolsep{0pt}%
    \put(0,0){\includegraphics[width=\unitlength,page=1]{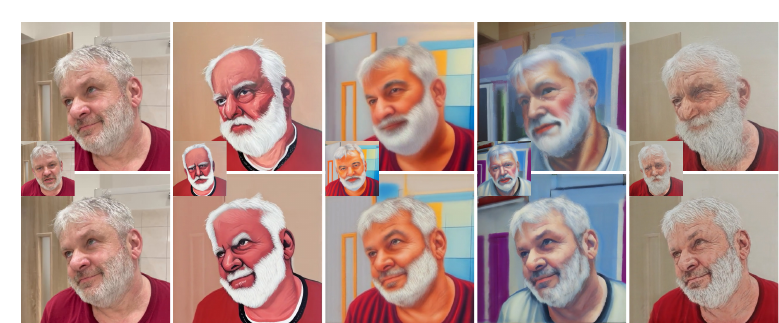}}%
    \put(0.12285387,0.39465448){\color[rgb]{0,0,0}\makebox(0,0)[t]{\lineheight{1.25}\smash{\begin{tabular}[t]{c}\targetframefigure{}\end{tabular}}}}%
    \put(0.31869846,0.39465448){\color[rgb]{0,0,0}\makebox(0,0)[t]{\lineheight{1.25}\smash{\begin{tabular}[t]{c}(a)~Ceylan et al.~\shortcite{ceylan_pix2video_2023}\end{tabular}}}}%
    \put(0.51308566,0.39465448){\color[rgb]{0,0,0}\makebox(0,0)[t]{\lineheight{1.25}\smash{\begin{tabular}[t]{c}(b)~Yang et al.~\shortcite{yang_rerender_2023}\end{tabular}}}}%
    \put(0.70863618,0.39465448){\color[rgb]{0,0,0}\makebox(0,0)[t]{\lineheight{1.25}\smash{\begin{tabular}[t]{c}(c)~Chu et al.~\shortcite{chu_medm_2024}\end{tabular}}}}%
    \put(0.90482707,0.39465448){\color[rgb]{0,0,0}\makebox(0,0)[t]{\lineheight{1.25}\smash{\begin{tabular}[t]{c}(d)~Geyer et al.~\shortcite{geyer_tokenflow_2024}\end{tabular}}}}%
    \put(0.02046819,0.29129200){\color[rgb]{0,0,0}\rotatebox{90}{\makebox(0,0)[t]{\lineheight{1.25}\smash{\begin{tabular}[t]{c}target frame\end{tabular}}}}}%
    \put(0.02046819,0.09614012){\color[rgb]{0,0,0}\rotatebox{90}{\makebox(0,0)[t]{\lineheight{1.25}\smash{\begin{tabular}[t]{c}our approach\end{tabular}}}}}%
  \end{picture}%
\endgroup%

%% file: figures/sota_comparison_lili/sota_comparison_lili.tex
\begingroup%
  \makeatletter%
  \providecommand\color[2][]{%
    \errmessage{(Inkscape) Color is used for the text in Inkscape, but the package 'color.sty' is not loaded}%
    \renewcommand\color[2][]{}%
  }%
  \providecommand\transparent[1]{%
    \errmessage{(Inkscape) Transparency is used (non-zero) for the text in Inkscape, but the package 'transparent.sty' is not loaded}%
    \renewcommand\transparent[1]{}%
  }%
  \providecommand\rotatebox[2]{#2}%
  \newcommand*\fsize{\dimexpr\f@size pt\relax}%
  \newcommand*\lineheight[1]{\fontsize{\fsize}{#1\fsize}\selectfont}%
  \ifx\svgwidth\undefined%
    \setlength{\unitlength}{373.57923132bp}%
    \ifx\svgscale\undefined%
      \relax%
    \else%
      \setlength{\unitlength}{\unitlength * \real{\svgscale}}%
    \fi%
  \else%
    \setlength{\unitlength}{\svgwidth}%
  \fi%
  \global\let\svgwidth\undefined%
  \global\let\svgscale\undefined%
  \makeatother%
  \begin{picture}(1,0.41499197)%
    \lineheight{1}%
    \setlength\tabcolsep{0pt}%
    \put(0,0){\includegraphics[width=\unitlength,page=1]{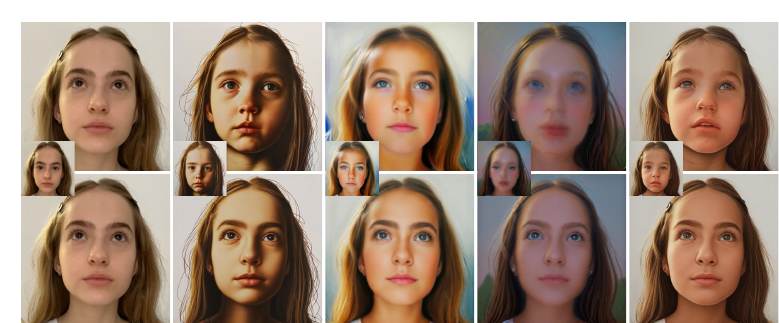}}%
    \put(0.12285387,0.39465448){\color[rgb]{0,0,0}\makebox(0,0)[t]{\lineheight{1.25}\smash{\begin{tabular}[t]{c}\targetframefigure{}\end{tabular}}}}%
    \put(0.31869846,0.39465448){\color[rgb]{0,0,0}\makebox(0,0)[t]{\lineheight{1.25}\smash{\begin{tabular}[t]{c}(a)~Ceylan et al.~\shortcite{ceylan_pix2video_2023}\end{tabular}}}}%
    \put(0.51308566,0.39465448){\color[rgb]{0,0,0}\makebox(0,0)[t]{\lineheight{1.25}\smash{\begin{tabular}[t]{c}(b)~Yang et al.~\shortcite{yang_rerender_2023}\end{tabular}}}}%
    \put(0.70863618,0.39465448){\color[rgb]{0,0,0}\makebox(0,0)[t]{\lineheight{1.25}\smash{\begin{tabular}[t]{c}(c)~Chu et al.~\shortcite{chu_medm_2024}\end{tabular}}}}%
    \put(0.90482707,0.39465448){\color[rgb]{0,0,0}\makebox(0,0)[t]{\lineheight{1.25}\smash{\begin{tabular}[t]{c}(d)~Geyer et al.~\shortcite{geyer_tokenflow_2024}\end{tabular}}}}%
    \put(0.02046819,0.29129200){\color[rgb]{0,0,0}\rotatebox{90}{\makebox(0,0)[t]{\lineheight{1.25}\smash{\begin{tabular}[t]{c}target frame\end{tabular}}}}}%
    \put(0.02046819,0.09614012){\color[rgb]{0,0,0}\rotatebox{90}{\makebox(0,0)[t]{\lineheight{1.25}\smash{\begin{tabular}[t]{c}our approach\end{tabular}}}}}%
  \end{picture}%
\endgroup%

%% file: figures/sota_comparison_zuzka/sota_comparison_zuzka.tex
\begingroup%
  \makeatletter%
  \providecommand\color[2][]{%
    \errmessage{(Inkscape) Color is used for the text in Inkscape, but the package 'color.sty' is not loaded}%
    \renewcommand\color[2][]{}%
  }%
  \providecommand\transparent[1]{%
    \errmessage{(Inkscape) Transparency is used (non-zero) for the text in Inkscape, but the package 'transparent.sty' is not loaded}%
    \renewcommand\transparent[1]{}%
  }%
  \providecommand\rotatebox[2]{#2}%
  \newcommand*\fsize{\dimexpr\f@size pt\relax}%
  \newcommand*\lineheight[1]{\fontsize{\fsize}{#1\fsize}\selectfont}%
  \ifx\svgwidth\undefined%
    \setlength{\unitlength}{373.57923132bp}%
    \ifx\svgscale\undefined%
      \relax%
    \else%
      \setlength{\unitlength}{\unitlength * \real{\svgscale}}%
    \fi%
  \else%
    \setlength{\unitlength}{\svgwidth}%
  \fi%
  \global\let\svgwidth\undefined%
  \global\let\svgscale\undefined%
  \makeatother%
  \begin{picture}(1,0.41499197)%
    \lineheight{1}%
    \setlength\tabcolsep{0pt}%
    \put(0,0){\includegraphics[width=\unitlength,page=1]{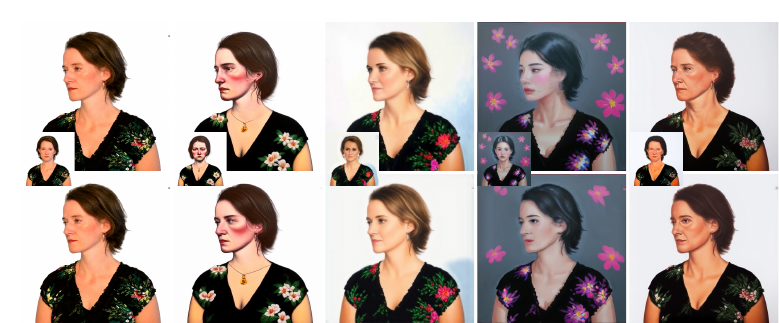}}%
    \put(0.12285387,0.39465448){\color[rgb]{0,0,0}\makebox(0,0)[t]{\lineheight{1.25}\smash{\begin{tabular}[t]{c}\targetframefigure{}\end{tabular}}}}%
    \put(0.31869846,0.39465448){\color[rgb]{0,0,0}\makebox(0,0)[t]{\lineheight{1.25}\smash{\begin{tabular}[t]{c}(a)~Ceylan et al.~\shortcite{ceylan_pix2video_2023}\end{tabular}}}}%
    \put(0.51308566,0.39465448){\color[rgb]{0,0,0}\makebox(0,0)[t]{\lineheight{1.25}\smash{\begin{tabular}[t]{c}(b)~Yang et al.~\shortcite{yang_rerender_2023}\end{tabular}}}}%
    \put(0.70863618,0.39465448){\color[rgb]{0,0,0}\makebox(0,0)[t]{\lineheight{1.25}\smash{\begin{tabular}[t]{c}(c)~Chu et al.~\shortcite{chu_medm_2024}\end{tabular}}}}%
    \put(0.90482707,0.39465448){\color[rgb]{0,0,0}\makebox(0,0)[t]{\lineheight{1.25}\smash{\begin{tabular}[t]{c}(d)~Geyer et al.~\shortcite{geyer_tokenflow_2024}\end{tabular}}}}%
    \put(0.02046819,0.29129200){\color[rgb]{0,0,0}\rotatebox{90}{\makebox(0,0)[t]{\lineheight{1.25}\smash{\begin{tabular}[t]{c}target frame\end{tabular}}}}}%
    \put(0.02046819,0.09614012){\color[rgb]{0,0,0}\rotatebox{90}{\makebox(0,0)[t]{\lineheight{1.25}\smash{\begin{tabular}[t]{c}our approach\end{tabular}}}}}%
  \end{picture}%
\endgroup%

%% file: figures/sota_comparison_joli/sota_comparison_joli.tex
\begingroup%
  \makeatletter%
  \providecommand\color[2][]{%
    \errmessage{(Inkscape) Color is used for the text in Inkscape, but the package 'color.sty' is not loaded}%
    \renewcommand\color[2][]{}%
  }%
  \providecommand\transparent[1]{%
    \errmessage{(Inkscape) Transparency is used (non-zero) for the text in Inkscape, but the package 'transparent.sty' is not loaded}%
    \renewcommand\transparent[1]{}%
  }%
  \providecommand\rotatebox[2]{#2}%
  \newcommand*\fsize{\dimexpr\f@size pt\relax}%
  \newcommand*\lineheight[1]{\fontsize{\fsize}{#1\fsize}\selectfont}%
  \ifx\svgwidth\undefined%
    \setlength{\unitlength}{373.57923132bp}%
    \ifx\svgscale\undefined%
      \relax%
    \else%
      \setlength{\unitlength}{\unitlength * \real{\svgscale}}%
    \fi%
  \else%
    \setlength{\unitlength}{\svgwidth}%
  \fi%
  \global\let\svgwidth\undefined%
  \global\let\svgscale\undefined%
  \makeatother%
  \begin{picture}(1,0.41499197)%
    \lineheight{1}%
    \setlength\tabcolsep{0pt}%
    \put(0,0){\includegraphics[width=\unitlength,page=1]{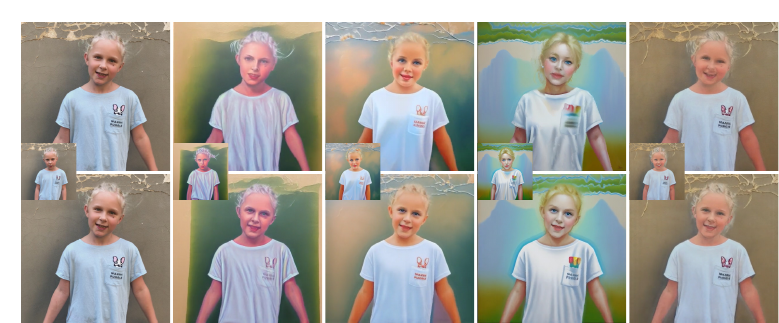}}%
    \put(0.12285387,0.39465448){\color[rgb]{0,0,0}\makebox(0,0)[t]{\lineheight{1.25}\smash{\begin{tabular}[t]{c}\targetframefigure{}\end{tabular}}}}%
    \put(0.31869846,0.39465448){\color[rgb]{0,0,0}\makebox(0,0)[t]{\lineheight{1.25}\smash{\begin{tabular}[t]{c}(a)~Ceylan et al.~\shortcite{ceylan_pix2video_2023}\end{tabular}}}}%
    \put(0.51308566,0.39465448){\color[rgb]{0,0,0}\makebox(0,0)[t]{\lineheight{1.25}\smash{\begin{tabular}[t]{c}(b)~Yang et al.~\shortcite{yang_rerender_2023}\end{tabular}}}}%
    \put(0.70863618,0.39465448){\color[rgb]{0,0,0}\makebox(0,0)[t]{\lineheight{1.25}\smash{\begin{tabular}[t]{c}(c)~Chu et al.~\shortcite{chu_medm_2024}\end{tabular}}}}%
    \put(0.90482707,0.39465448){\color[rgb]{0,0,0}\makebox(0,0)[t]{\lineheight{1.25}\smash{\begin{tabular}[t]{c}(d)~Geyer et al.~\shortcite{geyer_tokenflow_2024}\end{tabular}}}}%
    \put(0.02046819,0.29129200){\color[rgb]{0,0,0}\rotatebox{90}{\makebox(0,0)[t]{\lineheight{1.25}\smash{\begin{tabular}[t]{c}target frame\end{tabular}}}}}%
    \put(0.02046819,0.09614012){\color[rgb]{0,0,0}\rotatebox{90}{\makebox(0,0)[t]{\lineheight{1.25}\smash{\begin{tabular}[t]{c}our approach\end{tabular}}}}}%
  \end{picture}%
\endgroup%

%% file: figures/edits/edits.tex
\begingroup%
  \makeatletter%
  \providecommand\color[2][]{%
    \errmessage{(Inkscape) Color is used for the text in Inkscape, but the package 'color.sty' is not loaded}%
    \renewcommand\color[2][]{}%
  }%
  \providecommand\transparent[1]{%
    \errmessage{(Inkscape) Transparency is used (non-zero) for the text in Inkscape, but the package 'transparent.sty' is not loaded}%
    \renewcommand\transparent[1]{}%
  }%
  \providecommand\rotatebox[2]{#2}%
  \newcommand*\fsize{\dimexpr\f@size pt\relax}%
  \newcommand*\lineheight[1]{\fontsize{\fsize}{#1\fsize}\selectfont}%
  \ifx\svgwidth\undefined%
    \setlength{\unitlength}{531.32417454bp}%
    \ifx\svgscale\undefined%
      \relax%
    \else%
      \setlength{\unitlength}{\unitlength * \real{\svgscale}}%
    \fi%
  \else%
    \setlength{\unitlength}{\svgwidth}%
  \fi%
  \global\let\svgwidth\undefined%
  \global\let\svgscale\undefined%
  \makeatother%
  \begin{picture}(1,0.25)%
    \lineheight{1}%
    \setlength\tabcolsep{0pt}%
    \put(0,0){\includegraphics[width=\unitlength,page=1]{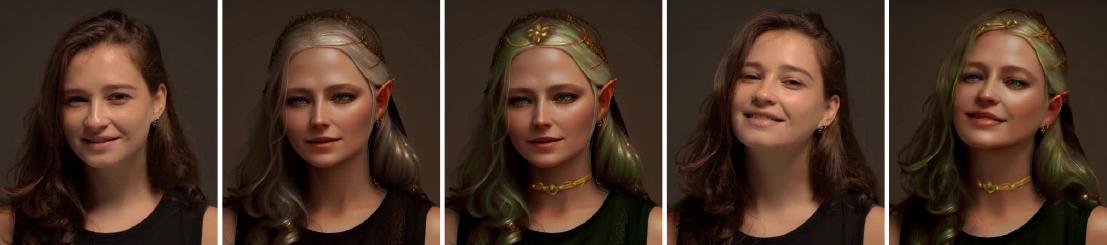}}%
    \put(0.00858475,0.015){\color[rgb]{1,1,1}\makebox(0,0)[lt]{\lineheight{1.25}\smash{\begin{tabular}[t]{l}(a)\end{tabular}}}}%
    \put(0.20973319,0.015){\color[rgb]{1,1,1}\makebox(0,0)[lt]{\lineheight{1.25}\smash{\begin{tabular}[t]{l}(b)\end{tabular}}}}%
    \put(0.41088161,0.015){\color[rgb]{1,1,1}\makebox(0,0)[lt]{\lineheight{1.25}\smash{\begin{tabular}[t]{l}(c)\end{tabular}}}}%
    \put(0.61203008,0.015){\color[rgb]{1,1,1}\makebox(0,0)[lt]{\lineheight{1.25}\smash{\begin{tabular}[t]{l}(d)\end{tabular}}}}%
    \put(0.81317851,0.015){\color[rgb]{1,1,1}\makebox(0,0)[lt]{\lineheight{1.25}\smash{\begin{tabular}[t]{l}(e)\end{tabular}}}}%

    \put(0.06,0.23029563){\color[rgb]{0,0,0}\makebox(0,0)[lt]{\lineheight{1.25}\smash{\begin{tabular}[t]{l}keyframe from Yang et al.~\shortcite{yang_rerender_2023} $(\mathbf{x}_r, \hat{\mathbf{x}}_r)$\end{tabular}}}}%
    \put(0.45,0.23029563){\color[rgb]{0,0,0}\makebox(0,0)[lt]{\lineheight{1.25}\smash{\begin{tabular}[t]{l}custom edit $\hat{\mathbf{x}}_e$\end{tabular}}}}%
    \put(0.65,0.23029563){\color[rgb]{0,0,0}\makebox(0,0)[lt]{\lineheight{1.25}\smash{\begin{tabular}[t]{l}target frame $\mathbf{y}$\end{tabular}}}}%
    \put(0.812,0.23029563){\color[rgb]{0,0,0}\makebox(0,0)[lt]{\lineheight{1.25}\smash{\begin{tabular}[t]{l}our method, stylization $\hat{\mathbf{y}}$\end{tabular}}}}%
  \end{picture}%
\endgroup%

%% file: figures/sota_comparison_dad_2/sota_comparison_dad_2.tex
\begingroup%
  \makeatletter%
  \providecommand\color[2][]{%
    \errmessage{(Inkscape) Color is used for the text in Inkscape, but the package 'color.sty' is not loaded}%
    \renewcommand\color[2][]{}%
  }%
  \providecommand\transparent[1]{%
    \errmessage{(Inkscape) Transparency is used (non-zero) for the text in Inkscape, but the package 'transparent.sty' is not loaded}%
    \renewcommand\transparent[1]{}%
  }%
  \providecommand\rotatebox[2]{#2}%
  \newcommand*\fsize{\dimexpr\f@size pt\relax}%
  \newcommand*\lineheight[1]{\fontsize{\fsize}{#1\fsize}\selectfont}%
  \ifx\svgwidth\undefined%
    \setlength{\unitlength}{485.03142109bp}%
    \ifx\svgscale\undefined%
      \relax%
    \else%
      \setlength{\unitlength}{\unitlength * \real{\svgscale}}%
    \fi%
  \else%
    \setlength{\unitlength}{\svgwidth}%
  \fi%
  \global\let\svgwidth\undefined%
  \global\let\svgscale\undefined%
  \makeatother%
  \begin{picture}(1,0.44150126)%
    \lineheight{1}%
    \setlength\tabcolsep{0pt}%
    \put(0,0){\includegraphics[width=\unitlength,page=1]{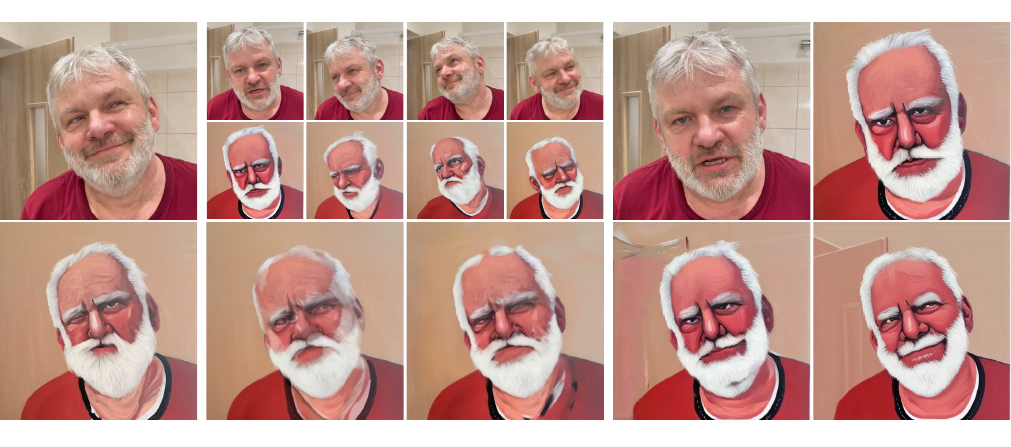}}%
    \put(0.09771721,0.42573636){\makebox(0,0)[t]{\lineheight{1.25}\smash{\begin{tabular}[t]{c}target frame $\mathbf{y}$\end{tabular}}}}%
    \put(0.40079691,0.42573636){\makebox(0,0)[t]{\lineheight{1.25}\smash{\begin{tabular}[t]{c}4 keyframes\end{tabular}}}}%
    \put(0.80293356,0.42573636){\makebox(0,0)[t]{\lineheight{1.25}\smash{\begin{tabular}[t]{c}1 keyframe\end{tabular}}}}%
    \put(0.09647897,0.00414757){\makebox(0,0)[t]{\lineheight{1.25}\smash{\begin{tabular}[t]{c}(a)~Ceylan et al.~\shortcite{ceylan_pix2video_2023}\end{tabular}}}}%
    \put(0.30173980,0.00414757){\makebox(0,0)[t]{\lineheight{1.25}\smash{\begin{tabular}[t]{c}(b)~Jamri\v{s}ka et al.~\shortcite{jamriska_stylizing_2019}\end{tabular}}}}%
    \put(0.50025673,0.00414757){\makebox(0,0)[t]{\lineheight{1.25}\smash{\begin{tabular}[t]{c}(c)~Texler et al.~\shortcite{texler_interactive_2020}\end{tabular}}}}%
    \put(0.70396224,0.00414757){\makebox(0,0)[t]{\lineheight{1.25}\smash{\begin{tabular}[t]{c}(d)~Futschik et al.~\shortcite{futschik_stalp_2021}\end{tabular}}}}%
    \put(0.90278594,0.00414757){\makebox(0,0)[t]{\lineheight{1.25}\smash{\begin{tabular}[t]{c}(e)~our approach\end{tabular}}}}%
  \end{picture}%
\endgroup%

%% file: figures/sota_comparison_lili_2/sota_comparison_lili_2.tex
\begingroup%
  \makeatletter%
  \providecommand\color[2][]{%
    \errmessage{(Inkscape) Color is used for the text in Inkscape, but the package 'color.sty' is not loaded}%
    \renewcommand\color[2][]{}%
  }%
  \providecommand\transparent[1]{%
    \errmessage{(Inkscape) Transparency is used (non-zero) for the text in Inkscape, but the package 'transparent.sty' is not loaded}%
    \renewcommand\transparent[1]{}%
  }%
  \providecommand\rotatebox[2]{#2}%
  \newcommand*\fsize{\dimexpr\f@size pt\relax}%
  \newcommand*\lineheight[1]{\fontsize{\fsize}{#1\fsize}\selectfont}%
  \ifx\svgwidth\undefined%
    \setlength{\unitlength}{485.03142109bp}%
    \ifx\svgscale\undefined%
      \relax%
    \else%
      \setlength{\unitlength}{\unitlength * \real{\svgscale}}%
    \fi%
  \else%
    \setlength{\unitlength}{\svgwidth}%
  \fi%
  \global\let\svgwidth\undefined%
  \global\let\svgscale\undefined%
  \makeatother%
  \begin{picture}(1,0.44150126)%
    \lineheight{1}%
    \setlength\tabcolsep{0pt}%
    \put(0,0){\includegraphics[width=\unitlength,page=1]{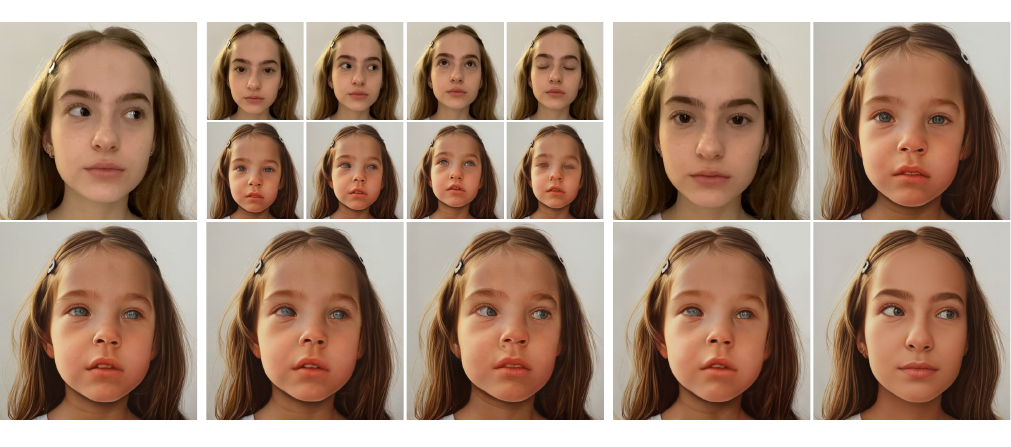}}%
    \put(0.09771721,0.42573636){\makebox(0,0)[t]{\lineheight{1.25}\smash{\begin{tabular}[t]{c}target frame $\mathbf{y}$\end{tabular}}}}%
    \put(0.40079691,0.42573636){\makebox(0,0)[t]{\lineheight{1.25}\smash{\begin{tabular}[t]{c}4 keyframes\end{tabular}}}}%
    \put(0.80293356,0.42573636){\makebox(0,0)[t]{\lineheight{1.25}\smash{\begin{tabular}[t]{c}1 keyframe\end{tabular}}}}%
    \put(0.09647897,0.00414757){\makebox(0,0)[t]{\lineheight{1.25}\smash{\begin{tabular}[t]{c}(a)~Geyer et al.~\shortcite{geyer_tokenflow_2024}\end{tabular}}}}%
    \put(0.30173980,0.00414757){\makebox(0,0)[t]{\lineheight{1.25}\smash{\begin{tabular}[t]{c}(b)~Jamri\v{s}ka et al.~\shortcite{jamriska_stylizing_2019}\end{tabular}}}}%
    \put(0.50025673,0.00414757){\makebox(0,0)[t]{\lineheight{1.25}\smash{\begin{tabular}[t]{c}(c)~Texler et al.~\shortcite{texler_interactive_2020}\end{tabular}}}}%
    \put(0.70396224,0.00414757){\makebox(0,0)[t]{\lineheight{1.25}\smash{\begin{tabular}[t]{c}(d)~Futschik et al.~\shortcite{futschik_stalp_2021}\end{tabular}}}}%
    \put(0.90278594,0.00414757){\makebox(0,0)[t]{\lineheight{1.25}\smash{\begin{tabular}[t]{c}(e)~our approach\end{tabular}}}}%
  \end{picture}%
\endgroup%

%% file: figures/sota_comparison_zuzka_2/sota_comparison_zuzka_2.tex
\begingroup%
  \makeatletter%
  \providecommand\color[2][]{%
    \errmessage{(Inkscape) Color is used for the text in Inkscape, but the package 'color.sty' is not loaded}%
    \renewcommand\color[2][]{}%
  }%
  \providecommand\transparent[1]{%
    \errmessage{(Inkscape) Transparency is used (non-zero) for the text in Inkscape, but the package 'transparent.sty' is not loaded}%
    \renewcommand\transparent[1]{}%
  }%
  \providecommand\rotatebox[2]{#2}%
  \newcommand*\fsize{\dimexpr\f@size pt\relax}%
  \newcommand*\lineheight[1]{\fontsize{\fsize}{#1\fsize}\selectfont}%
  \ifx\svgwidth\undefined%
    \setlength{\unitlength}{485.03142109bp}%
    \ifx\svgscale\undefined%
      \relax%
    \else%
      \setlength{\unitlength}{\unitlength * \real{\svgscale}}%
    \fi%
  \else%
    \setlength{\unitlength}{\svgwidth}%
  \fi%
  \global\let\svgwidth\undefined%
  \global\let\svgscale\undefined%
  \makeatother%
  \begin{picture}(1,0.44150126)%
    \lineheight{1}%
    \setlength\tabcolsep{0pt}%
    \put(0,0){\includegraphics[width=\unitlength,page=1]{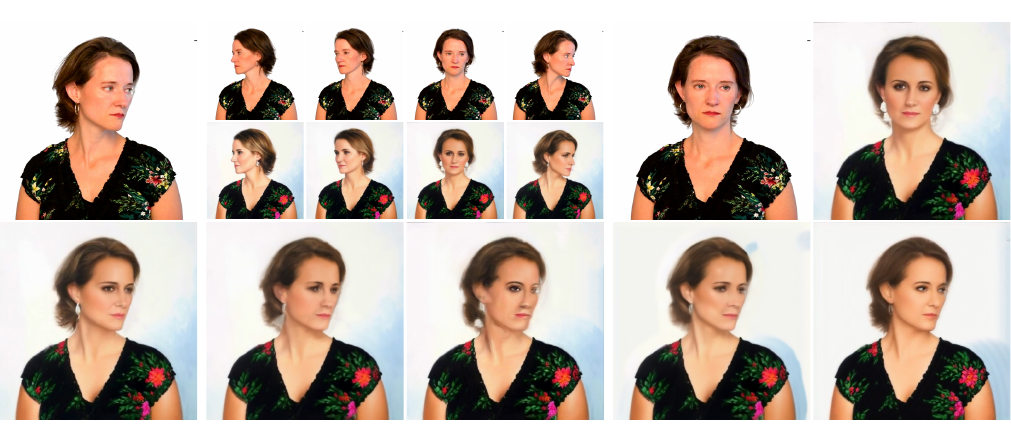}}%
    \put(0.09771721,0.42573636){\makebox(0,0)[t]{\lineheight{1.25}\smash{\begin{tabular}[t]{c}target frame \targetframefigure{}\end{tabular}}}}%
    \put(0.40079691,0.42573636){\makebox(0,0)[t]{\lineheight{1.25}\smash{\begin{tabular}[t]{c}4 keyframes\end{tabular}}}}%
    \put(0.80293356,0.42573636){\makebox(0,0)[t]{\lineheight{1.25}\smash{\begin{tabular}[t]{c}1 keyframe\end{tabular}}}}%
    \put(0.09647897,0.00414757){\makebox(0,0)[t]{\lineheight{1.25}\smash{\begin{tabular}[t]{c}(a)~Yang et al.~\shortcite{yang_rerender_2023}\end{tabular}}}}%
    \put(0.30173980,0.00414757){\makebox(0,0)[t]{\lineheight{1.25}\smash{\begin{tabular}[t]{c}(b)~Jamri\v{s}ka et al.~\shortcite{jamriska_stylizing_2019}\end{tabular}}}}%
    \put(0.50025673,0.00414757){\makebox(0,0)[t]{\lineheight{1.25}\smash{\begin{tabular}[t]{c}(c)~Texler et al.~\shortcite{texler_interactive_2020}\end{tabular}}}}%
    \put(0.70396224,0.00414757){\makebox(0,0)[t]{\lineheight{1.25}\smash{\begin{tabular}[t]{c}(d)~Futschik et al.~\shortcite{futschik_stalp_2021}\end{tabular}}}}%
    \put(0.90278594,0.00414757){\makebox(0,0)[t]{\lineheight{1.25}\smash{\begin{tabular}[t]{c}(e)~our approach\end{tabular}}}}%
  \end{picture}%
\endgroup%

%% file: figures/perceptual_study_temporal_consistency_style_structure/perceptual_study_temporal_consistency_style_structure.tex
\begingroup%
  \makeatletter%
  \providecommand\color[2][]{%
    \errmessage{(Inkscape) Color is used for the text in Inkscape, but the package 'color.sty' is not loaded}%
    \renewcommand\color[2][]{}%
  }%
  \providecommand\transparent[1]{%
    \errmessage{(Inkscape) Transparency is used (non-zero) for the text in Inkscape, but the package 'transparent.sty' is not loaded}%
    \renewcommand\transparent[1]{}%
  }%
  \providecommand\rotatebox[2]{#2}%
  \newcommand*\fsize{\dimexpr\f@size pt\relax}%
  \newcommand*\lineheight[1]{\fontsize{\fsize}{#1\fsize}\selectfont}%
  \ifx\svgwidth\undefined%
    \setlength{\unitlength}{360bp}%
    \ifx\svgscale\undefined%
      \relax%
    \else%
      \setlength{\unitlength}{\unitlength * \real{\svgscale}}%
    \fi%
  \else%
    \setlength{\unitlength}{\svgwidth}%
  \fi%
  \global\let\svgwidth\undefined%
  \global\let\svgscale\undefined%
  \makeatother%
  \begin{picture}(1,0.7)%
    \lineheight{1}%
    \setlength\tabcolsep{0pt}%
    \put(0,0){\includegraphics[width=\unitlength,page=1]{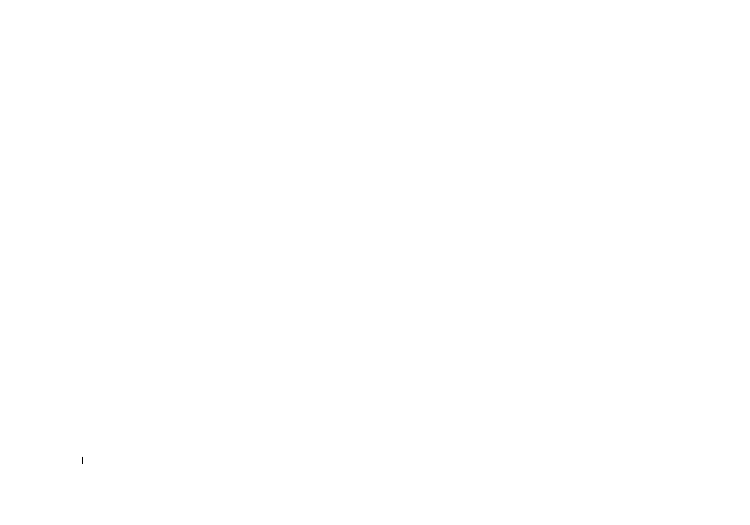}}%
    \put(0.08790798,0.0504618){\color[rgb]{0,0,0}\makebox(0,0)[lt]{\lineheight{1.25}\smash{\begin{tabular}[t]{l}0.0\end{tabular}}}}%
    \put(0,0){\includegraphics[width=\unitlength,page=2]{source_perceptual_study_temporal_consistency_style_structure.pdf}}%
    \put(0.26090798,0.0504618){\color[rgb]{0,0,0}\makebox(0,0)[lt]{\lineheight{1.25}\smash{\begin{tabular}[t]{l}0.2\end{tabular}}}}%
    \put(0,0){\includegraphics[width=\unitlength,page=3]{source_perceptual_study_temporal_consistency_style_structure.pdf}}%
    \put(0.43390796,0.0504618){\color[rgb]{0,0,0}\makebox(0,0)[lt]{\lineheight{1.25}\smash{\begin{tabular}[t]{l}0.4\end{tabular}}}}%
    \put(0,0){\includegraphics[width=\unitlength,page=4]{source_perceptual_study_temporal_consistency_style_structure.pdf}}%
    \put(0.60690796,0.0504618){\color[rgb]{0,0,0}\makebox(0,0)[lt]{\lineheight{1.25}\smash{\begin{tabular}[t]{l}0.6\end{tabular}}}}%
    \put(0,0){\includegraphics[width=\unitlength,page=5]{source_perceptual_study_temporal_consistency_style_structure.pdf}}%
    \put(0.77990795,0.0504618){\color[rgb]{0,0,0}\makebox(0,0)[lt]{\lineheight{1.25}\smash{\begin{tabular}[t]{l}0.8\end{tabular}}}}%
    \put(0,0){\includegraphics[width=\unitlength,page=6]{source_perceptual_study_temporal_consistency_style_structure.pdf}}%
    \put(0.95290795,0.0504618){\color[rgb]{0,0,0}\makebox(0,0)[lt]{\lineheight{1.25}\smash{\begin{tabular}[t]{l}1.0\end{tabular}}}}%
    \put(0.37498696,0.01248437){\color[rgb]{0,0,0}\makebox(0,0)[lt]{\lineheight{1.25}\smash{\begin{tabular}[t]{l}Temporal consistency $\square$\end{tabular}}}}%
    \put(0.04637153,0.08045312){\color[rgb]{0,0,0}\makebox(0,0)[lt]{\lineheight{1.25}\smash{\begin{tabular}[t]{l}0.0\end{tabular}}}}%
    \put(0,0){\includegraphics[width=\unitlength,page=8]{source_perceptual_study_temporal_consistency_style_structure.pdf}}%
    \put(0.04637153,0.18545312){\color[rgb]{0,0,0}\makebox(0,0)[lt]{\lineheight{1.25}\smash{\begin{tabular}[t]{l}0.2\end{tabular}}}}%
    \put(0,0){\includegraphics[width=\unitlength,page=9]{source_perceptual_study_temporal_consistency_style_structure.pdf}}%
    \put(0.04637153,0.29045313){\color[rgb]{0,0,0}\makebox(0,0)[lt]{\lineheight{1.25}\smash{\begin{tabular}[t]{l}0.4\end{tabular}}}}%
    \put(0,0){\includegraphics[width=\unitlength,page=10]{source_perceptual_study_temporal_consistency_style_structure.pdf}}%
    \put(0.04637153,0.3954531){\color[rgb]{0,0,0}\makebox(0,0)[lt]{\lineheight{1.25}\smash{\begin{tabular}[t]{l}0.6\end{tabular}}}}%
    \put(0,0){\includegraphics[width=\unitlength,page=11]{source_perceptual_study_temporal_consistency_style_structure.pdf}}%
    \put(0.04637153,0.5004531){\color[rgb]{0,0,0}\makebox(0,0)[lt]{\lineheight{1.25}\smash{\begin{tabular}[t]{l}0.8\end{tabular}}}}%
    \put(0,0){\includegraphics[width=\unitlength,page=12]{source_perceptual_study_temporal_consistency_style_structure.pdf}}%
    \put(0.04637153,0.6054531){\color[rgb]{0,0,0}\makebox(0,0)[lt]{\lineheight{1.25}\smash{\begin{tabular}[t]{l}1.0\end{tabular}}}}%
    \put(0.02948784,0.31860416){\color[rgb]{0,0,0}\rotatebox{90}{\makebox(0,0)[lt]{\lineheight{1.25}\smash{\begin{tabular}[t]{l}Style\end{tabular}}}}}%
    \put(0,0){\includegraphics[width=\unitlength,page=13]{source_perceptual_study_temporal_consistency_style_structure.pdf}}%
    \put(0.22813371,0.42429088){\color[rgb]{0,0,0}\makebox(0,0)[lt]{\lineheight{1.25}\smash{\begin{tabular}[t]{l}Ours vs. Geyer et al. \shortcite{geyer_tokenflow_2024}\end{tabular}}}}%
    \put(0,0){\includegraphics[width=\unitlength,page=14]{source_perceptual_study_temporal_consistency_style_structure.pdf}}%
    \put(0.22813371,0.38353566){\color[rgb]{0,0,0}\makebox(0,0)[lt]{\lineheight{1.25}\smash{\begin{tabular}[t]{l}Ours vs. Yang et al. \shortcite{yang_rerender_2023}\end{tabular}}}}%
    \put(0,0){\includegraphics[width=\unitlength,page=15]{source_perceptual_study_temporal_consistency_style_structure.pdf}}%
    \put(0.22813371,0.34278046){\color[rgb]{0,0,0}\makebox(0,0)[lt]{\lineheight{1.25}\smash{\begin{tabular}[t]{l}Ours vs. Chu et al. \shortcite{chu_medm_2024}\end{tabular}}}}%
    \put(0,0){\includegraphics[width=\unitlength,page=16]{source_perceptual_study_temporal_consistency_style_structure.pdf}}%
    \put(0.22813371,0.30202524){\color[rgb]{0,0,0}\makebox(0,0)[lt]{\lineheight{1.25}\smash{\begin{tabular}[t]{l}Ours vs. Ceylan et al. \shortcite{ceylan_pix2video_2023}\end{tabular}}}}%
    \put(0,0){\includegraphics[width=\unitlength,page=17]{source_perceptual_study_temporal_consistency_style_structure.pdf}}%
    \put(0.22813371,0.26127006){\color[rgb]{0,0,0}\makebox(0,0)[lt]{\lineheight{1.25}\smash{\begin{tabular}[t]{l}Ours vs. Futschik et al. \shortcite{futschik_stalp_2021}\end{tabular}}}}%
    \put(0,0){\includegraphics[width=\unitlength,page=18]{source_perceptual_study_temporal_consistency_style_structure.pdf}}%
    \put(0.22813371,0.22051484){\color[rgb]{0,0,0}\makebox(0,0)[lt]{\lineheight{1.25}\smash{\begin{tabular}[t]{l}Ours vs. Texler et al. \shortcite{texler_interactive_2020}\end{tabular}}}}%
    \put(0,0){\includegraphics[width=\unitlength,page=19]{source_perceptual_study_temporal_consistency_style_structure.pdf}}%
    \put(0.22813371,0.17884817){\color[rgb]{0,0,0}\makebox(0,0)[lt]{\lineheight{1.25}\smash{\begin{tabular}[t]{l}Ours vs. Jamri\v ska et al.~\shortcite{jamriska_stylizing_2019}\end{tabular}}}}%
    \put(0,0){\includegraphics[width=\unitlength,page=20]{source_perceptual_study_temporal_consistency_style_structure.pdf}}%
    \put(0.08790798,0.64121701){\color[rgb]{0,0,0}\makebox(0,0)[lt]{\lineheight{1.25}\smash{\begin{tabular}[t]{l}0.0\end{tabular}}}}%
    \put(0,0){\includegraphics[width=\unitlength,page=21]{source_perceptual_study_temporal_consistency_style_structure.pdf}}%
    \put(0.26090798,0.64121701){\color[rgb]{0,0,0}\makebox(0,0)[lt]{\lineheight{1.25}\smash{\begin{tabular}[t]{l}0.2\end{tabular}}}}%
    \put(0,0){\includegraphics[width=\unitlength,page=22]{source_perceptual_study_temporal_consistency_style_structure.pdf}}%
    \put(0.43390796,0.64121701){\color[rgb]{0,0,0}\makebox(0,0)[lt]{\lineheight{1.25}\smash{\begin{tabular}[t]{l}0.4\end{tabular}}}}%
    \put(0,0){\includegraphics[width=\unitlength,page=23]{source_perceptual_study_temporal_consistency_style_structure.pdf}}%
    \put(0.60690796,0.64121701){\color[rgb]{0,0,0}\makebox(0,0)[lt]{\lineheight{1.25}\smash{\begin{tabular}[t]{l}0.6\end{tabular}}}}%
    \put(0,0){\includegraphics[width=\unitlength,page=24]{source_perceptual_study_temporal_consistency_style_structure.pdf}}%
    \put(0.77990795,0.64121701){\color[rgb]{0,0,0}\makebox(0,0)[lt]{\lineheight{1.25}\smash{\begin{tabular}[t]{l}0.8\end{tabular}}}}%
    \put(0,0){\includegraphics[width=\unitlength,page=25]{source_perceptual_study_temporal_consistency_style_structure.pdf}}%
    \put(0.95290795,0.64121701){\color[rgb]{0,0,0}\makebox(0,0)[lt]{\lineheight{1.25}\smash{\begin{tabular}[t]{l}1.0\end{tabular}}}}%
    \put(0.46281249,0.67342184){\color[rgb]{0,0,0}\makebox(0,0)[lt]{\lineheight{1.25}\smash{\begin{tabular}[t]{l}Structure $\triangle$\end{tabular}}}}%
    \put(0,0){\includegraphics[width=\unitlength,page=26]{source_perceptual_study_temporal_consistency_style_structure.pdf}}%
  \end{picture}%
\endgroup%

%% file: figures/timelapse/timelapse.tex
\begingroup%
  \makeatletter%
  \providecommand\color[2][]{%
    \errmessage{(Inkscape) Color is used for the text in Inkscape, but the package 'color.sty' is not loaded}%
    \renewcommand\color[2][]{}%
  }%
  \providecommand\transparent[1]{%
    \errmessage{(Inkscape) Transparency is used (non-zero) for the text in Inkscape, but the package 'transparent.sty' is not loaded}%
    \renewcommand\transparent[1]{}%
  }%
  \providecommand\rotatebox[2]{#2}%
  \newcommand*\fsize{\dimexpr\f@size pt\relax}%
  \newcommand*\lineheight[1]{\fontsize{\fsize}{#1\fsize}\selectfont}%
  \ifx\svgwidth\undefined%
    \setlength{\unitlength}{486.02486197bp}%
    \ifx\svgscale\undefined%
      \relax%
    \else%
      \setlength{\unitlength}{\unitlength * \real{\svgscale}}%
    \fi%
  \else%
    \setlength{\unitlength}{\svgwidth}%
  \fi%
  \global\let\svgwidth\undefined%
  \global\let\svgscale\undefined%
  \makeatother%
  \begin{picture}(1,0.49768527)%
    \lineheight{1}%
    \setlength\tabcolsep{0pt}%
    \put(0,0){\includegraphics[width=\unitlength,page=1]{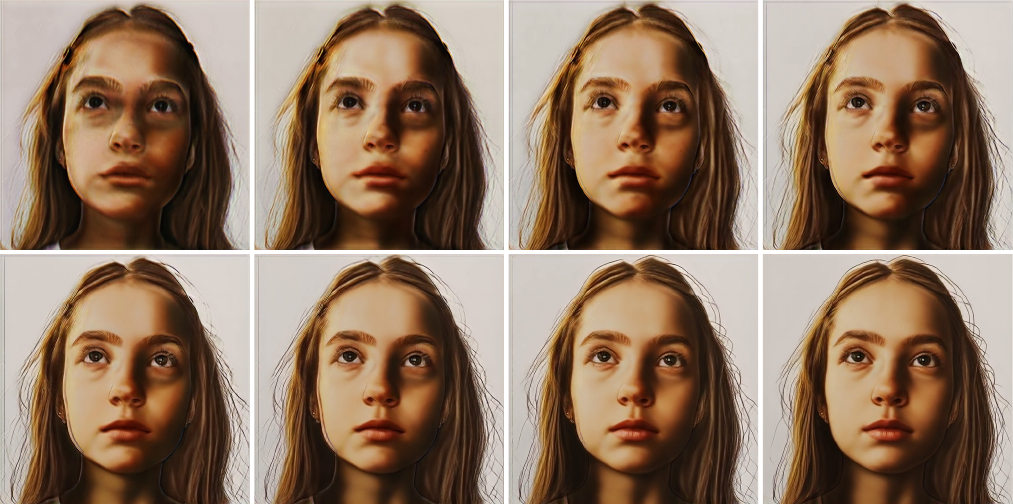}}%
    \put(0.00811427,0.46717562){\color[rgb]{0,0,0}\makebox(0,0)[lt]{\lineheight{1.25}\smash{\begin{tabular}[t]{l}1m\end{tabular}}}}%
    \put(0.25913042,0.46717562){\color[rgb]{0,0,0}\makebox(0,0)[lt]{\lineheight{1.25}\smash{\begin{tabular}[t]{l}2m\end{tabular}}}}%
    \put(0.51014657,0.46761164){\color[rgb]{0,0,0}\makebox(0,0)[lt]{\lineheight{1.25}\smash{\begin{tabular}[t]{l}3m\end{tabular}}}}%
    \put(0.76116271,0.46761164){\color[rgb]{0,0,0}\makebox(0,0)[lt]{\lineheight{1.25}\smash{\begin{tabular}[t]{l}6m\end{tabular}}}}%
    \put(0.00811427,0.21718819){\color[rgb]{0,0,0}\makebox(0,0)[lt]{\lineheight{1.25}\smash{\begin{tabular}[t]{l}10m\end{tabular}}}}%
    \put(0.25913042,0.21718819){\color[rgb]{0,0,0}\makebox(0,0)[lt]{\lineheight{1.25}\smash{\begin{tabular}[t]{l}20m\end{tabular}}}}%
    \put(0.51014657,0.21762422){\color[rgb]{0,0,0}\makebox(0,0)[lt]{\lineheight{1.25}\smash{\begin{tabular}[t]{l}45m\end{tabular}}}}%
    \put(0.76116271,0.21762422){\color[rgb]{0,0,0}\makebox(0,0)[lt]{\lineheight{1.25}\smash{\begin{tabular}[t]{l}90m\end{tabular}}}}%
  \end{picture}%
\endgroup%

%% file: figures/loss_stability/loss_stability.tex
\begingroup%
  \makeatletter%
  \providecommand\color[2][]{%
    \errmessage{(Inkscape) Color is used for the text in Inkscape, but the package 'color.sty' is not loaded}%
    \renewcommand\color[2][]{}%
  }%
  \providecommand\transparent[1]{%
    \errmessage{(Inkscape) Transparency is used (non-zero) for the text in Inkscape, but the package 'transparent.sty' is not loaded}%
    \renewcommand\transparent[1]{}%
  }%
  \providecommand\rotatebox[2]{#2}%
  \newcommand*\fsize{\dimexpr\f@size pt\relax}%
  \newcommand*\lineheight[1]{\fontsize{\fsize}{#1\fsize}\selectfont}%
  \ifx\svgwidth\undefined%
    \setlength{\unitlength}{790.5bp}%
    \ifx\svgscale\undefined%
      \relax%
    \else%
      \setlength{\unitlength}{\unitlength * \real{\svgscale}}%
    \fi%
  \else%
    \setlength{\unitlength}{\svgwidth}%
  \fi%
  \global\let\svgwidth\undefined%
  \global\let\svgscale\undefined%
  \makeatother%
  \begin{picture}(1,0.70113852)%
    \lineheight{1}%
    \setlength\tabcolsep{0pt}%
    \put(0,0){\includegraphics[width=\unitlength,page=1]{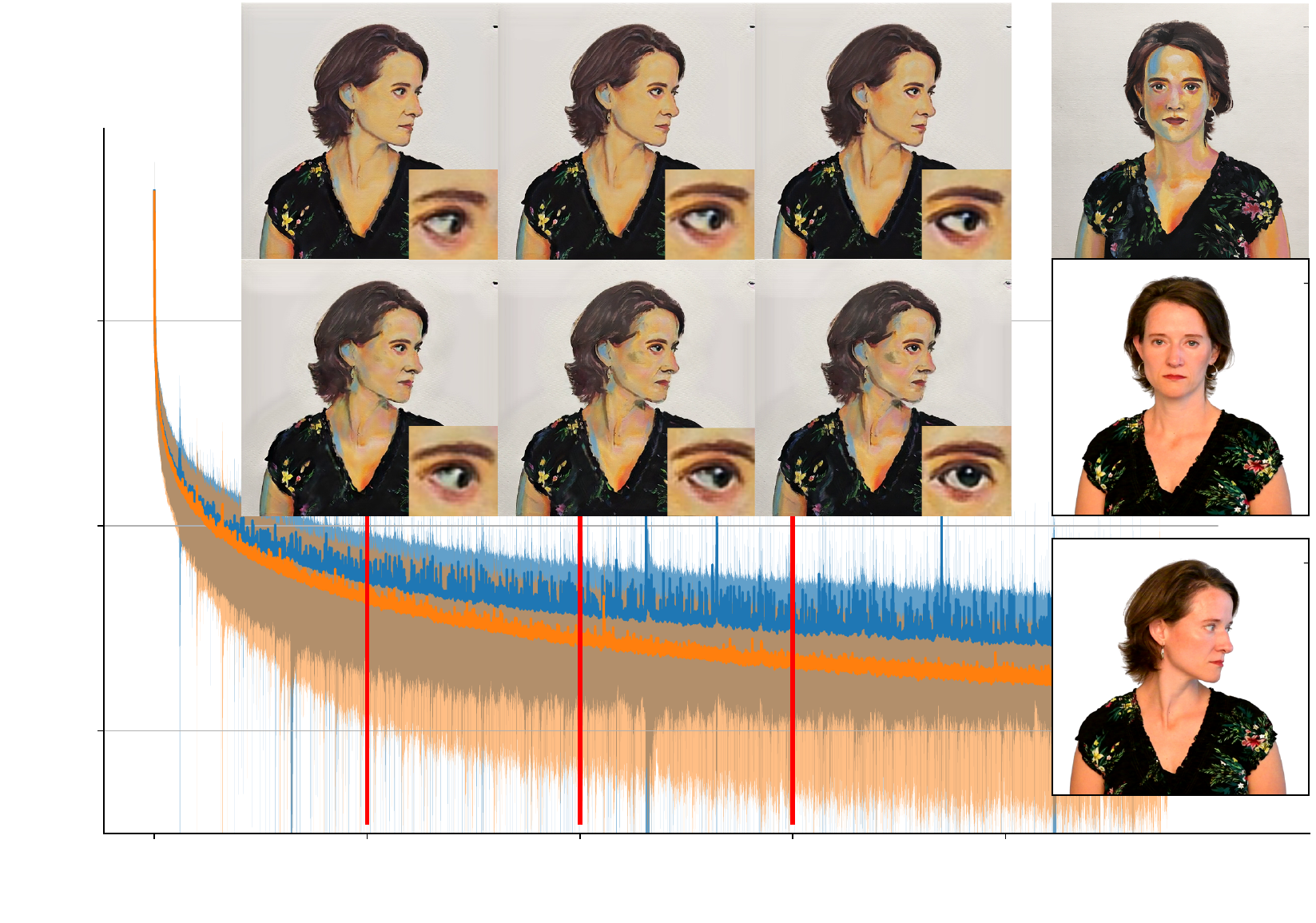}}%
    \put(0.11103947,0.034){\color[rgb]{0,0,0}\makebox(0,0)[lt]{\lineheight{1.25}\smash{\begin{tabular}[t]{l}$0$\end{tabular}}}}%
    \put(0.0,0.13852874){\color[rgb]{0,0,0}\makebox(0,0)[lt]{\lineheight{1.25}\smash{\begin{tabular}[t]{l}$10^{-3}$\end{tabular}}}}%
    \put(0.0,0.29431381){\color[rgb]{0,0,0}\makebox(0,0)[lt]{\lineheight{1.25}\smash{\begin{tabular}[t]{l}$10^{-2}$\end{tabular}}}}%
    \put(0.0,0.45009875){\color[rgb]{0,0,0}\makebox(0,0)[lt]{\lineheight{1.25}\smash{\begin{tabular}[t]{l}$10^{-1}$\end{tabular}}}}%
    \put(0.26066585,0.034){\color[rgb]{0,0,0}\makebox(0,0)[lt]{\lineheight{1.25}\smash{\begin{tabular}[t]{l}$30$\end{tabular}}}}%
    \put(0.420546,0.034){\color[rgb]{0,0,0}\makebox(0,0)[lt]{\lineheight{1.25}\smash{\begin{tabular}[t]{l}$60$\end{tabular}}}}%
    \put(0.45564875,0.00){\color[rgb]{0,0,0}\makebox(0,0)[lt]{\lineheight{1.25}\smash{\begin{tabular}[t]{l}Minutes\end{tabular}}}}%
    \put(0.0,0.20884049){\color[rgb]{0,0,0}\rotatebox{90}{\makebox(0,0)[lt]{\lineheight{1.25}\smash{\begin{tabular}[t]{l}Total\hspace{0.8em} loss\end{tabular}}}}}%
    \put(0.19,0.6654049){\color[rgb]{0,0,0}\makebox(0,0)[lt]{\lineheight{1.25}\smash{\begin{tabular}[t]{l}(a)\end{tabular}}}}
    \put(0.383,0.6654049){\color[rgb]{0,0,0}\makebox(0,0)[lt]{\lineheight{1.25}\smash{\begin{tabular}[t]{l}(b)\end{tabular}}}}
    \put(0.578,0.6654049){\color[rgb]{0,0,0}\makebox(0,0)[lt]{\lineheight{1.25}\smash{\begin{tabular}[t]{l}(c)\end{tabular}}}}
    \put(0.808,0.6654049){\color[rgb]{0,0,0}\makebox(0,0)[lt]{\lineheight{1.25}\smash{\begin{tabular}[t]{l}$\hat{\mathbf{x}}$\end{tabular}}}}
    \put(0.19,0.470){\color[rgb]{0,0,0}\makebox(0,0)[lt]{\lineheight{1.25}\smash{\begin{tabular}[t]{l}(d)\end{tabular}}}}
    \put(0.383,0.470){\color[rgb]{0,0,0}\makebox(0,0)[lt]{\lineheight{1.25}\smash{\begin{tabular}[t]{l}(e)\end{tabular}}}}
    \put(0.578,0.470){\color[rgb]{0,0,0}\makebox(0,0)[lt]{\lineheight{1.25}\smash{\begin{tabular}[t]{l}(f)\end{tabular}}}}
    \put(0.808,0.470){\color[rgb]{0,0,0}\makebox(0,0)[lt]{\lineheight{1.25}\smash{\begin{tabular}[t]{l}$\mathbf{x}$\end{tabular}}}}
    \put(0.808,0.260){\color[rgb]{0,0,0}\makebox(0,0)[lt]{\lineheight{1.25}\smash{\begin{tabular}[t]{l}$\mathbf{y}$\end{tabular}}}}
    \put(0.58362523,0.034){\color[rgb]{0,0,0}\makebox(0,0)[lt]{\lineheight{1.25}\smash{\begin{tabular}[t]{l}$90$\end{tabular}}}}%
    \put(0.7329516,0.034){\color[rgb]{0,0,0}\makebox(0,0)[lt]{\lineheight{1.25}\smash{\begin{tabular}[t]{l}$120$\end{tabular}}}}%
  \end{picture}%
\endgroup%

%% file: figures/loss_ablation/loss_ablation.tex
\begingroup%
  \makeatletter%
  \providecommand\color[2][]{%
    \errmessage{(Inkscape) Color is used for the text in Inkscape, but the package 'color.sty' is not loaded}%
    \renewcommand\color[2][]{}%
  }%
  \providecommand\transparent[1]{%
    \errmessage{(Inkscape) Transparency is used (non-zero) for the text in Inkscape, but the package 'transparent.sty' is not loaded}%
    \renewcommand\transparent[1]{}%
  }%
  \providecommand\rotatebox[2]{#2}%
  \newcommand*\fsize{\dimexpr\f@size pt\relax}%
  \newcommand*\lineheight[1]{\fontsize{\fsize}{#1\fsize}\selectfont}%
  \ifx\svgwidth\undefined%
    \setlength{\unitlength}{483.89468216bp}%
    \ifx\svgscale\undefined%
      \relax%
    \else%
      \setlength{\unitlength}{\unitlength * \real{\svgscale}}%
    \fi%
  \else%
    \setlength{\unitlength}{\svgwidth}%
  \fi%
  \global\let\svgwidth\undefined%
  \global\let\svgscale\undefined%
  \makeatother%
  \begin{picture}(1,0.24635067)%
    \lineheight{1}%
    \setlength\tabcolsep{0pt}%
    \put(0,0){\includegraphics[width=\unitlength,page=1]{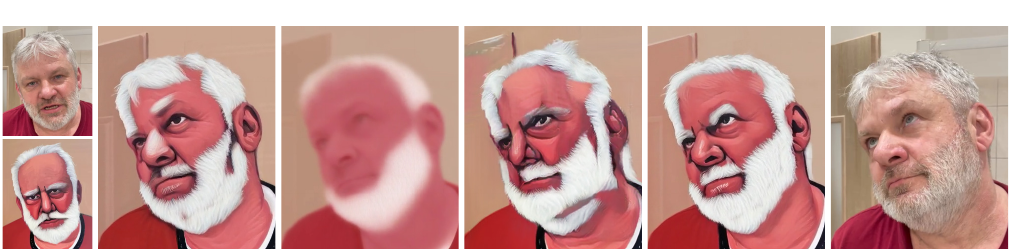}}%
    \put(0.04660861,0.23054874){\makebox(0,0)[t]{\lineheight{1.25}\smash{\begin{tabular}[t]{c}keyframe\end{tabular}}}}%
    \put(0.91224194,0.23054874){\makebox(0,0)[t]{\lineheight{1.25}\smash{\begin{tabular}[t]{c}target frame~$\mathbf{y}$\end{tabular}}}}%
    \put(0.18530079,0.23054874){\makebox(0,0)[t]{\lineheight{1.25}\smash{\begin{tabular}[t]{c}$\mathcal{L}_{\text{style}} + \mathcal{L}_{\text{structure}}$\end{tabular}}}}%
    \put(0.36698816,0.23054874){\makebox(0,0)[t]{\lineheight{1.25}\smash{\begin{tabular}[t]{c}$\mathcal{L}_{\text{key}} + \mathcal{L}_{\text{structure}}$\end{tabular}}}}%
    \put(0.54867557,0.23054874){\makebox(0,0)[t]{\lineheight{1.25}\smash{\begin{tabular}[t]{c}$\mathcal{L}_{\text{key}} + \mathcal{L}_{\text{style}}$\end{tabular}}}}%
    \put(0.73036289,0.23054874){\makebox(0,0)[t]{\lineheight{1.25}\smash{\begin{tabular}[t]{c}$\mathcal{L}_{\text{key}} + \mathcal{L}_{\text{style}} + \mathcal{L}_{\text{structure}}$\end{tabular}}}}%
    \put(0.12518793,0.00904366){\color[rgb]{1,1,1}\makebox(0,0)[t]{\lineheight{1.25}\smash{\begin{tabular}[t]{c}(a)\end{tabular}}}}%
    \put(0.30657718,0.00904366){\color[rgb]{1,1,1}\makebox(0,0)[t]{\lineheight{1.25}\smash{\begin{tabular}[t]{c}(b)\end{tabular}}}}%
    \put(0.48796686,0.00904366){\color[rgb]{1,1,1}\makebox(0,0)[t]{\lineheight{1.25}\smash{\begin{tabular}[t]{c}(c)\end{tabular}}}}%
    \put(0.66935774,0.00904366){\color[rgb]{1,1,1}\makebox(0,0)[t]{\lineheight{1.25}\smash{\begin{tabular}[t]{c}(d)\end{tabular}}}}%
  \end{picture}%
\endgroup%

%% file: figures/style_as_function_of_sds_attributes_3/style_as_function_of_sds_attributes_3.tex
\begingroup%
  \makeatletter%
  \providecommand\color[2][]{%
    \errmessage{(Inkscape) Color is used for the text in Inkscape, but the package 'color.sty' is not loaded}%
    \renewcommand\color[2][]{}%
  }%
  \providecommand\transparent[1]{%
    \errmessage{(Inkscape) Transparency is used (non-zero) for the text in Inkscape, but the package 'transparent.sty' is not loaded}%
    \renewcommand\transparent[1]{}%
  }%
  \providecommand\rotatebox[2]{#2}%
  \newcommand*\fsize{\dimexpr\f@size pt\relax}%
  \newcommand*\lineheight[1]{\fontsize{\fsize}{#1\fsize}\selectfont}%
  \ifx\svgwidth\undefined%
    \setlength{\unitlength}{447.65307809bp}%
    \ifx\svgscale\undefined%
      \relax%
    \else%
      \setlength{\unitlength}{\unitlength * \real{\svgscale}}%
    \fi%
  \else%
    \setlength{\unitlength}{\svgwidth}%
  \fi%
  \global\let\svgwidth\undefined%
  \global\let\svgscale\undefined%
  \makeatother%
  \begin{picture}(1,1.2)%
    \lineheight{1}%
    \setlength\tabcolsep{0pt}%
    \put(0,0){\includegraphics[width=\unitlength,page=1]{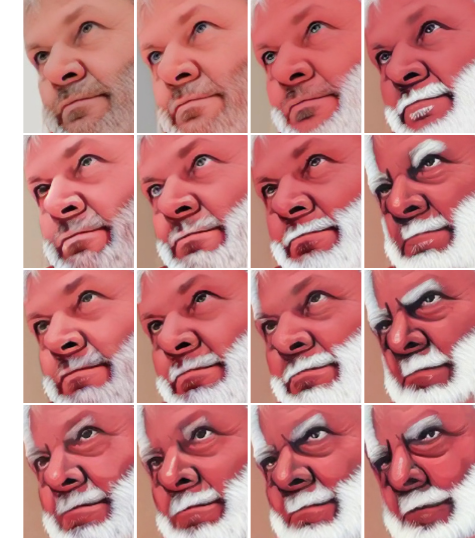}}%
    %
    %
    \put(0.17,0.0){\makebox(0,0)[t]{\lineheight{1.25}\smash{\begin{tabular}[t]{c}$t=16$\end{tabular}}}}%
    \put(0.41,0.0){\makebox(0,0)[t]{\lineheight{1.25}\smash{\begin{tabular}[t]{c}$t=20$\end{tabular}}}}%
    \put(0.65,0.0){\makebox(0,0)[t]{\lineheight{1.25}\smash{\begin{tabular}[t]{c}$t=24$\end{tabular}}}}%
    \put(0.89,0.0){\makebox(0,0)[t]{\lineheight{1.25}\smash{\begin{tabular}[t]{c}$t=28$\end{tabular}}}}%
    \put(0.01,0.950){\makebox(0,0)[t]{\lineheight{1.25}\smash{\begin{tabular}[t]{c}\rotatebox{90}{$\lambda_s=5 \cdot 10^{-5}$}\end{tabular}}}}%
    \put(0.01,0.662){\makebox(0,0)[t]{\lineheight{1.25}\smash{\begin{tabular}[t]{c}\rotatebox{90}{$\lambda_s=1 \cdot 10^{-5}$}\end{tabular}}}}%
    \put(0.01,0.376){\makebox(0,0)[t]{\lineheight{1.25}\smash{\begin{tabular}[t]{c}\rotatebox{90}{$\lambda_s=5 \cdot 10^{-6}$}\end{tabular}}}}%
    \put(0.01,0.090){\makebox(0,0)[t]{\lineheight{1.25}\smash{\begin{tabular}[t]{c}\rotatebox{90}{$\lambda_s=1 \cdot 10^{-6}$}\end{tabular}}}}%
  \end{picture}%
\endgroup%

%% file: figures/control_type_2/control_type_2.tex
\begingroup%
  \makeatletter%
  \providecommand\color[2][]{%
    \errmessage{(Inkscape) Color is used for the text in Inkscape, but the package 'color.sty' is not loaded}%
    \renewcommand\color[2][]{}%
  }%
  \providecommand\transparent[1]{%
    \errmessage{(Inkscape) Transparency is used (non-zero) for the text in Inkscape, but the package 'transparent.sty' is not loaded}%
    \renewcommand\transparent[1]{}%
  }%
  \providecommand\rotatebox[2]{#2}%
  \newcommand*\fsize{\dimexpr\f@size pt\relax}%
  \newcommand*\lineheight[1]{\fontsize{\fsize}{#1\fsize}\selectfont}%
  \ifx\svgwidth\undefined%
    \setlength{\unitlength}{283.03850069bp}%
    \ifx\svgscale\undefined%
      \relax%
    \else%
      \setlength{\unitlength}{\unitlength * \real{\svgscale}}%
    \fi%
  \else%
    \setlength{\unitlength}{.8\svgwidth}%
  \fi%
  \global\let\svgwidth\undefined%
  \global\let\svgscale\undefined%
  \makeatother%
  \begin{picture}(1,1.67995841)%
    \lineheight{1}%
    \setlength\tabcolsep{0pt}%
    \put(0,0){\includegraphics[width=\unitlength,page=1]{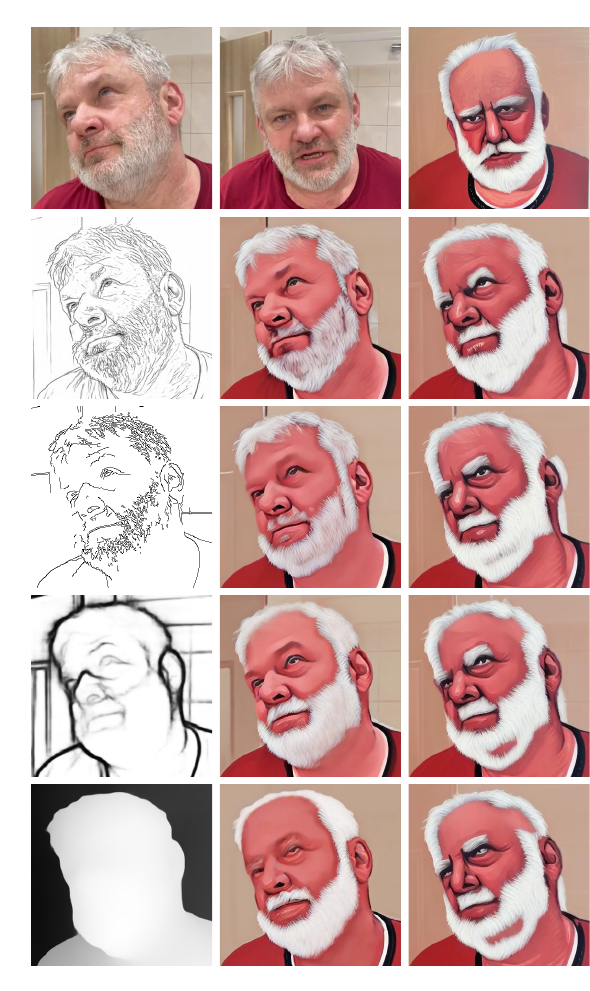}}%
    \put(0.03241882,1.15557210){\rotatebox{90}{\makebox(0,0)[t]{\lineheight{1.25}\smash{\begin{tabular}[t]{c}lineart\end{tabular}}}}}%
    \put(0.03241882,0.83632195){\rotatebox{90}{\makebox(0,0)[t]{\lineheight{1.25}\smash{\begin{tabular}[t]{c}Canny\end{tabular}}}}}%
    \put(0.03241882,0.51658523){\rotatebox{90}{\makebox(0,0)[t]{\lineheight{1.25}\smash{\begin{tabular}[t]{c}soft edge\end{tabular}}}}}%
    \put(0.03241882,0.19721107){\rotatebox{90}{\makebox(0,0)[t]{\lineheight{1.25}\smash{\begin{tabular}[t]{c}depth\end{tabular}}}}}%
    \put(0.20563719,0.00064181){\makebox(0,0)[t]{\lineheight{1.25}\smash{\begin{tabular}[t]{c}$\mathbf{c}$\end{tabular}}}}%
    \put(0.52684359,0.00064181){\makebox(0,0)[t]{\lineheight{1.25}\smash{\begin{tabular}[t]{c}$t=16$\end{tabular}}}}%
    \put(0.84703579,0.00064181){\makebox(0,0)[t]{\lineheight{1.25}\smash{\begin{tabular}[t]{c}$t=28$\\\end{tabular}}}}%
    \put(0.20653770,1.64753959){\makebox(0,0)[t]{\lineheight{1.25}\smash{\begin{tabular}[t]{c}target frame\end{tabular}}}}%
    \put(0.68632899,1.64753959){\makebox(0,0)[t]{\lineheight{1.25}\smash{\begin{tabular}[t]{c}stylized keyframe\end{tabular}}}}%
  \end{picture}%
\endgroup%

%% file: figures/difflineart/difflineart.tex
\begingroup%
  \makeatletter%
  \providecommand\color[2][]{%
    \errmessage{(Inkscape) Color is used for the text in Inkscape, but the package 'color.sty' is not loaded}%
    \renewcommand\color[2][]{}%
  }%
  \providecommand\transparent[1]{%
    \errmessage{(Inkscape) Transparency is used (non-zero) for the text in Inkscape, but the package 'transparent.sty' is not loaded}%
    \renewcommand\transparent[1]{}%
  }%
  \providecommand\rotatebox[2]{#2}%
  \newcommand*\fsize{\dimexpr\f@size pt\relax}%
  \newcommand*\lineheight[1]{\fontsize{\fsize}{#1\fsize}\selectfont}%
  \ifx\svgwidth\undefined%
    \setlength{\unitlength}{531.99723852bp}%
    \ifx\svgscale\undefined%
      \relax%
    \else%
      \setlength{\unitlength}{\unitlength * \real{\svgscale}}%
    \fi%
  \else%
    \setlength{\unitlength}{\svgwidth}%
  \fi%
  \global\let\svgwidth\undefined%
  \global\let\svgscale\undefined%
  \makeatother%
  \begin{picture}(1,0.60477941)%
    \lineheight{1}%
    \setlength\tabcolsep{0pt}%
    \put(0,0){\includegraphics[width=\unitlength,page=1]{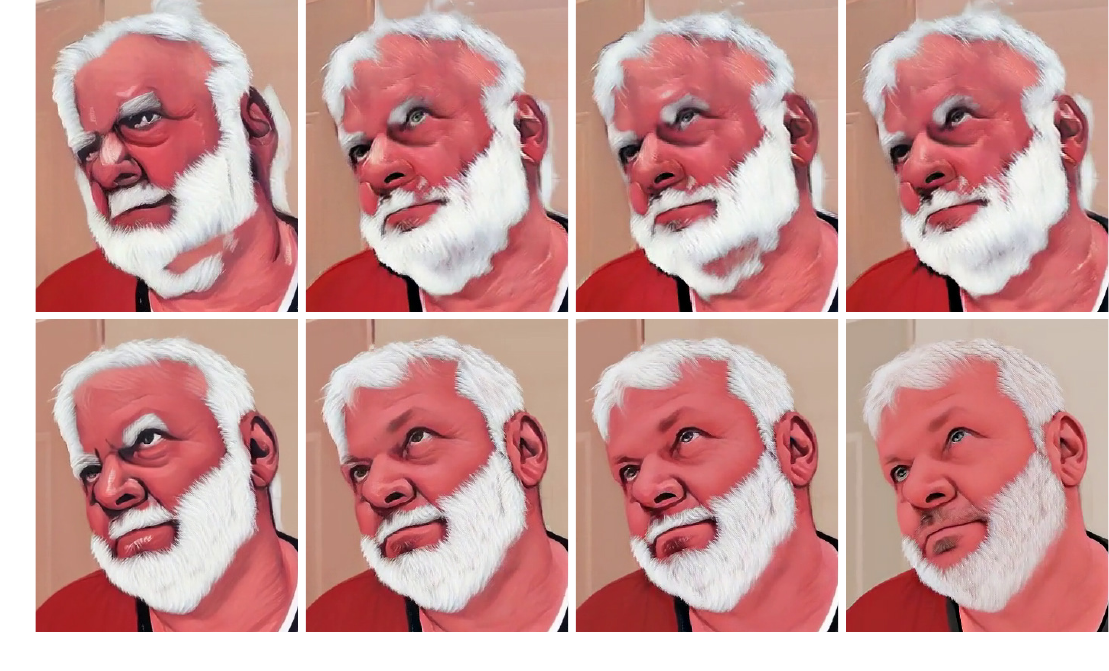}}%
    \put(0.01,0.17596831){\rotatebox{90}{\makebox(0,0)[t]{\lineheight{1.25}\smash{\begin{tabular}[t]{c}$\mathcal{L}_{\text{structure}}$\end{tabular}}}}}%
    \put(0.01,0.46426909){\rotatebox{90}{\makebox(0,0)[t]{\lineheight{1.25}\smash{\begin{tabular}[t]{c}$\mathcal{L}_{\text{lineart}}$\end{tabular}}}}}%
    \put(0.51723739,0.0){\makebox(0,0)[t]{\lineheight{1.25}\smash{\begin{tabular}[t]{c}$\xrightarrow{\hspace*{17.8mm}}$ increasing strength of loss term $\xrightarrow{\hspace*{17.8mm}}$\end{tabular}}}}%
  \end{picture}%
\endgroup%

%% file: supp/supplementary.tex
\appendix
\renewcommand\thefigure{\thesection.\arabic{figure}}

\section{Additional Results and Experiments}

In this supplementary material, we first present additional results and comparisons with the current state-of-the-art in diffusion-based video stylization~(see Sec.~\ref{sec:additional-results}). Four new sequences will be provided (see Figures~\ref{fig:sota_comparison_maruska}, \ref{fig:sota_comparison_cersei}, \ref{fig:sota_comparison_mom}, and \ref{fig:sota_comparison_lynx}) and for two of them we also prepared a comparison with keyframe-based methods~(see Figures~\ref{fig:sota_comparison_cersei_2} and~\ref{fig:sota_comparison_lynx_2}). In addition, we compare our method with the approach of Kim et al.~\shortcite{kim_collaborative_2024} (see Fig.~\ref{fig:csd_comparison}). Finally, we present results of the real-time video call scenario originally proposed by Texler et al.~\shortcite{texler_interactive_2020} (see Sec.~\ref{sec:rt-scenario} and Fig.~\ref{fig:skype_zuzka}).


\subsection{Additional results and comparisons}\label{sec:additional-results}

In this section, we present further comparisons of our method with state-of-the-art video stylization approaches.
The results of four additional sequences compared with diffusion-based methods are presented in Figures~\ref{fig:sota_comparison_maruska}, \ref{fig:sota_comparison_cersei}, \ref{fig:sota_comparison_mom}, and \ref{fig:sota_comparison_lynx}.
Two sequences are compared with keyframe-based methods in Figures~\ref{fig:sota_comparison_cersei_2} and \ref{fig:sota_comparison_lynx_2}.
In both cases, additional results support the discussion provided in the main paper (Sec.~4), i.e., it is visible that our approach maintains structural details better than diffusion-based as well as keyframe-based techniques and provides overall stability that was previously difficult to achieve using diffusion-based methods (see our supplementary video).

\begin{figure*}
\def\svgwidth{\hsize}\import{figures/sota_comparison_maruska/}{sota_comparison_maruska.tex}\caption{%
Results of our approach in comparison with the state-of-the-art in diffusion-based video stylization: The target video sequence (see a representative target frame~\targetframefigure{}) has been stylized using diffusion-based approaches (top row):  
(a)~Ceylan et al.~\shortcite{ceylan_pix2video_2023}, (b)~Yang et al.~\shortcite{yang_rerender_2023}, (c)~Chu et al.~\shortcite{chu_medm_2024}, and (d)~Geyer et al.~\shortcite{geyer_tokenflow_2024}. One frame from those stylized sequences was used as a keyframe (see small insets). The style of this keyframe has been propagated to the rest of the target sequence~$\textbf{y} \in \mathcal{Y}$ using our approach~(bottom row). Note how our approach better preserves structural details seen in the target frame. Also, see our supplementary video to compare consistency across the entire sequence. Diffusion-based approaches tend to suffer from notable structural flicker, whereas our approach keeps the structure consistent, yielding considerably more stable results.
}\label{fig:sota_comparison_maruska}
\end{figure*}

\begin{figure*}
\def\svgwidth{\hsize}\import{figures/sota_comparison_cersei/}{sota_comparison_cersei.tex}\caption{%
Results of our approach in comparison with the state-of-the-art in diffusion-based video stylization (cont.): See Fig.~\ref{fig:sota_comparison_maruska} for detailed explanation.
}\label{fig:sota_comparison_cersei}
\end{figure*}

\begin{figure*}
\def\svgwidth{\hsize}\import{figures/sota_comparison_mom/}{sota_comparison_mom.tex}\caption{%
Results of our approach in comparison with the state-of-the-art in diffusion-based video stylization (cont.): See Fig.~\ref{fig:sota_comparison_maruska} for detailed explanation.
}\label{fig:sota_comparison_mom}
\end{figure*}

\begin{figure*}
\def\svgwidth{\hsize}\import{figures/sota_comparison_lynx/}{sota_comparison_lynx.tex}\caption{%
Results of our approach in comparison with the state-of-the-art in diffusion-based video stylization (cont.): See Fig.~\ref{fig:sota_comparison_maruska} for detailed explanation.
}\label{fig:sota_comparison_lynx}
\end{figure*}

\begin{figure*}
\def\svgwidth{\hsize}\import{figures/sota_comparison_cersei_2/}{sota_comparison_cersei_2.tex}\caption{%
Results of our approach in comparison with the state-of-the-art in keyframe-based video stylization: The diffusion-based method of Yang et al.~\shortcite{yang_rerender_2023} has been used to generate a stylized sequence~(a) from which~$4$~keyframes were selected to perform video stylization using methods of Jamri\v{s}ka et al.~\shortcite{jamriska_stylizing_2019}~(b) and~Texler et al.~\shortcite{texler_interactive_2020}~(c), and~$1$ keyframe was selected for the method of Futschik et al.~\shortcite{futschik_stalp_2021}~(d) and our approach~(e). Note how our approach better preserves the structural details seen in the target frame. In our supplementary video, it is also visible that our approach keeps the structure consistent, yielding notably more stable results.
}\label{fig:sota_comparison_cersei_2} 
\end{figure*}

\begin{figure*}
\def\svgwidth{\hsize}\import{figures/sota_comparison_lynx_2/}{sota_comparison_lynx_2.tex}\caption{%
Results of our approach in comparison with the state-of-the-art in keyframe-based video stylization: Diffusion-based method of Geyer et al.~\shortcite{geyer_tokenflow_2024} has been used to generate a stylized sequence~(a).
See Fig.~\ref{fig:sota_comparison_cersei_2} for detailed explanation.
}\label{fig:sota_comparison_lynx_2}
\end{figure*}

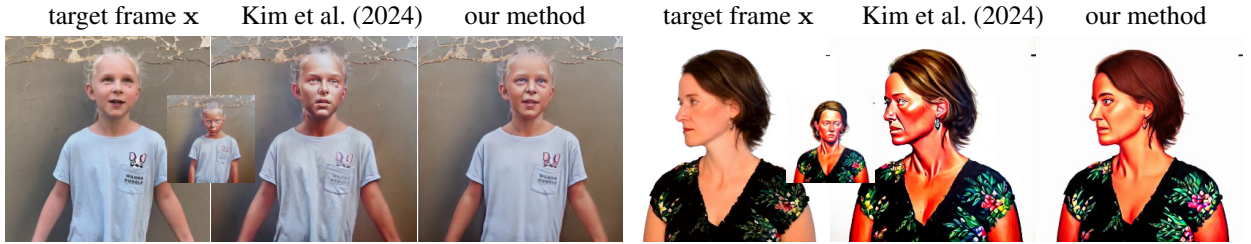
\begin{figure*}
\def\svgwidth{\hsize}\import{figures/csd_comparison/}{csd_comparison.tex}\caption{Comparison with the state-of-the-art in keyframe-based video stylization: Diffusion-based method of Kim et al.~\shortcite{kim_collaborative_2024} was used to generate a stylized sequence with the edit prompt: ``hyperrealistic detailed oil painting of a girl/woman.'' From that sequence, $1$~keyframe was selected (see small insets) to perform video stylization using our method. Note how our approach better preserves structural details seen in the target frame. In our supplementary video, you can see how our method outperforms the approach of Kim et al.~with respect to the overall structural stability.}\label{fig:csd_comparison}
\end{figure*}

In Fig.~\ref{fig:csd_comparison}, we present a comparison between our approach and the method of~Kim et al.~\shortcite{kim_collaborative_2024}. In the case of the sequence shown in the left half of the figure, our approach is better aligned with the structure of the target frame~$\mathbf{y}$. For the sequence in the right half of the figure, the result is not as distinctive, therefore we ask the reader to see our supplementary video where it is visible that our method clearly produces more temporally stable results.

We would like disambiguate the distinction between the conditioning used in our method and the one used by Kim et al.~\shortcite{kim_collaborative_2024}. In our method, we use the prior of a pre-trained text-to-image network conditioned by the guidance image~$\mathbf{c}$ to regularize the loss term~$\mathcal{L}_\text{style}$~(2). By regularizing this term, we enforce the structures from the target frame~$\mathbf{y}$ into the stylized frame~$\hat{\mathbf{y}}$. We sample the prior of the pre-trained network in every training step and just at a single step~$t$ of its denoising schedule. We do not use text conditioning. In contrast, the work of Kim et al.~\shortcite{kim_collaborative_2024} is built on the method of Brooks et al.~\shortcite{brooks_instructpix2pix_2023} that uses two-fold conditioning: the input image and the text instruction. Each conditioning has a guidance scale with which the degree of similarity between the generated samples and the input image is balanced, as well as the degree of similarity with the editing instruction. Both conditions are used during the entire image generation process. 
The method of Kim et al.~\shortcite{kim_collaborative_2024} has a significant memory footprint. Therefore, only~$90$ stylized frames are available for each sequence featured in our supplementary video.

\subsection{Real-time video call scenario}\label{sec:rt-scenario}

\begin{figure*}
\def\svgwidth{\hsize}\import{figures/skype_zuzka/}{skype_zuzka.tex}\caption{Our approach applied in the real-time video call scenario originally proposed by~Texler et al.~\shortcite{texler_interactive_2020}~(a) and later improved by Futschik et al.~\shortcite{futschik_stalp_2021}~(b) in comparison with three different settings of our approach: (c)~emphasizing style features from the stylized keyframe, (d)~achieving a balance between style and structure, and (e)~prioritizing structure from the target frame. All methods were trained with a single keyframe~(left). Note how our approach better preserves the structural details seen in the target frame for all three settings and also has better overall structural stability over the entire sequences as is visible in our supplementary video. The training time of our approach is comparable to the method of Futschik et al.~\shortcite{futschik_stalp_2021}.}\label{fig:skype_zuzka}
\end{figure*}
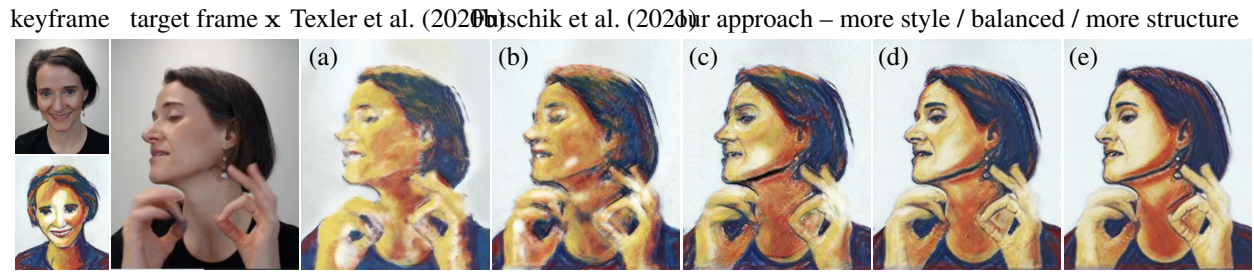

Texler et al.~\shortcite{texler_interactive_2020} proposed a real-time video call scenario in which the appearance of the call participant is stylized in real time. In this scenario, only a single artist-made keyframe is used to train our method and the methods of Texler et al.~\shortcite{texler_interactive_2020} and Futschik et al.~\shortcite{futschik_stalp_2021}. In Fig.~\ref{fig:skype_zuzka}, we observe a key advantage of our method: By increasing the strength of the loss term~$\mathcal{L}_\text{structure}$, the structure of the target frame~$\mathbf{y}$ is prioritized, and by decreasing its strength, the style characteristics of the style exemplar are emphasized. Compared to the results of Texler et al.~\shortcite{texler_interactive_2020}~(a) and Futschik et al.~\shortcite{futschik_stalp_2021}~(b), our proposed method produces a stylization that adheres better to the structural elements present in the target frame~$\mathbf{y}$. To emphasize this fact, we provide three different settings: (c)~emphasize more style features of the stylized keyframe, (d)~achieve a balance between style and structure, and (e)~prioritize structural details present in the target frame.

%% file: figures/sota_comparison_maruska/sota_comparison_maruska.tex
\begingroup%
  \makeatletter%
  \providecommand\color[2][]{%
    \errmessage{(Inkscape) Color is used for the text in Inkscape, but the package 'color.sty' is not loaded}%
    \renewcommand\color[2][]{}%
  }%
  \providecommand\transparent[1]{%
    \errmessage{(Inkscape) Transparency is used (non-zero) for the text in Inkscape, but the package 'transparent.sty' is not loaded}%
    \renewcommand\transparent[1]{}%
  }%
  \providecommand\rotatebox[2]{#2}%
  \newcommand*\fsize{\dimexpr\f@size pt\relax}%
  \newcommand*\lineheight[1]{\fontsize{\fsize}{#1\fsize}\selectfont}%
  \ifx\svgwidth\undefined%
    \setlength{\unitlength}{373.57923132bp}%
    \ifx\svgscale\undefined%
      \relax%
    \else%
      \setlength{\unitlength}{\unitlength * \real{\svgscale}}%
    \fi%
  \else%
    \setlength{\unitlength}{\svgwidth}%
  \fi%
  \global\let\svgwidth\undefined%
  \global\let\svgscale\undefined%
  \makeatother%
  \begin{picture}(1,0.41499197)%
    \lineheight{1}%
    \setlength\tabcolsep{0pt}%
    \put(0,0){\includegraphics[width=\unitlength,page=1]{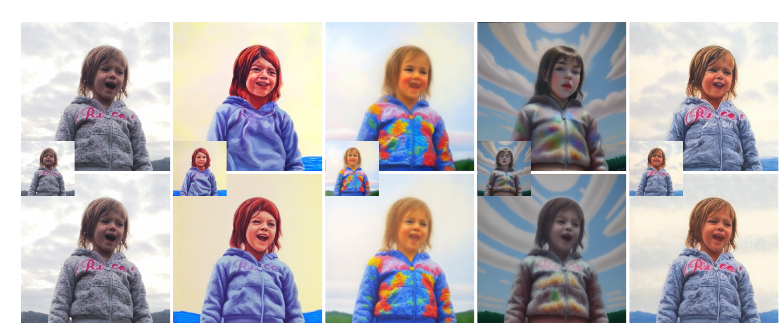}}%
    \put(0.12285387,0.39465448){\color[rgb]{0,0,0}\makebox(0,0)[t]{\lineheight{1.25}\smash{\begin{tabular}[t]{c}\targetframefigure{}\end{tabular}}}}%
    \put(0.31869846,0.39465448){\color[rgb]{0,0,0}\makebox(0,0)[t]{\lineheight{1.25}\smash{\begin{tabular}[t]{c}(a)~Ceylan et al.~\shortcite{ceylan_pix2video_2023}\end{tabular}}}}%
    \put(0.51308566,0.39465448){\color[rgb]{0,0,0}\makebox(0,0)[t]{\lineheight{1.25}\smash{\begin{tabular}[t]{c}(b)~Yang et al.~\shortcite{yang_rerender_2023}\end{tabular}}}}%
    \put(0.70863618,0.39465448){\color[rgb]{0,0,0}\makebox(0,0)[t]{\lineheight{1.25}\smash{\begin{tabular}[t]{c}(c)~Chu et al.~\shortcite{chu_medm_2024}\end{tabular}}}}%
    \put(0.90482707,0.39465448){\color[rgb]{0,0,0}\makebox(0,0)[t]{\lineheight{1.25}\smash{\begin{tabular}[t]{c}(d)~Geyer et al.~\shortcite{geyer_tokenflow_2024}\end{tabular}}}}%
    \put(0.02046819,0.29129200){\color[rgb]{0,0,0}\rotatebox{90}{\makebox(0,0)[t]{\lineheight{1.25}\smash{\begin{tabular}[t]{c}target frame\end{tabular}}}}}%
    \put(0.02046819,0.09614012){\color[rgb]{0,0,0}\rotatebox{90}{\makebox(0,0)[t]{\lineheight{1.25}\smash{\begin{tabular}[t]{c}our approach\end{tabular}}}}}%
  \end{picture}%
\endgroup%

%% file: figures/sota_comparison_cersei/sota_comparison_cersei.tex
\begingroup%
  \makeatletter%
  \providecommand\color[2][]{%
    \errmessage{(Inkscape) Color is used for the text in Inkscape, but the package 'color.sty' is not loaded}%
    \renewcommand\color[2][]{}%
  }%
  \providecommand\transparent[1]{%
    \errmessage{(Inkscape) Transparency is used (non-zero) for the text in Inkscape, but the package 'transparent.sty' is not loaded}%
    \renewcommand\transparent[1]{}%
  }%
  \providecommand\rotatebox[2]{#2}%
  \newcommand*\fsize{\dimexpr\f@size pt\relax}%
  \newcommand*\lineheight[1]{\fontsize{\fsize}{#1\fsize}\selectfont}%
  \ifx\svgwidth\undefined%
    \setlength{\unitlength}{373.57923132bp}%
    \ifx\svgscale\undefined%
      \relax%
    \else%
      \setlength{\unitlength}{\unitlength * \real{\svgscale}}%
    \fi%
  \else%
    \setlength{\unitlength}{\svgwidth}%
  \fi%
  \global\let\svgwidth\undefined%
  \global\let\svgscale\undefined%
  \makeatother%
  \begin{picture}(1,0.41499197)%
    \lineheight{1}%
    \setlength\tabcolsep{0pt}%
    \put(0,0){\includegraphics[width=\unitlength,page=1]{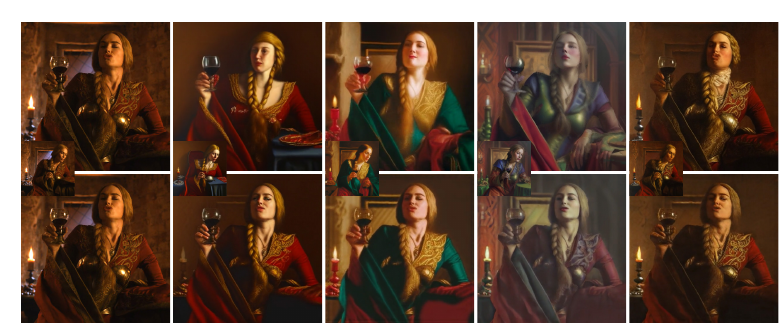}}%
    \put(0.12285387,0.39465448){\color[rgb]{0,0,0}\makebox(0,0)[t]{\lineheight{1.25}\smash{\begin{tabular}[t]{c}\targetframefigure{}\end{tabular}}}}%
    \put(0.31869846,0.39465448){\color[rgb]{0,0,0}\makebox(0,0)[t]{\lineheight{1.25}\smash{\begin{tabular}[t]{c}(a)~Ceylan et al.~\shortcite{ceylan_pix2video_2023}\end{tabular}}}}%
    \put(0.51308566,0.39465448){\color[rgb]{0,0,0}\makebox(0,0)[t]{\lineheight{1.25}\smash{\begin{tabular}[t]{c}(b)~Yang et al.~\shortcite{yang_rerender_2023}\end{tabular}}}}%
    \put(0.70863618,0.39465448){\color[rgb]{0,0,0}\makebox(0,0)[t]{\lineheight{1.25}\smash{\begin{tabular}[t]{c}(c)~Chu et al.~\shortcite{chu_medm_2024}\end{tabular}}}}%
    \put(0.90482707,0.39465448){\color[rgb]{0,0,0}\makebox(0,0)[t]{\lineheight{1.25}\smash{\begin{tabular}[t]{c}(d)~Geyer et al.~\shortcite{geyer_tokenflow_2024}\end{tabular}}}}%
    \put(0.02046819,0.29129200){\color[rgb]{0,0,0}\rotatebox{90}{\makebox(0,0)[t]{\lineheight{1.25}\smash{\begin{tabular}[t]{c}target frame\end{tabular}}}}}%
    \put(0.02046819,0.09614012){\color[rgb]{0,0,0}\rotatebox{90}{\makebox(0,0)[t]{\lineheight{1.25}\smash{\begin{tabular}[t]{c}our approach\end{tabular}}}}}%
  \end{picture}%
\endgroup%

%% file: figures/sota_comparison_mom/sota_comparison_mom.tex
\begingroup%
  \makeatletter%
  \providecommand\color[2][]{%
    \errmessage{(Inkscape) Color is used for the text in Inkscape, but the package 'color.sty' is not loaded}%
    \renewcommand\color[2][]{}%
  }%
  \providecommand\transparent[1]{%
    \errmessage{(Inkscape) Transparency is used (non-zero) for the text in Inkscape, but the package 'transparent.sty' is not loaded}%
    \renewcommand\transparent[1]{}%
  }%
  \providecommand\rotatebox[2]{#2}%
  \newcommand*\fsize{\dimexpr\f@size pt\relax}%
  \newcommand*\lineheight[1]{\fontsize{\fsize}{#1\fsize}\selectfont}%
  \ifx\svgwidth\undefined%
    \setlength{\unitlength}{373.57923132bp}%
    \ifx\svgscale\undefined%
      \relax%
    \else%
      \setlength{\unitlength}{\unitlength * \real{\svgscale}}%
    \fi%
  \else%
    \setlength{\unitlength}{\svgwidth}%
  \fi%
  \global\let\svgwidth\undefined%
  \global\let\svgscale\undefined%
  \makeatother%
  \begin{picture}(1,0.41499197)%
    \lineheight{1}%
    \setlength\tabcolsep{0pt}%
    \put(0,0){\includegraphics[width=\unitlength,page=1]{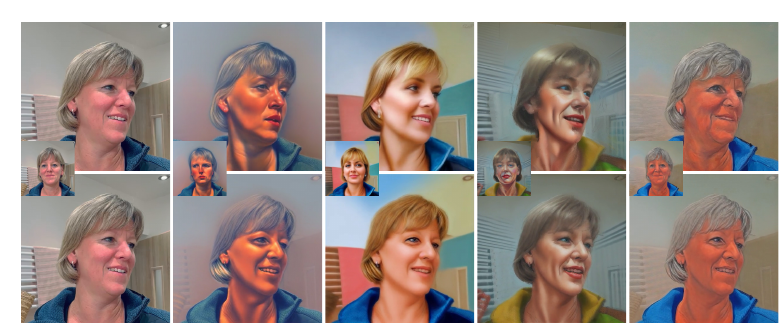}}%
    \put(0.12285387,0.39465448){\color[rgb]{0,0,0}\makebox(0,0)[t]{\lineheight{1.25}\smash{\begin{tabular}[t]{c}\targetframefigure{}\end{tabular}}}}%
    \put(0.31869846,0.39465448){\color[rgb]{0,0,0}\makebox(0,0)[t]{\lineheight{1.25}\smash{\begin{tabular}[t]{c}(a)~Ceylan et al.~\shortcite{ceylan_pix2video_2023}\end{tabular}}}}%
    \put(0.51308566,0.39465448){\color[rgb]{0,0,0}\makebox(0,0)[t]{\lineheight{1.25}\smash{\begin{tabular}[t]{c}(b)~Yang et al.~\shortcite{yang_rerender_2023}\end{tabular}}}}%
    \put(0.70863618,0.39465448){\color[rgb]{0,0,0}\makebox(0,0)[t]{\lineheight{1.25}\smash{\begin{tabular}[t]{c}(c)~Chu et al.~\shortcite{chu_medm_2024}\end{tabular}}}}%
    \put(0.90482707,0.39465448){\color[rgb]{0,0,0}\makebox(0,0)[t]{\lineheight{1.25}\smash{\begin{tabular}[t]{c}(d)~Geyer et al.~\shortcite{geyer_tokenflow_2024}\end{tabular}}}}%
    \put(0.02046819,0.29129200){\color[rgb]{0,0,0}\rotatebox{90}{\makebox(0,0)[t]{\lineheight{1.25}\smash{\begin{tabular}[t]{c}target frame\end{tabular}}}}}%
    \put(0.02046819,0.09614012){\color[rgb]{0,0,0}\rotatebox{90}{\makebox(0,0)[t]{\lineheight{1.25}\smash{\begin{tabular}[t]{c}our approach\end{tabular}}}}}%
  \end{picture}%
\endgroup%

%% file: figures/sota_comparison_lynx/sota_comparison_lynx.tex
\begingroup%
  \makeatletter%
  \providecommand\color[2][]{%
    \errmessage{(Inkscape) Color is used for the text in Inkscape, but the package 'color.sty' is not loaded}%
    \renewcommand\color[2][]{}%
  }%
  \providecommand\transparent[1]{%
    \errmessage{(Inkscape) Transparency is used (non-zero) for the text in Inkscape, but the package 'transparent.sty' is not loaded}%
    \renewcommand\transparent[1]{}%
  }%
  \providecommand\rotatebox[2]{#2}%
  \newcommand*\fsize{\dimexpr\f@size pt\relax}%
  \newcommand*\lineheight[1]{\fontsize{\fsize}{#1\fsize}\selectfont}%
  \ifx\svgwidth\undefined%
    \setlength{\unitlength}{373.57923132bp}%
    \ifx\svgscale\undefined%
      \relax%
    \else%
      \setlength{\unitlength}{\unitlength * \real{\svgscale}}%
    \fi%
  \else%
    \setlength{\unitlength}{\svgwidth}%
  \fi%
  \global\let\svgwidth\undefined%
  \global\let\svgscale\undefined%
  \makeatother%
  \begin{picture}(1,0.41499197)%
    \lineheight{1}%
    \setlength\tabcolsep{0pt}%
    \put(0,0){\includegraphics[width=\unitlength,page=1]{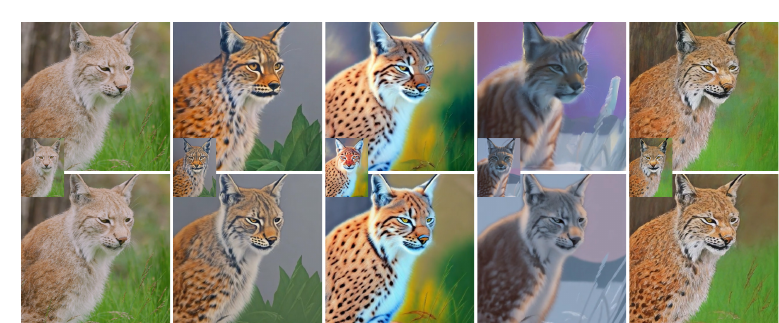}}%
    \put(0.12285387,0.39465448){\color[rgb]{0,0,0}\makebox(0,0)[t]{\lineheight{1.25}\smash{\begin{tabular}[t]{c}\targetframefigure{}\end{tabular}}}}%
    \put(0.31869846,0.39465448){\color[rgb]{0,0,0}\makebox(0,0)[t]{\lineheight{1.25}\smash{\begin{tabular}[t]{c}(a)~Ceylan et al.~\shortcite{ceylan_pix2video_2023}\end{tabular}}}}%
    \put(0.51308566,0.39465448){\color[rgb]{0,0,0}\makebox(0,0)[t]{\lineheight{1.25}\smash{\begin{tabular}[t]{c}(b)~Yang et al.~\shortcite{yang_rerender_2023}\end{tabular}}}}%
    \put(0.70863618,0.39465448){\color[rgb]{0,0,0}\makebox(0,0)[t]{\lineheight{1.25}\smash{\begin{tabular}[t]{c}(c)~Chu et al.~\shortcite{chu_medm_2024}\end{tabular}}}}%
    \put(0.90482707,0.39465448){\color[rgb]{0,0,0}\makebox(0,0)[t]{\lineheight{1.25}\smash{\begin{tabular}[t]{c}(d)~Geyer et al.~\shortcite{geyer_tokenflow_2024}\end{tabular}}}}%
    \put(0.02046819,0.29129200){\color[rgb]{0,0,0}\rotatebox{90}{\makebox(0,0)[t]{\lineheight{1.25}\smash{\begin{tabular}[t]{c}target frame\end{tabular}}}}}%
    \put(0.02046819,0.09614012){\color[rgb]{0,0,0}\rotatebox{90}{\makebox(0,0)[t]{\lineheight{1.25}\smash{\begin{tabular}[t]{c}our approach\end{tabular}}}}}%
  \end{picture}%
\endgroup%

%% file: figures/sota_comparison_cersei_2/sota_comparison_cersei_2.tex
\begingroup%
  \makeatletter%
  \providecommand\color[2][]{%
    \errmessage{(Inkscape) Color is used for the text in Inkscape, but the package 'color.sty' is not loaded}%
    \renewcommand\color[2][]{}%
  }%
  \providecommand\transparent[1]{%
    \errmessage{(Inkscape) Transparency is used (non-zero) for the text in Inkscape, but the package 'transparent.sty' is not loaded}%
    \renewcommand\transparent[1]{}%
  }%
  \providecommand\rotatebox[2]{#2}%
  \newcommand*\fsize{\dimexpr\f@size pt\relax}%
  \newcommand*\lineheight[1]{\fontsize{\fsize}{#1\fsize}\selectfont}%
  \ifx\svgwidth\undefined%
    \setlength{\unitlength}{485.03142109bp}%
    \ifx\svgscale\undefined%
      \relax%
    \else%
      \setlength{\unitlength}{\unitlength * \real{\svgscale}}%
    \fi%
  \else%
    \setlength{\unitlength}{\svgwidth}%
  \fi%
  \global\let\svgwidth\undefined%
  \global\let\svgscale\undefined%
  \makeatother%
  \begin{picture}(1,0.44150126)%
    \lineheight{1}%
    \setlength\tabcolsep{0pt}%
    \put(0,0){\includegraphics[width=\unitlength,page=1]{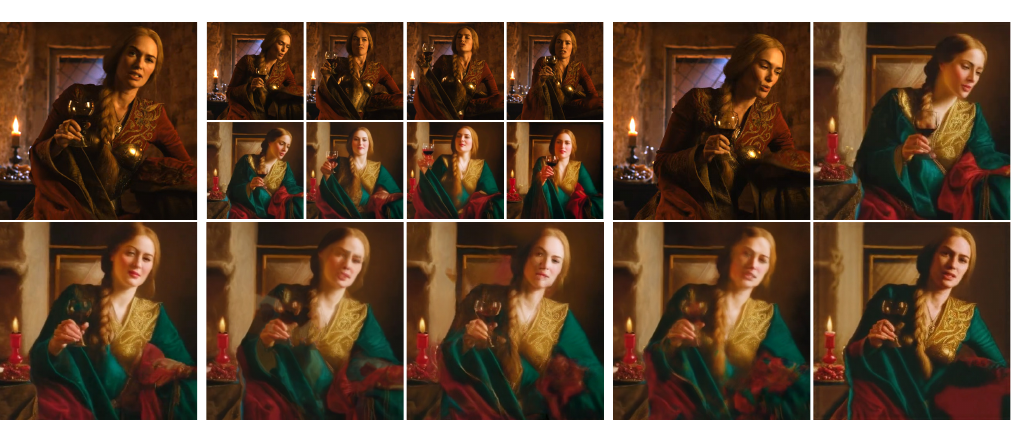}}%
    \put(0.09771721,0.42573636){\makebox(0,0)[t]{\lineheight{1.25}\smash{\begin{tabular}[t]{c}target frame \targetframefigure{}\end{tabular}}}}%
    \put(0.40079691,0.42573636){\makebox(0,0)[t]{\lineheight{1.25}\smash{\begin{tabular}[t]{c}4 keyframes\end{tabular}}}}%
    \put(0.80293356,0.42573636){\makebox(0,0)[t]{\lineheight{1.25}\smash{\begin{tabular}[t]{c}1 keyframe\end{tabular}}}}%
    \put(0.09647897,0.00414757){\makebox(0,0)[t]{\lineheight{1.25}\smash{\begin{tabular}[t]{c}(a)~Yang et al.~\shortcite{yang_rerender_2023}\end{tabular}}}}%
    \put(0.30173980,0.00414757){\makebox(0,0)[t]{\lineheight{1.25}\smash{\begin{tabular}[t]{c}(b)~Jamri\v{s}ka et al.~\shortcite{jamriska_stylizing_2019}\end{tabular}}}}%
    \put(0.50025673,0.00414757){\makebox(0,0)[t]{\lineheight{1.25}\smash{\begin{tabular}[t]{c}(c)~Texler et al.~\shortcite{texler_interactive_2020}\end{tabular}}}}%
    \put(0.70396224,0.00414757){\makebox(0,0)[t]{\lineheight{1.25}\smash{\begin{tabular}[t]{c}(d)~Futschik et al.~\shortcite{futschik_stalp_2021}\end{tabular}}}}%
    \put(0.90278594,0.00414757){\makebox(0,0)[t]{\lineheight{1.25}\smash{\begin{tabular}[t]{c}(e)~our approach\end{tabular}}}}%
  \end{picture}%
\endgroup%

%% file: figures/sota_comparison_lynx_2/sota_comparison_lynx_2.tex
\begingroup%
  \makeatletter%
  \providecommand\color[2][]{%
    \errmessage{(Inkscape) Color is used for the text in Inkscape, but the package 'color.sty' is not loaded}%
    \renewcommand\color[2][]{}%
  }%
  \providecommand\transparent[1]{%
    \errmessage{(Inkscape) Transparency is used (non-zero) for the text in Inkscape, but the package 'transparent.sty' is not loaded}%
    \renewcommand\transparent[1]{}%
  }%
  \providecommand\rotatebox[2]{#2}%
  \newcommand*\fsize{\dimexpr\f@size pt\relax}%
  \newcommand*\lineheight[1]{\fontsize{\fsize}{#1\fsize}\selectfont}%
  \ifx\svgwidth\undefined%
    \setlength{\unitlength}{485.03142109bp}%
    \ifx\svgscale\undefined%
      \relax%
    \else%
      \setlength{\unitlength}{\unitlength * \real{\svgscale}}%
    \fi%
  \else%
    \setlength{\unitlength}{\svgwidth}%
  \fi%
  \global\let\svgwidth\undefined%
  \global\let\svgscale\undefined%
  \makeatother%
  \begin{picture}(1,0.44150126)%
    \lineheight{1}%
    \setlength\tabcolsep{0pt}%
    \put(0,0){\includegraphics[width=\unitlength,page=1]{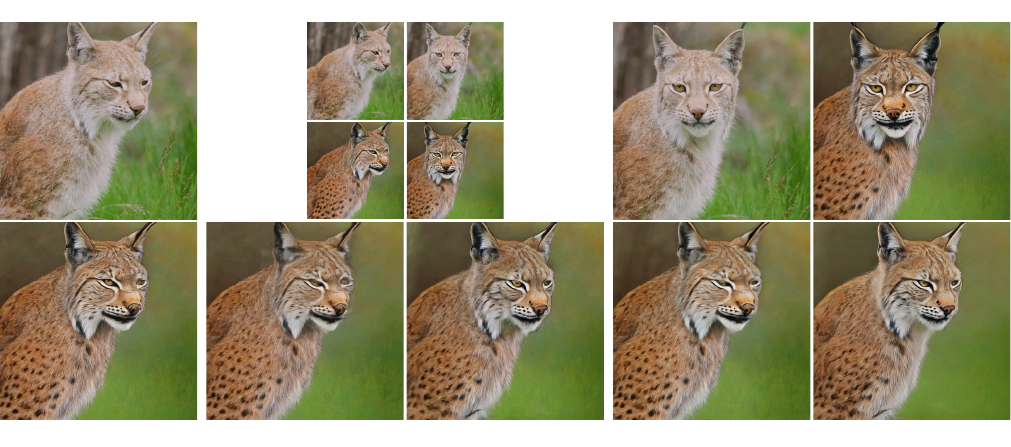}}%
    \put(0.09771721,0.42573636){\makebox(0,0)[t]{\lineheight{1.25}\smash{\begin{tabular}[t]{c}target frame $\mathbf{x}$\end{tabular}}}}%
    \put(0.40079691,0.42573636){\makebox(0,0)[t]{\lineheight{1.25}\smash{\begin{tabular}[t]{c}2 keyframes\end{tabular}}}}%
    \put(0.80293356,0.42573636){\makebox(0,0)[t]{\lineheight{1.25}\smash{\begin{tabular}[t]{c}1 keyframe\end{tabular}}}}%
    \put(0.09647897,0.00414757){\makebox(0,0)[t]{\lineheight{1.25}\smash{\begin{tabular}[t]{c}(a)~Geyer et al.~\shortcite{geyer_tokenflow_2024}\end{tabular}}}}%
    \put(0.30173980,0.00414757){\makebox(0,0)[t]{\lineheight{1.25}\smash{\begin{tabular}[t]{c}(b)~Jamri\v{s}ka et al.~\shortcite{jamriska_stylizing_2019}\end{tabular}}}}%
    \put(0.50025673,0.00414757){\makebox(0,0)[t]{\lineheight{1.25}\smash{\begin{tabular}[t]{c}(c)~Texler et al.~\shortcite{texler_interactive_2020}\end{tabular}}}}%
    \put(0.70396224,0.00414757){\makebox(0,0)[t]{\lineheight{1.25}\smash{\begin{tabular}[t]{c}(d)~Futschik et al.~\shortcite{futschik_stalp_2021}\end{tabular}}}}%
    \put(0.90278594,0.00414757){\makebox(0,0)[t]{\lineheight{1.25}\smash{\begin{tabular}[t]{c}(e)~our approach\end{tabular}}}}%
  \end{picture}%
\endgroup%

%% file: figures/csd_comparison/csd_comparison.tex
\begingroup%
  \makeatletter%
  \providecommand\color[2][]{%
    \errmessage{(Inkscape) Color is used for the text in Inkscape, but the package 'color.sty' is not loaded}%
    \renewcommand\color[2][]{}%
  }%
  \providecommand\transparent[1]{%
    \errmessage{(Inkscape) Transparency is used (non-zero) for the text in Inkscape, but the package 'transparent.sty' is not loaded}%
    \renewcommand\transparent[1]{}%
  }%
  \providecommand\rotatebox[2]{#2}%
  \newcommand*\fsize{\dimexpr\f@size pt\relax}%
  \newcommand*\lineheight[1]{\fontsize{\fsize}{#1\fsize}\selectfont}%
  \ifx\svgwidth\undefined%
    \setlength{\unitlength}{531.32417454bp}%
    \ifx\svgscale\undefined%
      \relax%
    \else%
      \setlength{\unitlength}{\unitlength * \real{\svgscale}}%
    \fi%
  \else%
    \setlength{\unitlength}{\svgwidth}%
  \fi%
  \global\let\svgwidth\undefined%
  \global\let\svgscale\undefined%
  \makeatother%
  \begin{picture}(1,0.195)%
    \lineheight{1}%
    \setlength\tabcolsep{0pt}%
    \put(0,0){\includegraphics[width=\unitlength,page=1]{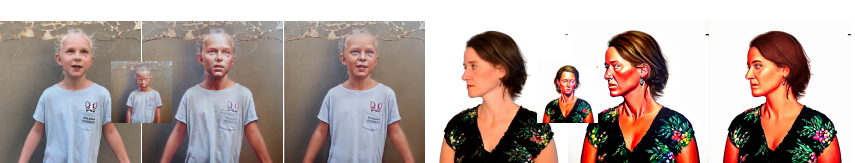}}%
    \put(0.035,0.175){\color[rgb]{0,0,0}\makebox(0,0)[lt]{\lineheight{1.25}\smash{\begin{tabular}[t]{l}target frame $\mathbf{x}$\end{tabular}}}}%
    \put(0.19,0.175){\color[rgb]{0,0,0}\makebox(0,0)[lt]{\lineheight{1.25}\smash{\begin{tabular}[t]{l}Kim et al.~\shortcite{kim_collaborative_2024}\end{tabular}}}}%
    \put(0.37,0.175){\color[rgb]{0,0,0}\makebox(0,0)[lt]{\lineheight{1.25}\smash{\begin{tabular}[t]{l}our method\end{tabular}}}}%
    \put(0.53,0.175){\color[rgb]{0,0,0}\makebox(0,0)[lt]{\lineheight{1.25}\smash{\begin{tabular}[t]{l}target frame $\mathbf{x}$\end{tabular}}}}%
    \put(0.69,0.175){\color[rgb]{0,0,0}\makebox(0,0)[lt]{\lineheight{1.25}\smash{\begin{tabular}[t]{l}Kim et al.~\shortcite{kim_collaborative_2024}\end{tabular}}}}%
    \put(0.87,0.175){\color[rgb]{0,0,0}\makebox(0,0)[lt]{\lineheight{1.25}\smash{\begin{tabular}[t]{l}our method\end{tabular}}}}%
  \end{picture}%
\endgroup%

%% file: figures/skype_zuzka/skype_zuzka.tex
\begingroup%
  \makeatletter%
  \providecommand\color[2][]{%
    \errmessage{(Inkscape) Color is used for the text in Inkscape, but the package 'color.sty' is not loaded}%
    \renewcommand\color[2][]{}%
  }%
  \providecommand\transparent[1]{%
    \errmessage{(Inkscape) Transparency is used (non-zero) for the text in Inkscape, but the package 'transparent.sty' is not loaded}%
    \renewcommand\transparent[1]{}%
  }%
  \providecommand\rotatebox[2]{#2}%
  \newcommand*\fsize{\dimexpr\f@size pt\relax}%
  \newcommand*\lineheight[1]{\fontsize{\fsize}{#1\fsize}\selectfont}%
  \ifx\svgwidth\undefined%
    \setlength{\unitlength}{483.89468216bp}%
    \ifx\svgscale\undefined%
      \relax%
    \else%
      \setlength{\unitlength}{\unitlength * \real{\svgscale}}%
    \fi%
  \else%
    \setlength{\unitlength}{\svgwidth}%
  \fi%
  \global\let\svgwidth\undefined%
  \global\let\svgscale\undefined%
  \makeatother%
  \begin{picture}(1,0.21635067)%
    \lineheight{1}%
    \setlength\tabcolsep{0pt}%
    \put(0,0){\includegraphics[width=\unitlength,page=1]{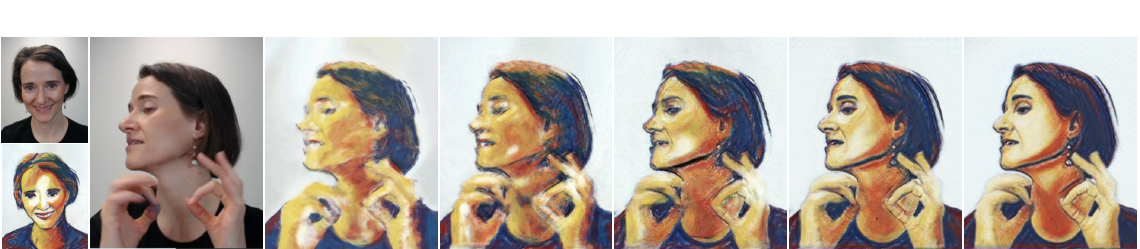}}%
    \put(0.038,0.195){\makebox(0,0)[t]{\lineheight{1.25}\smash{\begin{tabular}[t]{c}keyframe\end{tabular}}}}%
    %
    \put(0.155,0.195){\makebox(0,0)[t]{\lineheight{1.25}\smash{\begin{tabular}[t]{c}target frame $\mathbf{x}$\end{tabular}}}}%
    \put(0.31,0.195){\makebox(0,0)[t]{\lineheight{1.25}\smash{\begin{tabular}[t]{c}Texler et al.~\shortcite{texler_interactive_2020}\end{tabular}}}}%
    \put(0.46,0.195){\makebox(0,0)[t]{\lineheight{1.25}\smash{\begin{tabular}[t]{c}Futschik et al.~\shortcite{futschik_stalp_2021}\end{tabular}}}}%
    \put(0.760,0.195){\makebox(0,0)[t]{\lineheight{1.25}\smash{\begin{tabular}[t]{c}our approach -- more style / balanced / more structure\end{tabular}}}}%
    \put(0.25,0.165){\color[rgb]{0,0,0}\makebox(0,0)[t]{\lineheight{1.25}\smash{\begin{tabular}[t]{c}(a)\end{tabular}}}}%
    \put(0.403,0.165){\color[rgb]{0,0,0}\makebox(0,0)[t]{\lineheight{1.25}\smash{\begin{tabular}[t]{c}(b)\end{tabular}}}}%
    \put(0.556,0.165){\color[rgb]{0,0,0}\makebox(0,0)[t]{\lineheight{1.25}\smash{\begin{tabular}[t]{c}(c)\end{tabular}}}}%
    \put(0.709,0.165){\color[rgb]{0,0,0}\makebox(0,0)[t]{\lineheight{1.25}\smash{\begin{tabular}[t]{c}(d)\end{tabular}}}}%
    \put(0.862,0.165){\color[rgb]{0,0,0}\makebox(0,0)[t]{\lineheight{1.25}\smash{\begin{tabular}[t]{c}(e)\end{tabular}}}}%
  \end{picture}%
\endgroup%

%% file: stylereiser.bbl

%% file: stylereiser.bbl
\begin{thebibliography}{44}


\ifx \showCODEN    \undefined \def \showCODEN     #1{\unskip}     \fi
\ifx \showDOI      \undefined \def \showDOI       #1{#1}\fi
\ifx \showISBNx    \undefined \def \showISBNx     #1{\unskip}     \fi
\ifx \showISBNxiii \undefined \def \showISBNxiii  #1{\unskip}     \fi
\ifx \showISSN     \undefined \def \showISSN      #1{\unskip}     \fi
\ifx \showLCCN     \undefined \def \showLCCN      #1{\unskip}     \fi
\ifx \shownote     \undefined \def \shownote      #1{#1}          \fi
\ifx \showarticletitle \undefined \def \showarticletitle #1{#1}   \fi
\ifx \showURL      \undefined \def \showURL       {\relax}        \fi
\providecommand\bibfield[2]{#2}
\providecommand\bibinfo[2]{#2}
\providecommand\natexlab[1]{#1}
\providecommand\showeprint[2][]{arXiv:#2}

\bibitem[Bénard et~al\mbox{.}(2013)]%
        {benard_stylizing_2013}
\bibfield{author}{\bibinfo{person}{Pierre Bénard}, \bibinfo{person}{Forrester Cole}, \bibinfo{person}{Michael Kass}, \bibinfo{person}{Igor Mordatch}, \bibinfo{person}{James Hegarty}, \bibinfo{person}{Martin~Sebastian Senn}, \bibinfo{person}{Kurt~W Fleischer}, \bibinfo{person}{Davide Pesare}, {and} \bibinfo{person}{Katherine Breeden}.} \bibinfo{year}{2013}\natexlab{}.
\newblock \showarticletitle{Stylizing animation by example}.
\newblock \bibinfo{journal}{\emph{ACM Transactions on Graphics}} \bibinfo{volume}{32}, \bibinfo{number}{4} (\bibinfo{year}{2013}), \bibinfo{pages}{119}.
\newblock


\bibitem[Canny(1986)]%
        {canny_computational_1986}
\bibfield{author}{\bibinfo{person}{John Canny}.} \bibinfo{year}{1986}\natexlab{}.
\newblock \showarticletitle{A computational approach to edge detection}.
\newblock \bibinfo{journal}{\emph{IEEE Transactions on pattern analysis and machine intelligence}} \bibinfo{volume}{8}, \bibinfo{number}{6} (\bibinfo{year}{1986}), \bibinfo{pages}{679--698}.
\newblock


\bibitem[Ceylan et~al\mbox{.}(2023)]%
        {ceylan_pix2video_2023}
\bibfield{author}{\bibinfo{person}{Duygu Ceylan}, \bibinfo{person}{Chun-Hao~P Huang}, {and} \bibinfo{person}{Niloy~J Mitra}.} \bibinfo{year}{2023}\natexlab{}.
\newblock \showarticletitle{Pix2video: {Video} editing using image diffusion}. In \bibinfo{booktitle}{\emph{Proceedings of {IEEE} {Conference} on {Computer} {Vision} and {Pattern} {Recognition}}}. \bibinfo{pages}{23206--23217}.
\newblock


\bibitem[Chen et~al\mbox{.}(2017)]%
        {chen_coherent_2017}
\bibfield{author}{\bibinfo{person}{Dongdong Chen}, \bibinfo{person}{Jing Liao}, \bibinfo{person}{Lu Yuan}, \bibinfo{person}{Nenghai Yu}, {and} \bibinfo{person}{Gang Hua}.} \bibinfo{year}{2017}\natexlab{}.
\newblock \showarticletitle{Coherent online video style transfer}. In \bibinfo{booktitle}{\emph{Proceedings of {IEEE} {International} {Conference} on {Computer} {Vision}}}. \bibinfo{pages}{1105--1114}.
\newblock


\bibitem[Chu et~al\mbox{.}(2024)]%
        {chu_medm_2024}
\bibfield{author}{\bibinfo{person}{Ernie Chu}, \bibinfo{person}{Tzuhsuan Huang}, \bibinfo{person}{Shuo-Yen Lin}, {and} \bibinfo{person}{Jun-Cheng Chen}.} \bibinfo{year}{2024}\natexlab{}.
\newblock \showarticletitle{{MeDM}: {Mediating} image diffusion models for video-to-video translation with temporal correspondence guidance}. In \bibinfo{booktitle}{\emph{Proceedings of the {AAAI} {Conference} on {Artificial} {Intelligence}}}.
\newblock


\bibitem[Curtis et~al\mbox{.}(1997)]%
        {curtis_computer-generated_1997}
\bibfield{author}{\bibinfo{person}{Cassidy~J Curtis}, \bibinfo{person}{Sean~E Anderson}, \bibinfo{person}{Joshua~E Seims}, \bibinfo{person}{Kurt~W Fleischer}, {and} \bibinfo{person}{David~H Salesin}.} \bibinfo{year}{1997}\natexlab{}.
\newblock \showarticletitle{Computer-generated watercolor}. In \bibinfo{booktitle}{\emph{{SIGGRAPH} {Conference} {Proceedings}}}. \bibinfo{pages}{421--430}.
\newblock


\bibitem[Fišer et~al\mbox{.}(2017)]%
        {fiser_example-based_2017}
\bibfield{author}{\bibinfo{person}{Jakub Fišer}, \bibinfo{person}{Ondřej Jamriška}, \bibinfo{person}{David Simons}, \bibinfo{person}{Eli Shechtman}, \bibinfo{person}{Jingwan Lu}, \bibinfo{person}{Paul Asente}, \bibinfo{person}{Michal Lukáč}, {and} \bibinfo{person}{Daniel Sýkora}.} \bibinfo{year}{2017}\natexlab{}.
\newblock \showarticletitle{Example-{Based} {Synthesis} of {Stylized} {Facial} {Animations}}.
\newblock \bibinfo{journal}{\emph{ACM Transactions on Graphics}} \bibinfo{volume}{36}, \bibinfo{number}{4} (\bibinfo{year}{2017}), \bibinfo{pages}{155}.
\newblock


\bibitem[Frigo et~al\mbox{.}(2016)]%
        {frigo_split_2016}
\bibfield{author}{\bibinfo{person}{Oriel Frigo}, \bibinfo{person}{Neus Sabater}, \bibinfo{person}{Julie Delon}, {and} \bibinfo{person}{Pierre Hellier}.} \bibinfo{year}{2016}\natexlab{}.
\newblock \showarticletitle{Split and {Match}: {Example}-based adaptive patch sampling for unsupervised style transfer}. In \bibinfo{booktitle}{\emph{Proceedings of {IEEE} {Conference} on {Computer} {Vision} and {Pattern} {Recognition}}}. \bibinfo{pages}{553--561}.
\newblock


\bibitem[Futschik et~al\mbox{.}(2019)]%
        {futschik_real-time_2019}
\bibfield{author}{\bibinfo{person}{David Futschik}, \bibinfo{person}{Menglei Chai}, \bibinfo{person}{Chen Cao}, \bibinfo{person}{Chongyang Ma}, \bibinfo{person}{Aleksei Stoliar}, \bibinfo{person}{Sergey Korolev}, \bibinfo{person}{Sergey Tulyakov}, \bibinfo{person}{Michal Kučera}, {and} \bibinfo{person}{Daniel Sýkora}.} \bibinfo{year}{2019}\natexlab{}.
\newblock \showarticletitle{Real-time patch-based stylization of portraits using generative adversarial network.}. In \bibinfo{booktitle}{\emph{Proceedings of the {ACM}/{EG} {Expressive} {Symposium}}}. \bibinfo{pages}{33--42}.
\newblock


\bibitem[Futschik et~al\mbox{.}(2021)]%
        {futschik_stalp_2021}
\bibfield{author}{\bibinfo{person}{David Futschik}, \bibinfo{person}{Michal Kučera}, \bibinfo{person}{Michal Lukáč}, \bibinfo{person}{Zhaowen Wang}, \bibinfo{person}{Eli Shechtman}, {and} \bibinfo{person}{Daniel Sýkora}.} \bibinfo{year}{2021}\natexlab{}.
\newblock \showarticletitle{{STALP}: {Style} transfer with auxiliary limited pairing}.
\newblock \bibinfo{journal}{\emph{Computer Graphics Forum}} \bibinfo{volume}{40}, \bibinfo{number}{2} (\bibinfo{year}{2021}), \bibinfo{pages}{563--573}.
\newblock


\bibitem[Gatys et~al\mbox{.}(2016)]%
        {gatys_image_2016}
\bibfield{author}{\bibinfo{person}{Leon~A. Gatys}, \bibinfo{person}{Alexander~S. Ecker}, {and} \bibinfo{person}{Matthias Bethge}.} \bibinfo{year}{2016}\natexlab{}.
\newblock \showarticletitle{Image style transfer using convolutional neural networks}. In \bibinfo{booktitle}{\emph{Proceedings of {IEEE} {Conference} on {Computer} {Vision} and {Pattern} {Recognition}}}. \bibinfo{pages}{2414--2423}.
\newblock


\bibitem[Geyer et~al\mbox{.}(2024)]%
        {geyer_tokenflow_2024}
\bibfield{author}{\bibinfo{person}{Michal Geyer}, \bibinfo{person}{Omer Bar-Tal}, \bibinfo{person}{Shai Bagon}, {and} \bibinfo{person}{Tali Dekel}.} \bibinfo{year}{2024}\natexlab{}.
\newblock \showarticletitle{{TokenFlow}: {Consistent} diffusion features for consistent video editing}. In \bibinfo{booktitle}{\emph{Proceedings of {International} {Conference} on {Learning} {Representations}}}.
\newblock


\bibitem[Hertzmann et~al\mbox{.}(2001)]%
        {hertzmann_image_2001}
\bibfield{author}{\bibinfo{person}{Aaron Hertzmann}, \bibinfo{person}{Charles~E. Jacobs}, \bibinfo{person}{Nuria Oliver}, \bibinfo{person}{Brian Curless}, {and} \bibinfo{person}{David~H. Salesin}.} \bibinfo{year}{2001}\natexlab{}.
\newblock \showarticletitle{Image analogies}. In \bibinfo{booktitle}{\emph{{SIGGRAPH} {Conference} {Proceedings}}}. \bibinfo{pages}{327--340}.
\newblock


\bibitem[Ho et~al\mbox{.}(2020)]%
        {ho_denoising_2020}
\bibfield{author}{\bibinfo{person}{Jonathan Ho}, \bibinfo{person}{Ajay Jain}, {and} \bibinfo{person}{Pieter Abbeel}.} \bibinfo{year}{2020}\natexlab{}.
\newblock \showarticletitle{Denoising diffusion probabilistic models}. In \bibinfo{booktitle}{\emph{Advances in {Neural} {Information} {Processing} {Systems}}}. \bibinfo{pages}{6840--6851}.
\newblock


\bibitem[Isola et~al\mbox{.}(2017)]%
        {isola_image--image_2017}
\bibfield{author}{\bibinfo{person}{Phillip Isola}, \bibinfo{person}{Jun-Yan Zhu}, \bibinfo{person}{Tinghui Zhou}, {and} \bibinfo{person}{Alexei~A Efros}.} \bibinfo{year}{2017}\natexlab{}.
\newblock \showarticletitle{Image-to-image translation with conditional adversarial networks}. In \bibinfo{booktitle}{\emph{Proceedings of the {IEEE} conference on computer vision and pattern recognition}}. \bibinfo{pages}{1125--1134}.
\newblock


\bibitem[Jamriška et~al\mbox{.}(2019)]%
        {jamriska_stylizing_2019}
\bibfield{author}{\bibinfo{person}{Ondřej Jamriška}, \bibinfo{person}{Šárka Sochorová}, \bibinfo{person}{Ondřej Texler}, \bibinfo{person}{Michal Lukáč}, \bibinfo{person}{Jakub Fišer}, \bibinfo{person}{Jingwan Lu}, \bibinfo{person}{Eli Shechtman}, {and} \bibinfo{person}{Daniel Sýkora}.} \bibinfo{year}{2019}\natexlab{}.
\newblock \showarticletitle{Stylizing video by example}.
\newblock \bibinfo{journal}{\emph{ACM Transactions on Graphics}} \bibinfo{volume}{38}, \bibinfo{number}{4} (\bibinfo{year}{2019}), \bibinfo{pages}{107}.
\newblock


\bibitem[Johnson et~al\mbox{.}(2016)]%
        {johnson_perceptual_2016}
\bibfield{author}{\bibinfo{person}{Justin Johnson}, \bibinfo{person}{Alexandre Alahi}, {and} \bibinfo{person}{Li Fei-Fei}.} \bibinfo{year}{2016}\natexlab{}.
\newblock \showarticletitle{Perceptual losses for real-time style transfer and super-resolution}. In \bibinfo{booktitle}{\emph{Proceedings of {European} {Conference} on {Computer} {Vision}}}. \bibinfo{pages}{694--711}.
\newblock


\bibitem[Kasten et~al\mbox{.}(2021)]%
        {kasten_layered_2021}
\bibfield{author}{\bibinfo{person}{Yoni Kasten}, \bibinfo{person}{Dolev Ofri}, \bibinfo{person}{Oliver Wang}, {and} \bibinfo{person}{Tali Dekel}.} \bibinfo{year}{2021}\natexlab{}.
\newblock \showarticletitle{Layered neural atlases for consistent video editing}.
\newblock \bibinfo{journal}{\emph{ACM Transactions on Graphics}} \bibinfo{volume}{40}, \bibinfo{number}{6} (\bibinfo{year}{2021}), \bibinfo{pages}{210}.
\newblock


\bibitem[Khandelwal(2023)]%
        {khandelwal_infusion_2023}
\bibfield{author}{\bibinfo{person}{Anant Khandelwal}.} \bibinfo{year}{2023}\natexlab{}.
\newblock \showarticletitle{{InFusion}: {Inject} and attention fusion for multi concept zero shot text based video editing}.
\newblock \bibinfo{journal}{\emph{arXiv:2308.00135}} (\bibinfo{year}{2023}).
\newblock


\bibitem[Kim et~al\mbox{.}(2024)]%
        {kim_collaborative_2024}
\bibfield{author}{\bibinfo{person}{Subin Kim}, \bibinfo{person}{Kyungmin Lee}, \bibinfo{person}{June~Suk Choi}, \bibinfo{person}{Jongheon Jeong}, \bibinfo{person}{Kihyuk Sohn}, {and} \bibinfo{person}{Jinwoo Shin}.} \bibinfo{year}{2024}\natexlab{}.
\newblock \showarticletitle{Collaborative score distillation for consistent visual editing}. In \bibinfo{booktitle}{\emph{Advances in {Neural} {Information} {Processing} {Systems}}}.
\newblock


\bibitem[Kolkin et~al\mbox{.}(2019)]%
        {kolkin_style_2019}
\bibfield{author}{\bibinfo{person}{Nicholas Kolkin}, \bibinfo{person}{Jason Salavon}, {and} \bibinfo{person}{Gregory Shakhnarovich}.} \bibinfo{year}{2019}\natexlab{}.
\newblock \showarticletitle{Style transfer by relaxed optimal transport and self-similarity}. In \bibinfo{booktitle}{\emph{Proceedings of {IEEE} {Conference} on {Computer} {Vision} and {Pattern} {Recognition}}}. \bibinfo{pages}{10051--10060}.
\newblock


\bibitem[Li et~al\mbox{.}(2017)]%
        {li_universal_2017}
\bibfield{author}{\bibinfo{person}{Yijun Li}, \bibinfo{person}{Chen Fang}, \bibinfo{person}{Jimei Yang}, \bibinfo{person}{Zhaowen Wang}, \bibinfo{person}{Xin Lu}, {and} \bibinfo{person}{Ming-Hsuan Yang}.} \bibinfo{year}{2017}\natexlab{}.
\newblock \showarticletitle{Universal style transfer via feature transforms}. In \bibinfo{booktitle}{\emph{Advances in {Neural} {Information} {Processing} {Systems}}}. \bibinfo{pages}{385--395}.
\newblock


\bibitem[Liu et~al\mbox{.}(2019)]%
        {liu_few-shot_2019}
\bibfield{author}{\bibinfo{person}{Ming-Yu Liu}, \bibinfo{person}{Xun Huang}, \bibinfo{person}{Arun Mallya}, \bibinfo{person}{Tero Karras}, \bibinfo{person}{Timo Aila}, \bibinfo{person}{Jaakko Lehtinen}, {and} \bibinfo{person}{Jan Kautz}.} \bibinfo{year}{2019}\natexlab{}.
\newblock \showarticletitle{Few-shot unsupervised image-to-image translation}. In \bibinfo{booktitle}{\emph{Proceedings of {IEEE} {Conference} on {Computer} {Vision} and {Pattern} {Recognition}}}. \bibinfo{pages}{10551--10560}.
\newblock


\bibitem[Loshchilov and Hutter(2017)]%
        {loshchilov_decoupled_2017}
\bibfield{author}{\bibinfo{person}{Ilya Loshchilov} {and} \bibinfo{person}{Frank Hutter}.} \bibinfo{year}{2017}\natexlab{}.
\newblock \showarticletitle{Decoupled weight decay regularization}.
\newblock \bibinfo{journal}{\emph{arXiv:1711.05101}} (\bibinfo{year}{2017}).
\newblock


\bibitem[Paszke et~al\mbox{.}(2019)]%
        {paszke_pytorch_2019}
\bibfield{author}{\bibinfo{person}{Adam Paszke}, \bibinfo{person}{Sam Gross}, \bibinfo{person}{Francisco Massa}, \bibinfo{person}{Adam Lerer}, \bibinfo{person}{James Bradbury}, \bibinfo{person}{Gregory Chanan}, \bibinfo{person}{Trevor Killeen}, \bibinfo{person}{Zeming Lin}, \bibinfo{person}{Natalia Gimelshein}, \bibinfo{person}{Luca Antiga}, \bibinfo{person}{Alban Desmaison}, \bibinfo{person}{Andreas Kopf}, \bibinfo{person}{Edward Yang}, \bibinfo{person}{Zachary DeVito}, \bibinfo{person}{Martin Raison}, \bibinfo{person}{Alykhan Tejani}, \bibinfo{person}{Sasank Chilamkurthy}, \bibinfo{person}{Benoit Steiner}, \bibinfo{person}{Lu Fang}, \bibinfo{person}{Junjie Bai}, {and} \bibinfo{person}{Soumith Chintala}.} \bibinfo{year}{2019}\natexlab{}.
\newblock \showarticletitle{{PyTorch}: {An} imperative style, high-performance deep learning library}. In \bibinfo{booktitle}{\emph{Advances in {Neural} {Information} {Processing} {Systems}}}, Vol.~\bibinfo{volume}{32}.
\newblock


\bibitem[Poole et~al\mbox{.}(2022)]%
        {poole_dreamfusion_2022}
\bibfield{author}{\bibinfo{person}{Ben Poole}, \bibinfo{person}{Ajay Jain}, \bibinfo{person}{Jonathan~T. Barron}, {and} \bibinfo{person}{Ben Mildenhall}.} \bibinfo{year}{2022}\natexlab{}.
\newblock \showarticletitle{{DreamFusion}: {Text}-to-{3D} using {2D} diffusion}.
\newblock \bibinfo{journal}{\emph{arXiv:2209.14988}} (\bibinfo{year}{2022}).
\newblock


\bibitem[Praun et~al\mbox{.}(2001)]%
        {praun_real-time_2001}
\bibfield{author}{\bibinfo{person}{Emil Praun}, \bibinfo{person}{Hugues Hoppe}, \bibinfo{person}{Matthew Webb}, {and} \bibinfo{person}{Adam Finkelstein}.} \bibinfo{year}{2001}\natexlab{}.
\newblock \showarticletitle{Real-time hatching}. In \bibinfo{booktitle}{\emph{{SIGGRAPH} {Conference} {Proceedings}}}. \bibinfo{pages}{581--586}.
\newblock


\bibitem[Rav-Acha et~al\mbox{.}(2008)]%
        {rav-acha_unwrap_2008}
\bibfield{author}{\bibinfo{person}{Alex Rav-Acha}, \bibinfo{person}{Pushmeet Kohli}, \bibinfo{person}{Carsten Rother}, {and} \bibinfo{person}{Andrew Fitzgibbon}.} \bibinfo{year}{2008}\natexlab{}.
\newblock \showarticletitle{Unwrap {Mosaics}: {A} new representation for video editing}.
\newblock \bibinfo{journal}{\emph{ACM Transactions on Graphics}} \bibinfo{volume}{27}, \bibinfo{number}{3} (\bibinfo{year}{2008}), \bibinfo{pages}{17}.
\newblock


\bibitem[Rombach et~al\mbox{.}(2022)]%
        {rombach_high-resolution_2022}
\bibfield{author}{\bibinfo{person}{Robin Rombach}, \bibinfo{person}{Andreas Blattmann}, \bibinfo{person}{Dominik Lorenz}, \bibinfo{person}{Patrick Esser}, {and} \bibinfo{person}{Björn Ommer}.} \bibinfo{year}{2022}\natexlab{}.
\newblock \showarticletitle{High-resolution image synthesis with latent diffusion models}. In \bibinfo{booktitle}{\emph{Proceedings of {IEEE} {Conference} on {Computer} {Vision} and {Pattern} {Recognition}}}. \bibinfo{pages}{10684--10695}.
\newblock


\bibitem[Ronneberger et~al\mbox{.}(2015)]%
        {ronneberger_u-net_2015}
\bibfield{author}{\bibinfo{person}{Olaf Ronneberger}, \bibinfo{person}{Philipp Fischer}, {and} \bibinfo{person}{Thomas Brox}.} \bibinfo{year}{2015}\natexlab{}.
\newblock \showarticletitle{U-{Net}: {Convolutional} networks for biomedical image segmentation}. In \bibinfo{booktitle}{\emph{Proceedings of {International} {Conference} on {Medical} {Image} {Computing} and {Computer} {Assisted} {Intervention}}}. \bibinfo{pages}{234--241}.
\newblock


\bibitem[Ruder et~al\mbox{.}(2018)]%
        {ruder_artistic_2018}
\bibfield{author}{\bibinfo{person}{Manuel Ruder}, \bibinfo{person}{Alexey Dosovitskiy}, {and} \bibinfo{person}{Thomas Brox}.} \bibinfo{year}{2018}\natexlab{}.
\newblock \showarticletitle{Artistic style transfer for videos and spherical images}.
\newblock \bibinfo{journal}{\emph{International Journal of Computer Vision}} \bibinfo{volume}{126}, \bibinfo{number}{11} (\bibinfo{year}{2018}), \bibinfo{pages}{1199--1219}.
\newblock


\bibitem[Salisbury et~al\mbox{.}(1997)]%
        {salisbury_orientable_1997}
\bibfield{author}{\bibinfo{person}{Michael~P Salisbury}, \bibinfo{person}{Michael~T Wong}, \bibinfo{person}{John~F Hughes}, {and} \bibinfo{person}{David~H Salesin}.} \bibinfo{year}{1997}\natexlab{}.
\newblock \showarticletitle{Orientable textures for image-based pen-and-ink illustration}. In \bibinfo{booktitle}{\emph{{SIGGRAPH} {Conference} {Proceedings}}}. \bibinfo{pages}{401--406}.
\newblock


\bibitem[Shin et~al\mbox{.}(2024)]%
        {shin_edit--video_2024}
\bibfield{author}{\bibinfo{person}{Chaehun Shin}, \bibinfo{person}{Heeseung Kim}, \bibinfo{person}{Che~Hyun Lee}, \bibinfo{person}{Sang-gil Lee}, {and} \bibinfo{person}{Sungroh Yoon}.} \bibinfo{year}{2024}\natexlab{}.
\newblock \showarticletitle{Edit-{A}-{Video}: {Single} video editing with object-aware consistency}. In \bibinfo{booktitle}{\emph{Proceedings of {Asian} {Conference} on {Machine} {Learning}}}. \bibinfo{pages}{1215--1230}.
\newblock


\bibitem[Siarohin et~al\mbox{.}(2019a)]%
        {siarohin_animating_2019}
\bibfield{author}{\bibinfo{person}{Aliaksandr Siarohin}, \bibinfo{person}{Stéphane Lathuilière}, \bibinfo{person}{Sergey Tulyakov}, \bibinfo{person}{Elisa Ricci}, {and} \bibinfo{person}{Nicu Sebe}.} \bibinfo{year}{2019}\natexlab{a}.
\newblock \showarticletitle{Animating arbitrary objects via deep motion transfer}. In \bibinfo{booktitle}{\emph{Proceedings of {IEEE} {Conference} on {Computer} {Vision} and {Pattern} {Recognition}}}. \bibinfo{pages}{2377--2386}.
\newblock


\bibitem[Siarohin et~al\mbox{.}(2019b)]%
        {siarohin_first_2019}
\bibfield{author}{\bibinfo{person}{Aliaksandr Siarohin}, \bibinfo{person}{Stéphane Lathuilière}, \bibinfo{person}{Sergey Tulyakov}, \bibinfo{person}{Elisa Ricci}, {and} \bibinfo{person}{Nicu Sebe}.} \bibinfo{year}{2019}\natexlab{b}.
\newblock \showarticletitle{First order motion model for image animation}. In \bibinfo{booktitle}{\emph{Advances in {Neural} {Information} {Processing} {Systems}}}.
\newblock


\bibitem[Simonyan and Zisserman(2014)]%
        {simonyan_very_2014}
\bibfield{author}{\bibinfo{person}{Karen Simonyan} {and} \bibinfo{person}{Andrew Zisserman}.} \bibinfo{year}{2014}\natexlab{}.
\newblock \showarticletitle{Very deep convolutional networks for large-scale image recognition}.
\newblock \bibinfo{journal}{\emph{arXiv:1409.1556}} (\bibinfo{year}{2014}).
\newblock


\bibitem[Texler et~al\mbox{.}(2020a)]%
        {texler_arbitrary_2020}
\bibfield{author}{\bibinfo{person}{Ondřej Texler}, \bibinfo{person}{David Futschik}, \bibinfo{person}{Jakub Fišer}, \bibinfo{person}{Michal Lukáč}, \bibinfo{person}{Jingwan Lu}, \bibinfo{person}{Eli Shechtman}, {and} \bibinfo{person}{Daniel Sýkora}.} \bibinfo{year}{2020}\natexlab{a}.
\newblock \showarticletitle{Arbitrary style transfer using neurally-guided patch-based synthesis}.
\newblock \bibinfo{journal}{\emph{Computers \& Graphics}}  \bibinfo{volume}{87} (\bibinfo{year}{2020}), \bibinfo{pages}{62--71}.
\newblock


\bibitem[Texler et~al\mbox{.}(2020b)]%
        {texler_interactive_2020}
\bibfield{author}{\bibinfo{person}{Ondřej Texler}, \bibinfo{person}{David Futschik}, \bibinfo{person}{Michal Kučera}, \bibinfo{person}{Ondřej Jamriška}, \bibinfo{person}{Šárka Sochorová}, \bibinfo{person}{Menclei Chai}, \bibinfo{person}{Sergey Tulyakov}, {and} \bibinfo{person}{Daniel Sýkora}.} \bibinfo{year}{2020}\natexlab{b}.
\newblock \showarticletitle{Interactive video stylization using few-shot patch-based training}.
\newblock \bibinfo{journal}{\emph{ACM Transactions on Graphics}} \bibinfo{volume}{39}, \bibinfo{number}{4} (\bibinfo{year}{2020}), \bibinfo{pages}{73}.
\newblock


\bibitem[Ulyanov et~al\mbox{.}(2016)]%
        {ulyanov_instance_2016}
\bibfield{author}{\bibinfo{person}{Dmitry Ulyanov}, \bibinfo{person}{Andrea Vedaldi}, {and} \bibinfo{person}{Victor Lempitsky}.} \bibinfo{year}{2016}\natexlab{}.
\newblock \showarticletitle{Instance normalization: {The} missing ingredient for fast stylization}.
\newblock \bibinfo{journal}{\emph{arXiv:1607.08022}} (\bibinfo{year}{2016}).
\newblock


\bibitem[Wang et~al\mbox{.}(2019)]%
        {wang_few-shot_2019}
\bibfield{author}{\bibinfo{person}{Ting-Chun Wang}, \bibinfo{person}{Ming-Yu Liu}, \bibinfo{person}{Andrew Tao}, \bibinfo{person}{Guilin Liu}, \bibinfo{person}{Bryan Catanzaro}, {and} \bibinfo{person}{Jan Kautz}.} \bibinfo{year}{2019}\natexlab{}.
\newblock \showarticletitle{Few-shot video-to-video synthesis}. In \bibinfo{booktitle}{\emph{Advances in {Neural} {Information} {Processing} {Systems}}}. \bibinfo{pages}{5014--5025}.
\newblock


\bibitem[Yang et~al\mbox{.}(2023)]%
        {yang_rerender_2023}
\bibfield{author}{\bibinfo{person}{Shuai Yang}, \bibinfo{person}{Yifan Zhou}, \bibinfo{person}{Ziwei Liu}, {and} \bibinfo{person}{Chen~Change Loy}.} \bibinfo{year}{2023}\natexlab{}.
\newblock \showarticletitle{Rerender {A} {Video}: {Zero}-shot text-guided video-to-video translation}. In \bibinfo{booktitle}{\emph{{SIGGRAPH} {Asia} {Conference} {Papers}}}. \bibinfo{pages}{95}.
\newblock


\bibitem[Zhang et~al\mbox{.}(2023b)]%
        {zhang_adding_2023}
\bibfield{author}{\bibinfo{person}{Lvmin Zhang}, \bibinfo{person}{Anyi Rao}, {and} \bibinfo{person}{Maneesh Agrawala}.} \bibinfo{year}{2023}\natexlab{b}.
\newblock \showarticletitle{Adding {Conditional} {Control} to {Text}-to-{Image} {Diffusion} {Models}}. In \bibinfo{booktitle}{\emph{Proceedings of {IEEE} {International} {Conference} on {Computer} {Vision}}}. \bibinfo{pages}{3836--3847}.
\newblock


\bibitem[Zhang et~al\mbox{.}(2023a)]%
        {zhang_towards_2023}
\bibfield{author}{\bibinfo{person}{Zicheng Zhang}, \bibinfo{person}{Bonan Li}, \bibinfo{person}{Xuecheng Nie}, \bibinfo{person}{Congying Han}, \bibinfo{person}{Tiande Guo}, {and} \bibinfo{person}{Luoqi Liu}.} \bibinfo{year}{2023}\natexlab{a}.
\newblock \showarticletitle{Towards consistent video editing with text-to-image diffusion models}. In \bibinfo{booktitle}{\emph{Advances in {Neural} {Information} {Processing} {Systems}}}. \bibinfo{pages}{58508--58519}.
\newblock


\bibitem[Zhao et~al\mbox{.}(2024)]%
        {zhao_unipc_2024}
\bibfield{author}{\bibinfo{person}{Wenliang Zhao}, \bibinfo{person}{Lujia Bai}, \bibinfo{person}{Yongming Rao}, \bibinfo{person}{Jie Zhou}, {and} \bibinfo{person}{Jiwen Lu}.} \bibinfo{year}{2024}\natexlab{}.
\newblock \showarticletitle{Unipc: {A} unified predictor-corrector framework for fast sampling of diffusion models}.
\newblock \bibinfo{journal}{\emph{Advances in Neural Information Processing Systems}}  \bibinfo{volume}{36} (\bibinfo{year}{2024}).
\newblock


\end{thebibliography}
